\documentclass{CUP-JNL-DCE}%

\usepackage{latexsym}
\usepackage{graphicx}
\usepackage{multicol,multirow}
\usepackage{amsmath,amssymb,amsfonts}
\usepackage{mathrsfs}
\usepackage{amsthm}
\usepackage{apacite}
\usepackage{rotating}
\usepackage{appendix}
\usepackage[authoryear]{natbib}
\usepackage{ifpdf}
\usepackage[T1]{fontenc}
\usepackage{times}
\usepackage{sourcesanspro}
\usepackage{newtxmath}
\usepackage{textcomp}%
\usepackage{xcolor}%
\usepackage{hyperref}
\usepackage{cleveref}
\usepackage{subcaption}
\usepackage{siunitx}
\usepackage[normalem]{ulem}

\usepackage{tikz}
\usetikzlibrary{arrows.meta,trees,calc,patterns,decorations.pathmorphing,decorations.markings}

\articletype{SURVEY PAPER}
\jname{Data-Centric Engineering}
\jyear{2023}
\jdoi{XX.XXXX/dce.XXXX.X}
\begin{document}

\begin{Frontmatter}

% \title[Article Title]{Applications of Physics-Enhanced Machine Learning in Structural Mechanics: A Survey}
\title[Article Title]{Discussing the Spectrum of Physics-Enhanced Machine Learning; a Survey on Structural Mechanics Applications}

% There is no need to include ORCID IDs in your .pdf; this information is captured by the submission portal when a manuscript is submitted. 
\author*[1]{Marcus Haywood-Alexander}\orcid{0000-0003-3384-2346}\email{mhaywood@ethz.ch}
\author[2,3]{Wei Liu}\orcid{0000-0002-1103-7699}
\author[1,3]{Kiran Bacsa}\orcid{0000-0002-0834-185X}
\author[4,5]{Zhilu Lai}\orcid{0000-0001-6227-6123}
\author[1,3]{Eleni Chatzi}\orcid{0000-0002-6870-240X}

\authormark{M. Haywood-Alexander \textit{et al}.}
\address[1]{\orgdiv{Department of Civil, Environmental and Geomatic Engineering}, \orgname{ETH Zürich}, \orgaddress{\city{Zürich}, \postcode{8049}, \country{Switzerland}}}
\address[2]{\orgdiv{Department of Industrial Systems Engineering and Management}, \orgname{National University of Singapore}, \country{Singapore}}
\address[3]{\orgdiv{Future Resilient Systems}, \orgname{Singapore-ETH Centre}, \country{Singapore}}
\address[4]{\orgdiv{Internet of Things Thrust}, \orgname{HKUST(GZ)}, \country{Guangzhou, People’s Republic of China}}
\address[5]{\orgdiv{Department of Civil and Environmental Engineering}, \orgname{HKUST}, \country{Hong Kong, People’s
Republic of China}}

\keywords{physics enhanced, physics-based, physics-guided, physics-encoded, data driven, hybrid learning, structural mechanics}

% Abstract: 250 words
\abstract{
The intersection of physics and machine learning has given rise to the \emph{physics-enhanced} machine learning (\emph{PEML}) paradigm, aiming to improve the capabilities and reduce the individual shortcomings of data- or physics-only methods. %
In this paper, the spectrum of physics-enhanced machine learning methods, expressed across the defining axes of physics and data, is discussed by engaging in a comprehensive exploration of its characteristics, usage, and motivations. %
In doing so, we present a survey of recent applications and developments of \emph{PEML} techniques, revealing the potency of \emph{PEML} in addressing complex challenges. We further demonstrate application of select such schemes on the simple working example of a single degree-of-freedom Duffing oscillator, which allows to highlight the individual characteristics and motivations of different `genres' of \emph{PEML} approaches. %
To promote collaboration and transparency, and to provide practical examples for the reader, the code generating these working examples is provided alongside this paper. %
As a foundational contribution, this paper underscores the significance of \emph{PEML} in pushing the boundaries of scientific and engineering research, underpinned by the synergy of physical insights and machine learning capabilities. %
}

\begin{policy}[Impact Statement]
    This paper discusses methods born from fusion of physics and machine learning, known as \emph{physics-enhanced} machine learning (\emph{PEML}) schemes. %
    By considering their characteristics, this work clarifies and categorizes \emph{PEML} techniques, aiding researchers and users to targetedly select methods on the basis of specific problem characteristics and requirements. %
    The discussion of \emph{PEML} techniques is framed around a survey of recent applications/developments of \emph{PEML} in the field of structural mechanics. %
    A running example of a Duffing oscillator is used to highlight the traits and potential of diverse \emph{PEML} approaches. %
    Additionally, code is provided to foster transparency and collaboration. %
    The work advocates the pivotal role of \emph{PEML} in advancing computing for engineering through the merger of physics based knowledge and machine learning capabilities. %
\end{policy}

\end{Frontmatter}

% ------------------------------------------------------------------------------------
\section{Introduction}
\label{sec:introduction}

%% Motivation behind PE-ML
With the increase in both computing power and data availability, machine learning (ML) and deep learning (DL) are in scientific and engineering applications \citep{reich1997machine,hey2020machine,zhong2021machine,cuomo2022scientific}. %
Such methods have shown enormous potential in yielding efficient and accurate estimates over highly complex domains, such as those with high-dimensionality, or ill-posed problem definitions. %
The use of data-driven methods is wide reaching in science, from fields such as fluid dynamics \citep{zhang2015machine}, geoscience \citep{bergen2019machine}, bioinformatics \citep{olson2018data}, and more \citep{brunton2022data}. %
Data-driven schemes are particularly suited for the case of monitored systems, where availability of data is ensured via measurement of engineering quantities through the use of appropriate sensors \citep{sohn2003review,lynch2007overview,farrar2007introduction}. %

%In a solely-data-driven context, the aforementioned availability of data allows one to generate accurate estimates of \emph{model solutions}. %
However, such data-driven models are known to be restricted to the domain of the instance in which the data was collected; i.e.\ they lack generalisability \citep{o2019physics,karniadakis2021physics}, as a result of a lack of physical connotation. %
This challenge is often met when dealing with data-driven approaches for environmental and operational normalisation \citep{cross2011cointegration_a,cross2011cointegration_b,avendano2017gaussian}; it is impossible, or impractical, to collect data over the full environmental/operational (E/O) envelope \citep{figueiredo2011machine}. %
Particularly to what concerns data gathered from large-scale engineered systems, it is common to meet a scarcity of training samples across a system's comprehensive operational envelope \citep{Sohn2007}. %
These variables frequently exhibit intricate and non-stationary patterns changing over time. %
Consequently, the limited pool of labeled samples available for training or cross-validation can fall short of accurately capturing the intrinsic relationships for scientific discovery tasks, potentially resulting in misleading extrapolations \citep{d2019machine}. %
This scarcity of representative samples sets scientific problems apart from more mainstream concerns like language translation or object recognition, where copious amounts of labeled or unlabeled data have underpinned recent advancements in deep learning \citep{jordan2015machine, sharifani2023machine}. %
Discussions on and examples of the challenge posed by comparatively small datasets in scientific machine learning can be found in \citep{zhang2018strategy}, \citep{shaikhina2015machine}, and \citep{dou2023machine}. %

While black box data-driven schemes are often sufficient for delivering an actionable system model, able to act as an estimator or classifier, a common pursuit within the context of mechanics lies in knowledge discovery \citep{geyer2021explainable,naser2021engineer,cuomo2022scientific}. In this case, it is imperative to deliver models that are explainable/interpretable and generalisable \citep{linardatos2020explainable}. %
This entails revealing and comprehending the cause-and-effect mechanisms underpinning the workings of a particular engineered system. %
Consequently, even if a black-box model attains marginally superior accuracy, its inability to unravel the fundamental underlying processes renders it inadequate for furthering downstream scientific applications \citep{langley1994selection}. %
Conversely, an interpretable model rooted in explainable theories is better poised to guard against the learning of spurious data-driven patterns that lack interpretability \citep{molnar2020interpretable}. %
This becomes particularly crucial for practices where predictive models are of the essence for risk-based assessment and decision support, such as the domains of structural health monitoring \citep{farrar2012structural} and resilience \citep{shadabfar2022resilience}.

In modelling complex systems, there is a need for a balanced approach that combines physics-based and data-driven models \citep{pawar2021model}. %
Modern engineering systems, involve complex materials, geometries and often intricate energy harvesting and vibration mitigation mechanisms, which may be associated with complex mechanics and failure patterns \citep{duenas2009cascading,van2012computational,kim2017failure}. This results in behaviour that cannot be trivially described purely on the basis of data observations or via common, and often simplified, modelling assumptions. In efficiently modelling such systems, a viable approach is to integrate the aspect of physics, which is linked to forward modelling with the aspect of learning from data (via machine learning tools), which can account for modelling uncertainties and imprecision. This fusion has been referred to via the term 'physics-enhanced machine learning (\emph{PEML})' \citep{faroughi2022physics}, which we also adopt herein. This term is used to denote that, in some form, prior physics knowledge is embedded to the learner  \citep{o2019physics,choudhary2020physics,xiaowei2021physics}. %
which typically results in more interpretable models. 
%A main goal of such techniques is to blend physical principles and ML in a `seamless’ manner, meaning that the fusion of both elements is well-integrated, efficient, and generalizable. %

%% Uses of structural mechanics, and the need for more complex models
In this work, we focus on applications of \emph{PEML} in the domain of structural mechanics; a field that impacts the design, building, monitoring, maintenance, and disuse of critical structures and infrastructures. %
Some of the greatest impact comes from large scale infrastructure, such as bridges, wind turbines, and transport systems. %
However, accurate and robust numerical models of complex structures are non-trivial to establish for tasks such as Digital Twinning (DT) and Structural Health Monitoring (SHM), where both precision and computational efficiency are of the essence  \citep{farrar2012structural,yuan2020machine}. %
%% Machine learning in Structural Mechanics
This has motivated the increased adoption of ML or DL approaches for generating models of such structures, overcoming the challenges presented by the complexity. %
Further to the extended use in the DT and SHM contexts, data-driven approaches have further been adopted for optimising the design of materials and structures \citep{guo2021artificial, sun2021machine}. %
Multi-scale modelling of structures has also benefited from use of ML approaches, typically via replacement of computationally-costly representative volume element simulations with ML models, such as neural networks \citep{huang2020machine} or support vector regression and random forest regression \citep{reimann2019modeling}. %

\begin{figure}[h!]
    \centering
    \includegraphics[width=1\textwidth]{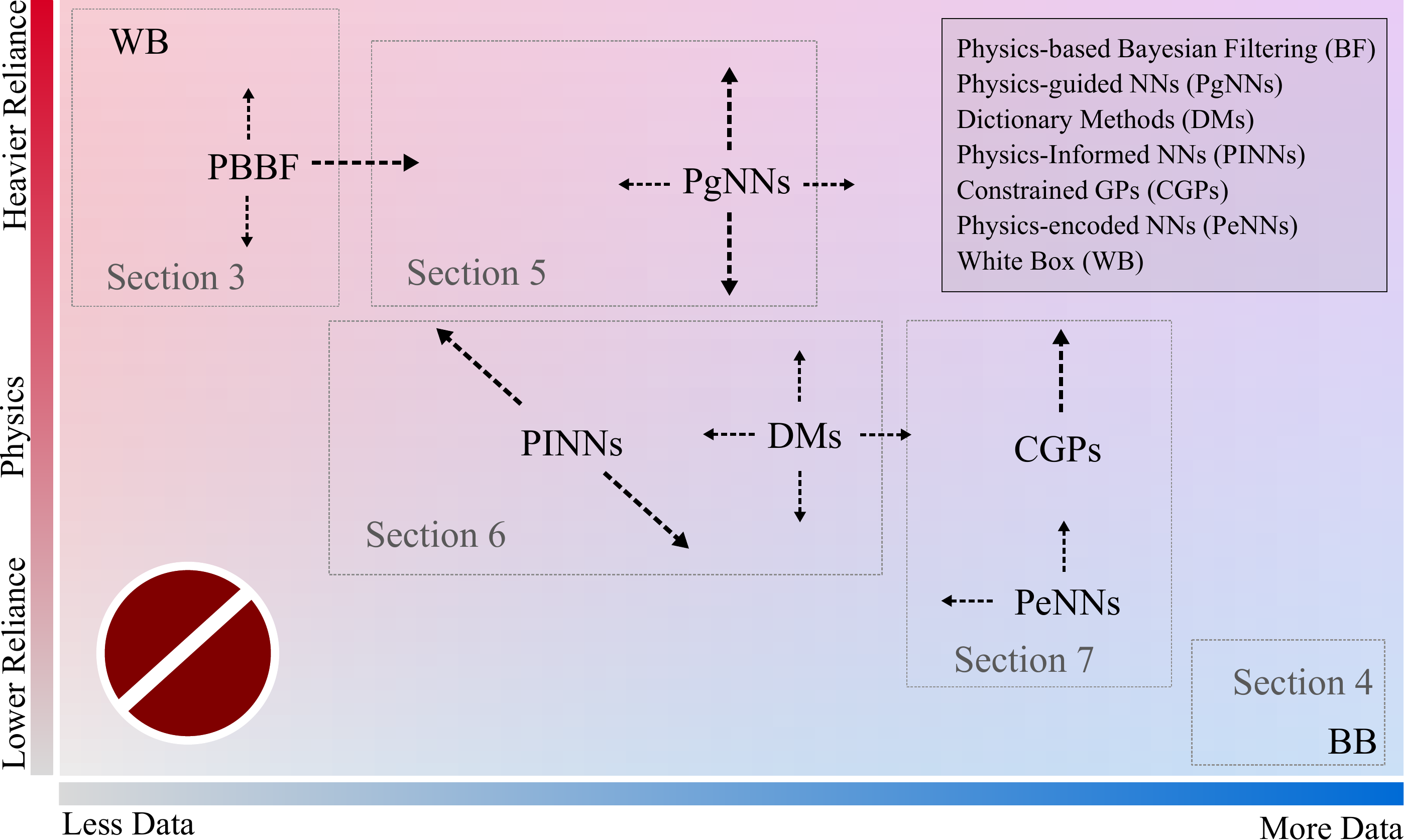}
    \caption{The spectrum of Physics-Enhanced Machine Learning (\emph{PEML}) schemes surveyed in this paper}
    \label{fig:paper_layout}
\end{figure}

%% How physics is embedded
In order to contextualise \emph{PEML} for use within the realm of structural mechanics applications, we here employ a characterisation that adopts the idea of a spectrum, as opposed to a categorization in purely white, black and grey box models. This is inspired by the categorisation put forth in the recent works of \cite{cross2022physics}, which discusses the placement of \emph{PEML} methods on a two-dimensional spectrum of physics and data, and \cite{faroughi2022physics} which categorizes these schemes based on the implementation of physics within the ML architecture. %
In the context of the previously-used one-dimensional spectrum, the `darker' end of the said spectrum relies more heavily on data, whilst the `lighter' end relies more heavily on the portion of physics that is considered known (Figure \ref{fig:paper_layout}). %
One can envision this one-dimensional spectrum lying equivalently along the diagonal from the red (top left), to the blue (bottom right) corners of the two-dimensional spectrum. %
Under this definition, `off-the-shelf' ML approaches more customarily fit the black end of the spectrum, while purely analytical solutions would sit on white end of the spectrum \citep{rebillat2023physically}. %
Generally, the position along this spectrum is driven by both the amount of data available, and the level of physics constraints that are applied.  However, it is important to note, that the inclusion of data is not a requirement for \emph{PEML}. %
An example of the latter is found in methods such as \emph{physics-informed neural networks} (PINNs) \citep{raissi2019physics}, which exploit the capabilities of ML methods to act as forward modellers, where no observations or measured data are necessarily used for the formulation of the loss function. %
In these cases, prescribed boundary/initial conditions, physics equations, and system inputs are provided and the ML algorithm is used to `learn' the solutions; one such method is the observation-absent \emph{PINN} \citep{rezaei2022mixed}. %
% In the past decade, machine learning methods have been used with known boundary conditions to solve complex equations, which one can argue is a similar solution approach as numerical solvers. %
%However, in more modern approaches and discussions to \emph{PEML}, descriptions have been improved to include the level of physics and data included as two distinct spectra. %

The reliance on the physics can be quantified in terms of the level of strictness of the physics model prescription. %
The level of strictness refers to the degree to which the prescribed model form incorporates and adheres to the underlying physical principles, and concurrently defines the set of systems which the prescribed model can emulate. %
For example, when system parameters are assumed known, the physics is more strictly prescribed. %
% This categorisation concurrently defines the set of systems which the prescribed physics model can emulate - a high level of strictness results in a narrower set of systems, and \emph{vice-versa}. %
Using a solid mechanics example, a strictly prescribed model would correspond to use of a specified a Finite Element model as the underlying physics structure. This strictness is somewhat relaxed when it is assumed that the model parameters are uncertain and subject to updating \citep{papadimitriou2004bayesian}. %
An example of a low degree of strictness corresponds to prescribing the system output as a function of the derivative of system inputs with respect to time; such a more loose prescription would be capable of emulating structural dynamics system \citep{bacsa2023symplectic}, as well as further system and problem types, such as heat transfer \citep{dhadphale2022neural}, or virus spreading \citep{nunez2023forecasting}. %
In this work, the vertical axis of the \emph{PEML} spectrum is defined in terms of the reliance on the imposed physics-based model form, which is earlier referred to as the level of strictness.
A separate notion to consider, which is not reflected in the included axes, pertains to the level of constraint of the employed \emph{PEML} architecture, which describes the degree to which the learner must adhere to a prescribed model. %
As an example, residual modelling techniques \citep{christodoulou2007structural} have a relatively low level of constraint, as the solution space is not limited to that which is posited by the physics model. 
The combination of the strictness in the prescription of a model form and the learner constraints defines the overall \emph{flexibility} of the \emph{PEML} scheme; this refers to its capability to emulate systems of varying types and complexities \citep{karniadakis2021physics}. 
%

% Define the description of level of physics knowledge
When selecting the type, or `genre' of \emph{PEML} model, the confidence in the physics that is known \emph{a priori} guides the selection of the appropriate reliance on physics in the form of the level of strictness of the prescribed model and/or constraints of the learner. %
Different prior knowledge can be in the form of an appropriate model structure, i.e.\ an equivalent MDOF system, or an appropriate finite element model, or it could be that appropriate material/property values are prescribed. %
The term appropriate here refers to availability of adequate information on models and parameters, which approximate well the behaviour and traits of the true system. %
If one wishes to delve deeper into such a categorization, the level of knowledge can be appraised in further sub-types, i.e., the discrete number of, or confidence in, known material parameters, or the complexity of the model in relation to the real structure. %

%% Paper introduction
The remainder of the paper is organised into methods corresponding to different collective areas over the \emph{PEML} spectrum, as indicated in \Cref{fig:paper_layout}. %
We initiate with a discussion and corresponding examples that are closer to a \emph{white-box} approach in \Cref{sec:white_box}, where physics-based models of specified form are fused with data via a Bayesian Filtering (BF) approach. %
This is followed by \Cref{sec:black_box}, which shows a brief survey of solely data-driven methods, which embed no prior knowledge on the underlying physics. %
After introducing instances of methods that are situated near the extreme corners of the spectrum (black- and white-box), the main motivation of the paper, namely the overview of \emph{physics-enhanced machine learning} schemes initiates. %
The breakdown into the subsections of \emph{PEML} techniques is driven by a combination of the reliance on the prescription of the physics model form and the method of physics embedding, these are broken down below and their naming conventions explained. %

Firstly, \Cref{sec:guided} surveys and discusses \emph{physics-guided} machine learning (PGML) techniques, in which the physics model prescriptions are embedded as proposed solutions, and act in parallel to the data-driven learner in the full \emph{PEML} model architecture. %
\emph{PGML} schemes steer the learner toward a desired solution by prescribing models with a relatively large degree of strictness, therefore neighbouring the similarly strict construct of BF methods. %
Physics-guided methods often benefit from a reduced data requirement since the physics embedding allows for estimation in absence of dense observations from the system. %
However, depending on the formulation, or the type of method used, such schemes can still suffer from data-sparsity. %
In \Cref{sec:informed}, \emph{physics-informed} methods are presented, which correspond to a a heavier reliance on data,while still retaining a moderate reliance on the prescribed physics. %
These schemes are so named as physics is embedded as prior information, from which an objective or loss function is constructed, which the learner is prompted to follow. %
Compared to physics-guided methods, in physics-informed schemes, the physics is embedded in a less constrained manner, i.e.\ it is weakly imposed. 
In this sense, such schemes are often formed by way of minimising a loss or objective function, which vanishes when all the imposed physics are satisfied. %
% In this sense, a target physical principle is given, to which the learner is prompted to adhere to through the use of a loss or objective function to be minimised. %
The survey portion of the paper concludes with a discussion on \emph{physics-encoded} learners in \Cref{sec:encoded}. %
Physics-encoded methods embed the imposed physics directly within the architecture of the learner, via selection of operators, kernels, or transforms. %
As a result, these methods are often less reliant on the model form (e.g.\ they may simply impose derivatives), but they are highly constrained in the fact that this imposed model is always adhered to. %
The position of \emph{PgNNs} and \emph{PINNs} compared to \emph{constrained Gaussian Processes} (CGPs) is less indicative of the higher requirement of \emph{GCPs} for data, but more indicative of the lower requirement of \emph{PgNNs} and \emph{PINNs} for data, as these methods are capable of proposing viable solutions with fewer data. However, this is dependent on the physics and ML model form, thus arrows are included to indicate their mobility on the spectrum. %

% CGPs and PeNNs both seem to operate in the same way in that they draw from a family of functions, and prescribe the general shape of the function using known physics, as opposed to a high level of physics model strictness.

Throughout the paper, a working example of a single-degree-of-freedom Duffing oscillator, the details of which are offered in \Cref{sec:working_examp}, is used to demonstrate the methods surveyed. %
As previously mentioned, the code used to generate the fundamental versions of these methods is provided alongside this paper in a Github repository\footnote{Python code and data available at: \hyperlink{https://github.com/ETH-IBK-SMECH/PIDyNN}{https://github.com/ETH-IBK-SMECH/PIDyNN}}. %
This code is written in Python and primarily built with the freely available Pytorch package \citep{NEURIPS2019_pytorch}. %

\section{A Working Example}
\label{sec:working_examp}
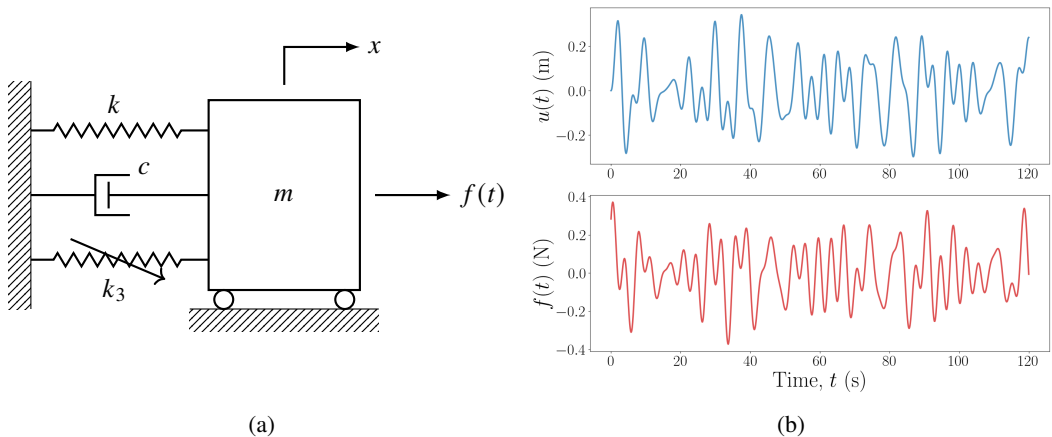
\begin{figure}[h!]
    \centering
    \begin{subfigure}{0.48\textwidth}
        \centering
        \begin{tikzpicture}[every node/.style={draw,outer sep=0pt,thick}]
\tikzstyle{spring}=[thick,decorate,decoration={zigzag,pre length=0.3cm,post length=0.3cm,segment length=6}]
\tikzstyle{nonlin_spring}=[thick,decorate,
    decoration={zigzag,pre length=0.3cm,post length=0.3cm,segment length=6},
    postaction={decorate,decoration={markings,mark=between positions 0.4 and 0.6 step 0.5 with {\draw [->,thick] (-4mm,-3mm) -- (7mm,5mm);}
    }}
    ]
\tikzstyle{damper}=[thick,decoration={markings,  
  mark connection node=dmp,
  mark=at position 0.4 with 
  {
    \node (dmp) [thick,inner sep=0pt,transform shape,rotate=-90,minimum width=15pt,minimum height=5pt,draw=none] {};
    \draw [thick] ($(dmp.north east)+(8pt,0)$) -- (dmp.south east) -- (dmp.south west) -- ($(dmp.north west)+(8pt,0)$);
    \draw [thick] ($(dmp.north)+(0,-5pt)$) -- ($(dmp.north)+(0,5pt)$);
  }
}, decorate]
\tikzstyle{ground}=[fill,pattern=north east lines,draw=none,minimum width=0.75cm,minimum height=0.3cm]

\node (M) [minimum width=2cm, minimum height=2.5cm, xshift=0.5cm] {$m$};

\node (ground) [ground,anchor=north,yshift=-0.25cm,minimum width=2.5cm] at (M.south) {};
\draw (ground.north east) -- (ground.north west);
\draw [thick] (M.south west) ++ (0.2cm,-0.125cm) circle (0.125cm)  (M.south east) ++ (-0.2cm,-0.125cm) circle (0.125cm);

\node (wall) [ground, rotate=-90, minimum width=3cm,yshift=-3cm] {};
\draw (wall.north east) -- (wall.north west);

\draw [spring] (wall.170) -- ($(M.north west)!(wall.170)!(M.south west)$);
\node[draw=none, xshift=1.1cm, yshift=0.35cm] at (wall.170) {$k$};
\draw [damper] (wall.90) -- ($(M.north west)!(wall.90)!(M.south west)$);
\node[draw=none, xshift=1.5cm, yshift=0.35cm] at (wall.90) {$c$};
\draw [nonlin_spring] (wall.10) -- ($(M.north west)!(wall.10)!(M.south west)$);
\node[draw=none, xshift=1.1cm, yshift=-0.4cm] at (wall.10) {$k_3$};

\draw [-latex, thick] (M.east) ++ (0.2cm,0) -- +(1cm,0) node[right, draw=none] {$f(t)$};

\draw [-latex, thick] (M.north) ++(0,0.2cm) -- +(0,0.5cm) -- +(1cm,0.5cm) node[right, draw=none] {$x$};

\end{tikzpicture}
        \caption{}
        \label{fig:duffing_osc_diag}
    \end{subfigure}
    \begin{subfigure}{0.48\textwidth}
        \centering
        \includegraphics[width=\textwidth]{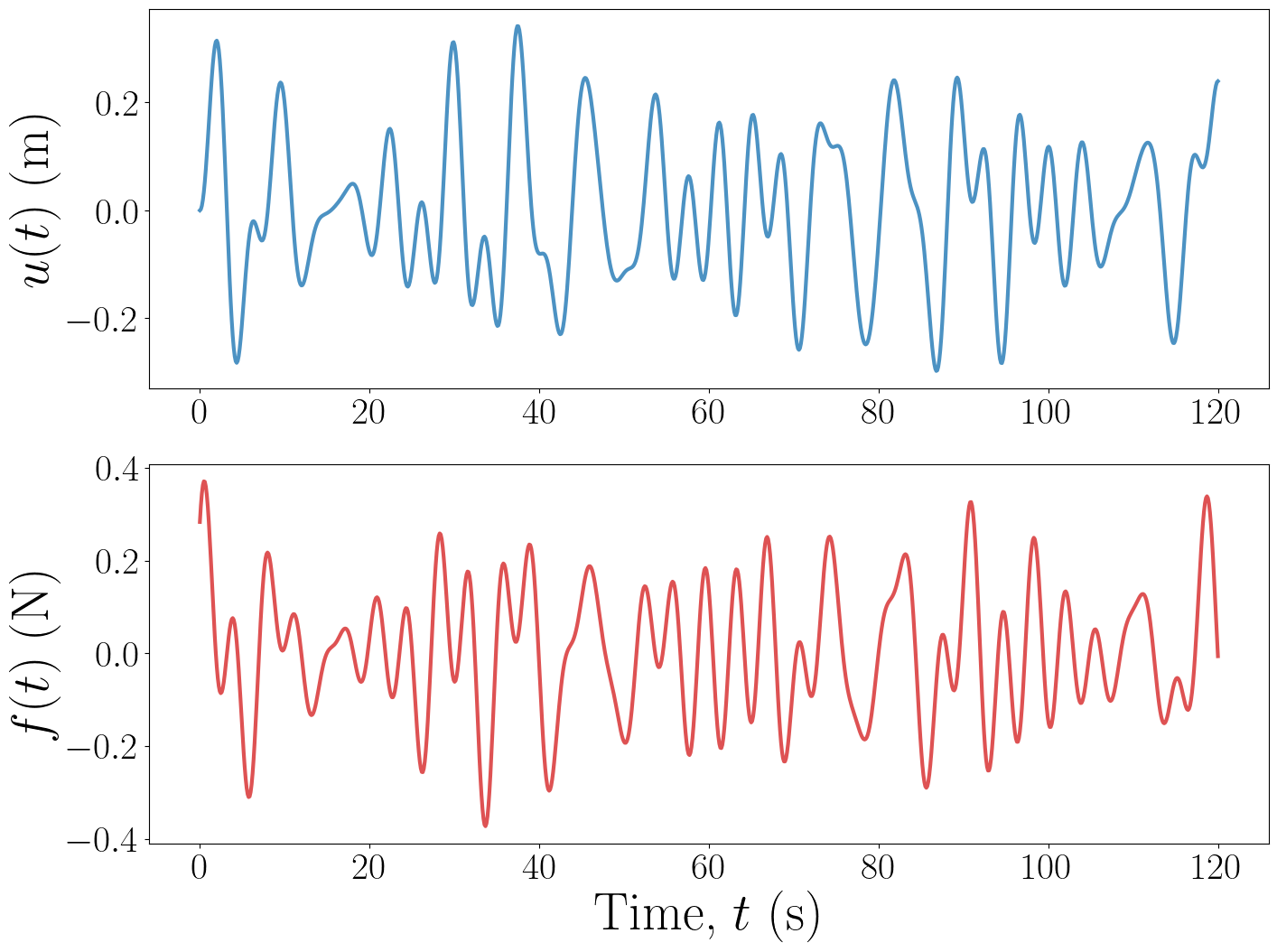}
        \caption{}
        \label{fig:duffing_osc_sig}
    \end{subfigure}
    \caption{(a) Diagram of the working example used throughout this paper, corresponding to a Duffing Oscillator; instances of the (b) displacement (top) and forcing signal (bottom) produced during simulation}
    \label{fig:duffing_osc}
\end{figure}

Aiding the survey and discussion in each aspect of \emph{PEML}, an example of a dynamic system will be used throughout the paper to provide a tangible example for the reader. %
A variety of \emph{PEML} methods will be applied to the presented model, the aim of which is not to showcase any particularly novel applications of the methods, but to help illustrate and discuss the characteristics of the \emph{PEML} variants for a simple example, while highlighting emerging schemes and their placement across the spectrum of \Cref{fig:paper_layout}. %
To this end, we employ a single-degree-of-freedom (SDOF) Duffing oscillator, shown in \Cref{fig:duffing_osc_diag}, as a running example. %
The equation of motion of this oscillator is defined as,
\begin{equation}
    m\ddot{u}(t) + c\dot{u}(t) + ku(t) + k_3 u^3(t) = f(t)
    \label{eq:duffing_osc}
\end{equation}
where the values for the physical parameters $m$, $c$, $k$, and $k_3$ are 10\si{kg}, 1\si{Ns/m}, 15\si{N/m}, and 100\si{N/m^3}, respectively. %
To be consistent with problem formulations in this paper, this is defined in state-space form as follows,
\begin{equation}
    \dot{\mathbf{z}} = \mathbf{A}\mathbf{z} + \mathbf{A}_nu^3 + \mathbf{B}f
    \label{eq:duffing_state_space}
\end{equation}
where $\mathbf{z}=\{u,\dot{u}\}^T$ is the system state, and the state matrices are,
\begin{equation*}
    \mathbf{A} = \begin{bmatrix}
        0 & 1 \\ -m^{-1}k & -m^{-1}c
    \end{bmatrix}, \qquad 
    \mathbf{A}_n = \begin{bmatrix}
        0 \\ -m^{-1}k_3
    \end{bmatrix}, \qquad
    \mathbf{B} = \begin{bmatrix}
        0 \\ m^{-1}
    \end{bmatrix}
\end{equation*}
For this example, the forcing signal of the system consists of a random-phase multi-sine signal containing frequencies of 0.7, 0.85, 1.6 and 1.8 \si{rad/s}. %
The Duffing oscillator is simulated using a 4th-order Runge-Kutta integration. %
The forcing and resulting displacement for 1024 samples at an equivalent sample rate of 8.525Hz is shown in \Cref{fig:duffing_osc_sig}. %
These data are then used as the ground truth for the examples shown throughout the paper. %

\paragraph{A note on data and domains}

The interdisciplinary nature of \emph{PEML} can lead to confusion regarding the terms defining the data and domain for the model. %
To enhance clarity for readers with diverse backgrounds, we provide clarifications on the nomenclature in this paper. %
Firstly, \emph{data} refers to all measured or known values that are used in the overall architecture/methodology, not exclusively that which is used as inputs to the ML model, or as target observations. %
This may include, but is not limited to, measured data, system parameters, and scaling information. %
In the context of the example above, the data encompasses measured values of the state, $\mathbf{z}^*$, and force, $f^*$, along with system parameters ${m,c,k,k_3}$ (where the asterisk denotes observations of a value). %
Importantly, the use of the term \emph{observation data} is akin to the classic ML definition of \emph{training data}, which is the scope of data used in traditional learning paradigms which minimise the discrepancy between the model output and some observed target values. %
This change is employed here as the training stages in many instances of physics-enhanced machine learning demonstrate the learner's ability to make predictions beyond the scope of these observations of target values. %

\begin{figure}[h!]
    \centering
    \includegraphics[width=0.9\textwidth]{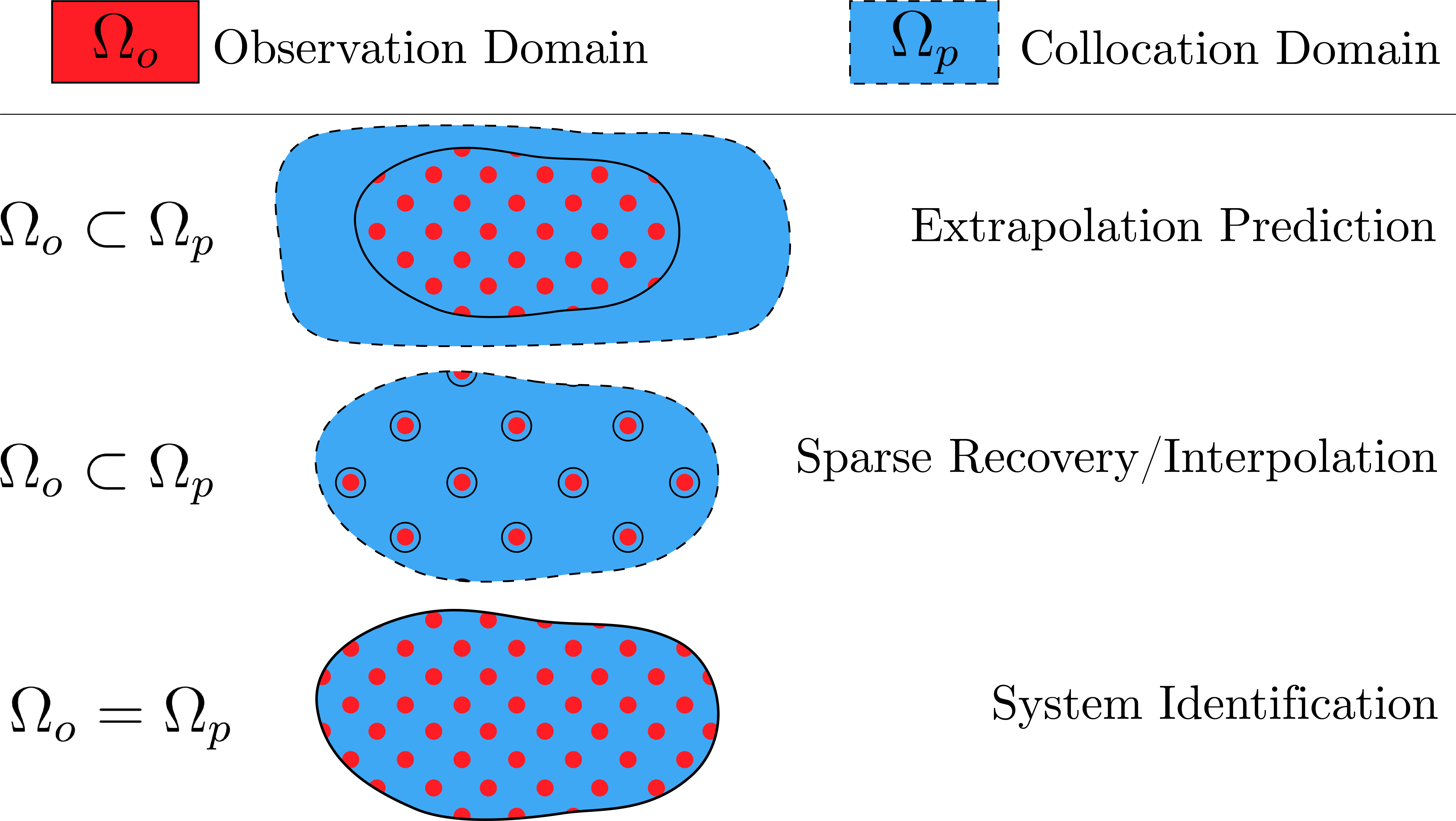}
    \caption{Visualisation of domain definitions for schemes and motivations that can employ \emph{PEML}. The blue areas represent the continuous collocation domain, and the red dots represent the coverage and sparsity of the discrete observation domain. The dashed and solid lines represent the scope of the collocation and observation domains, respectively}
    \label{fig:pinn_domains}
\end{figure}

This goal of extending the scope of prediction also prompts a clarification of the term \emph{domain}. %
The \emph{domain} here is similar to the definition of the domain of a function, representing the set of values passed as input to the model — in this case, the set of time values, $t$. %
The domain where \emph{measured} values of the model output are available, is termed the \emph{observation domain}, $\Omega_o$. %
The overall domain in which the model is \emph{trained}, and predictions can be made, is the \emph{collocation domain}, $\Omega_c$. %
For example, if one provides measurements of the state for the first third of the signal in \Cref{fig:duffing_osc_sig} but proposes the model to learn (and therefore predict) over the full signal range, the observation domain would be $\Omega_o\in 0\leq t\leq40s$ and the collocation domain $\Omega_c\in 0\leq t\leq120s$. %
It's crucial to note that these domains are not restricted to a range (scope), and the discrete nature of the observation domain influences the motivation for interpolation schemes. %
For example, in sparse data recovery schemes, the observation domain can be defined in a discrete manner. %
% and when defining them, the application's purpose should be considered -- data from the observation domain will be discrete, but the collocation domain will be continuous. %
\Cref{fig:pinn_domains} provides a visualisation of commonly used domain types for a selection of schemes which can employ \emph{PEML} methods. %

%----------------------------------------------------------------
% SECTION 1:
%----------------------------------------------------------------
\section{White Box Case - Physics-based Bayesian Filtering}
\label{sec:white_box}

Prior to overviewing the mentioned \emph{PEML} classes and their adoption within the SHM and twinning context, we briefly recall a class of methods, which is situated near the \emph{white-box} end of the spectrum in \Cref{fig:paper_layout}, i.e., Bayesian Filtering (BF). Perhaps one of the most typical examples of a hybrid approach to monitoring of dynamical systems is delivered in such BF estimators, which couple a system model (typically in state-space form), with sparse and noisy monitoring data. The employed state-space model can be either derived via a data-driven approach, e.g. via use of a system identification approach such as a Stochastic Subspace Identification \citep{peeters2001stochastic}, or alternatively, it may be inferred on the basis of an a priori assumed numerical (e.g. finite element) model. We here refer to the former case, which we refer to as physics-based Bayesian Filtering. Such Bayesian filters can be used for estimation tasks of different complexity, including pure response (state) estimation, joint or dual state-parameter estimation \citep{Chatzi2009}, input-state estimation \citep{vettori2023assessment,EFTEKHARAZAM2015866,MAES2018292,SEDEHI2019659}, joint state-parameter-input identification \citep{dertimanis2019input}, or damage detection \citep{erazo2019vibration}. Bayesian filters draw their potency from their capacity to deal with uncertainties stemming from modelling errors, disturbances, lacking information on the structural system's configuration, and noise corruption. However, they are limited by the requirement for a model structure, which should be representative of the system's dynamics. 

In the general case, the equation of motion of a multi degree of freedom linear time invariant (LTI) dynamic system can be formulated as:
\begin{equation}
    \label{eq:eqmotion}
    \mathbf{M} \ddot{\mathbf{u}}(t)+\mathbf{D} \dot{\mathbf{u}}(t)+\mathbf{K}(t)=\mathbf{S}_i\mathbf{f}(t)
\end{equation}

where $\mathbf{u}(t) \in \mathbb{R}^{n_{dof}}$ is the vector of displacements, often linked to the Degrees of Freedom (DOFs) of a numerical system model, $\mathbf{M} \in \mathbb{R}^{n_{dof}\times n_{dof}}$, $\mathbf{D} \in \mathbb{R}^{n_{dof}\times n_{dof}}$ and $\mathbf{K} \in \mathbb{R}^{n_{dof}\times n_{dof}}$ denote the mass, damping and stiffness matrices respectively;  $\mathbf{f}(t)\in\mathbb{R}^{n_i}$ (with $n_i$ representing the number of loads) is the input vector and $\mathbf{S}_{i}\in\mathbb{R}^{n_{dof}\times n_{i}}$ is a Boolean input shape matrix for load assignment. 
As an optional step, a Reduced Order Model (ROM) can be adopted, often derived via superposition of modal contributions $\mathbf{u} (t)\: \approx \:  \mathbf{\Psi p}(t)$, where $\mathbf{\Psi}\in\mathbb{R}^{n_{dof}\times{n_r}}$ is the reduction basis and  $\mathbf{p}\in\mathbb{R}^{n_r}$ is the vector of the generalised coordinates of the system, with $n_r$ denoting the reduced system dimension. This allows to rewrite \autoref{eq:eqmotion} as: 
\begin{equation}
    \label{eq:redeqmotion}
    \mathbf{M}_r \ddot{\mathbf{p}}(t)+\mathbf{D}_r\dot{\mathbf{p}}(t)+\mathbf{K}_r \mathbf{p}(t)=\mathbf{S}_r \mathbf{f}(t)
\end{equation}
where the mass, damping, stiffness and input shape matrices of the reduced system are obtained as $\mathbf{M}_r = \mathbf{\Psi}^{T}\mathbf{M\Psi}$, $\mathbf{D}_r=\mathbf{\Psi}^{T}\mathbf{D\Psi}$, $\mathbf{K}_r = \mathbf{\Psi}^{T}\mathbf{K\Psi}$ and $\mathbf{S}_r=\mathbf{\Psi}^{T}\mathbf{S}_i$. 

Assuming availability of response measurements, $\mathbf{x}_{k} \in \mathbb{R}^m$, at a finite set of $m$ DOFs, such an LTI system can be eventually brought into a combined deterministic-stochastic state-space model, which forms the basis of application of Bayesian filtering schemes \citep{vettori2023adaptive}:
	\begin{equation} 
		\label{eq:BDM}
		\begin{cases}
		\mathbf{z}_{k} = \mathbf{A}_{d}\mathbf{z}_{k-1}+\mathbf{B}_d\mathbf{f}_{k-1}+\mathbf{w}_{k-1} \\ 
        \mathbf{x}_k = \mathbf{C} \mathbf{z}_k+\mathbf{G}\mathbf{f}_k+\mathbf{v}_k.
		\end{cases}	
	\end{equation}

where the state vector $c = \begin{bmatrix} {\mathbf{p}_k}^T & {\mathbf{\dot{p}}_k}^T \end{bmatrix}^T \in \mathbb{R}^{2n_{r}}$ reflects a random variable following a Gaussian distribution with mean $\mathbf{\hat{z}}_k \in \mathbb{R}^{2n_r}$ and covariance matrix $\mathbf{P}_k \in \mathbb{R}^{2n_r \times 2n_r}$. Stationary zero-mean uncorrelated white noise sources $\mathbf{w}_k$ and $\mathbf{v}_k$ of respective covariance $\mathbf{Q}_k:\mathbf{w}_{k}\sim{\mathcal {N}}\left(0, \mathbf{Q}_k\right)$ and  $\mathbf{R}_k:\mathbf{v}_{k}\sim{\mathcal {N}}\left(0, \mathbf{R}_k\right)$ are introduced to account for model uncertainties and measurement noise. A common issue in BF schemes lies in calibrating the defining noise covariance parameters, which is often tackled via offline schemes \citep{ODELSON2006303}, or online variants, as those proposed recently for more involved inference tasks \citep{yang2020structure,vettori2023adaptive,doi:10.1061/AJRUA6.0000839}. %

Bayesian filters exploit this hybrid formulation to extract an improved posterior estimate of the complete response of the system $\mathbf{z}_k$, i.e. even in unmeasured DOFs, on the basis of a ``predict" and ``update" procedure. Variants of these filters are formed to operate on linear (Kalman Filter - KF) or nonlinear systems (Extended KF - EKF, Unscented KF - UKF, Particle Filter - PF, etc) for diverse estimation tasks. Moreover, depending on the level of reduction achieved, BF estimators can feasibly operate in real, or near real-time. It becomes, however, obvious that these estimators are restricted by the rather strictly imposed model form. 

\begin{figure}[h!]
    \centering
    % \begin{subfigure}{0.57\textwidth}
    %     \centering
    %     \includegraphics[width=\textwidth]{figures/BF_States.png}
    %     \caption{}
    % \end{subfigure}
    % \begin{subfigure}{0.41\textwidth}
    %     \centering
    %     \includegraphics[width=\textwidth]{figures/BF_Parameters.png}
    %     \caption{}
    % \end{subfigure}
    \includegraphics[width=0.99\textwidth]{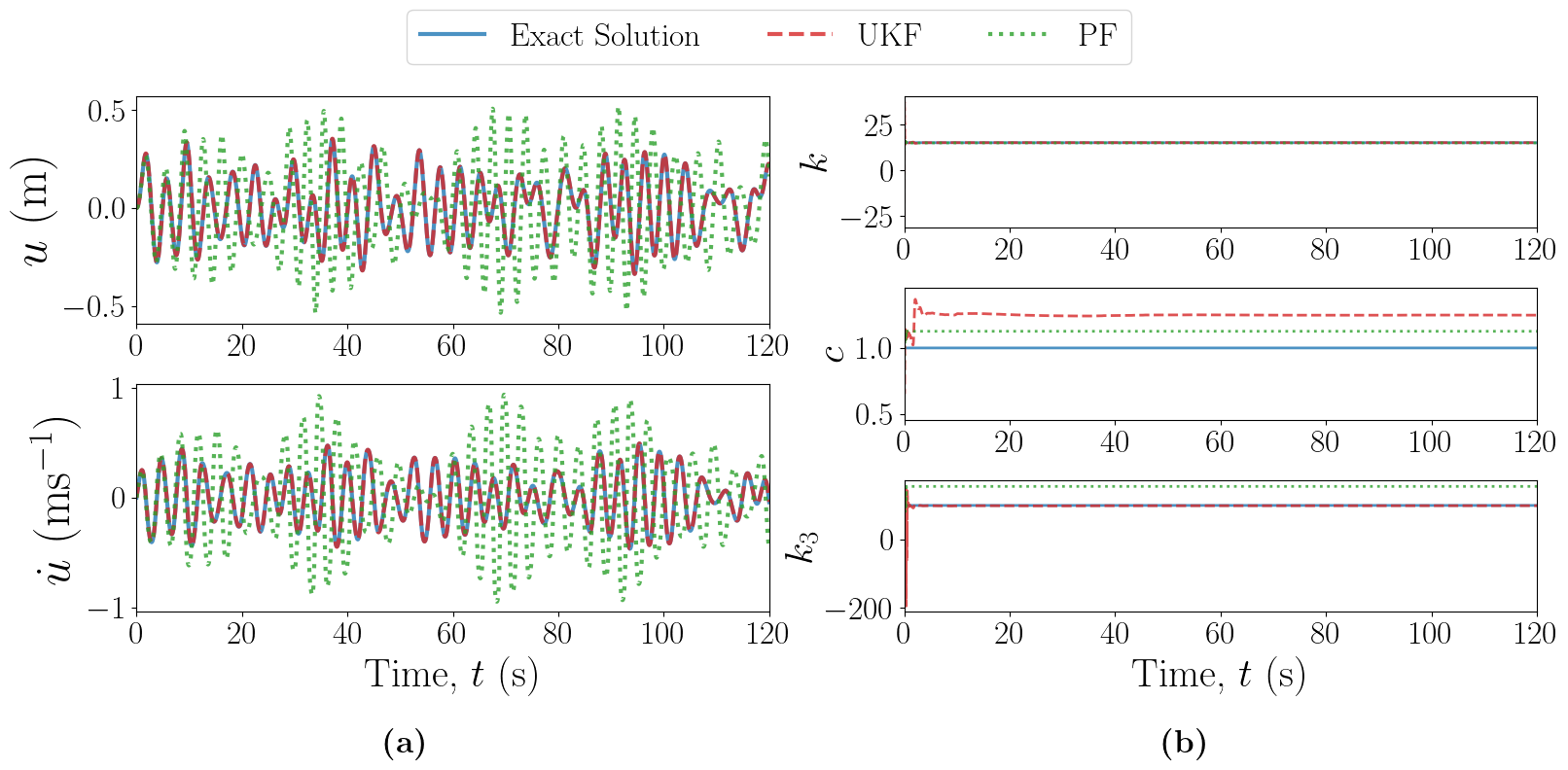}
    \caption{(a) State (response) estimation results for the nonlinear SDOF working example, assuming availability of acceleration measurements and precise knowledge of the model form, albeit under the assumption of unknown model parameters. The performance is illustrated for use of the UKF and PF, contrasted against the reference simulation; (b) Parameter estimation convergence via use of the UKF and PF contrasted against the reference values for the nonlinear SDOF working example}
    \label{fig:BF_estimation}
\end{figure}

In order to exemplify the functionality of such Bayesian filters for the purpose of system identification and state (response) prediction, we present application of two nonlinear variants of the Kalman Filter on our Duffing oscillator working example. %
The system is simulated using the model parameters and inputs defined in \Cref{sec:working_examp}. %
We further assume that the system is monitored via use of a typical vibration sensor, namely an accelerometer, which delivers a noisy measurement of $\ddot{u}$. %
We further contaminate the simulated acceleration with zero mean Gaussian noise corresponding to 8.5\% Root Mean Square (RMS) noise to signal ratio. %
For the purpose of this simulation, we assume accurate knowledge of the model form describing the dynamics, on the basis of engineering intuition. %
However, we assume that the model parameters are unknown, or rather uncertain. %
The UKF and PF are adopted in order to identify the unknown system parameters, namely the linear stiffness $k$, mass, $m$, and nonlinear stiffness $k_3$. %
The parameter identification is achieved via augmenting the state vector to include the time invariant parameters. %
A random walk assumption is made on the evolution of the parameters. %
The UKF employs a further augmentation of the state to include 2 dimensions for the process and measurement noise sources, resulting in this case in $2*9+1=19$ Sigma points to simulate the system. %
It further initiates from an initial guess on the unknown parameters, set as: $k_0=1$ \si{N/m}, $c_0=0.5$ \si{Ns/m}, $k_{30}=40$ \si{N/m^3}, which is significantly off with respect to the true parameters. %
The PF typically employs a larger number of sample points in an effort to more appropriately approximate the posterior distribution of the sate. %
We here employ 2000 sample points and initiate the parameter space in the interval $k \in \{5 \; 20\}$ \si{N/m}, $c \in \{0.5 \; 2\}$ \si{Ns/m}, $k_3 \in \{50 \; 160\}$ \si{N/m^3}. %
In all cases, a zero mean Gaussian process noise of covariance $1e-18$ (added to the velocity states) and a zero mean Gaussian measurement noise of covariance $1e-18$ is assumed. %
Figure \ref{fig:BF_estimation} demonstrates the results of the filter for the purpose of state estimation (left subplot) and parameter estimation (right subplot). %
The plotted result reveals a closer matching of the states for the UKF, while both filters sufficiently approximate the unknown parameters. %
More details on implementation of these filters is found in \citep{Chatzi2009,CHATZI2010,Kamariotis_2023}, while a Python library is made available in association to the following tutorial on Nonlinear Bayesian filtering \citep{Tatsis_2023}. %

Recent advances/applications of Bayesian filtering in structural mechanics include the following works. %
The problem of virtual sensing has been further explored in \citep{TATSIS2021107223,TATSIS2022108558} adopting a sub-structuring formulation, which allows to tackle problems, where only a portion of the domain is monitored.
For clarity, sub-structuring involves dividing a complex domain into smaller, more manageable components, which are solved independently before integrated back into the full structure. %
Employing a lower-level of reliance on the physics model form, by embedding physical concepts in the form of physics-domain knowledge, \citep{tchemodanova2021strain} proposed a novel approach, where they combined a modal expansion with an augmented Kalman filter for output-only virtual sensing of vibration measurements. %
\cite{gres2021kalman} proposed a Kalman filter-based approach to perform subspace identification on output-only data, where in the input force is unmeasured. %
In this case, only the \emph{periodic nature} of the input force is known, and so this (unparameterised) information is also embedded within the model learning architecture. %
In comparison to filtering techniques with an assumed known force, this approach is less reliant on the physics model prescription, and so this approach has the advantage that it may be applied to a wide variety of similar problems/instances. %
The problem of unknown  inputs has recently led to the adoption of Gaussian Process Latent Force Models (GPLFMs), which move beyond the typical assumption of a random walk model, that are meant to describe the evolution of the input depending on the problem at hand \citep{NAYEK2019497,vettori2023assessment,ZOU2023110488,ROGERS2020106580}. Such an approach now moves toward a grey-like method (as discussed in the later sections), since Gaussian Processes, which are trained on sample data are required for data-driven inference and characterization of the unknown input model.

In relaxing the strictness of the imposed physics model, BF inference schemes can include model parameters in the inference task. Such an example is delivered in joint or dual state-parameter estimation methods \citep{dertimanis2019input,TEYMOURI2023109758}, which are further extended to state-input-parameter estimation schemes. %
% For simultaneous state-parameter estimation, Bayesian methods are often used in order to allow for discrepancies between the prescribed model and the data, via inclusion of stochastic interpretations of the model. %
In this context, \cite{naets2015online} couple reduced-order modelling with Extended Kalman Filters to achieve online state-input-parameter estimation, while \cite{dertimanis2019input} combine a dual and an Unscented Kalman filter, to this end; the former for estimating the unknown structural excitation, and the latter for the combined state-parameter estimation. 
Naturally, when the inference task targets multiple quantities, it is important to ensure sufficiency of the available observations, a task which can be achieved by checking appropriate observability identifiability, and invertibility criteria \citep{SHI2022108345,MAES2019378,Chatzis2015}.
\cite{feng2020force} proposed a `sparse Kalman filter', using Bayesian logic, to effectively localise and reconstruct time-domain force signals on a fixed beam. %
As another example in the context of damage detection strategies, \cite{nandakumar2021structural} presented a method for identifying cracks in a structure, from the state space model, using a combined Observer Kalman filter identification, and Eigen Realisation Algorithm methods. %
Another approach to overcome to challenge of model-system discrepancy is to utilise ML approaches along with BF techniques to `bridge the gap', but more will be discussed on this in \Cref{sec:guided}, as these are no longer white-box models. %

%----------------------------------------------------------------
% SECTION 2:
%----------------------------------------------------------------
\section{The Black Box Case - Deep Learning Models}
\label{sec:black_box}

Many modern ML methods are based on, or form extensions of, perhaps one of the most well-known methods, the neural network (NN). %
The NN can be used as a universal function approximator, where more complex models will generally require deeper and/or wider networks. %
When using multiple layers within the network, the method falls in the \emph{deep-learning} (DL) class. %
For a regression problem, the aim of an NN is to determine an estimate of the mapping from the input $\mathbf{x}$, to the output $\mathbf{y}$. %
A fully-connected, feed-forward NN is formed by $N$ hidden layers, each with $n^{(N)}$ nodes. %
The nodes of each layer are connected to every node in the next layer and the values are passed through an activation function $\sigma$. %
For $N$ hidden layers, the output of the neural network can be defined as,
\begin{equation}
    \mathcal{N}_{\mathbf{y}}(\mathbf{x};\mathbf{W}, \mathbf{B}) := \sigma(\mathbf{w}^l x^{l-1} + \mathbf{b}^l), \quad \mathrm{for}\; l = 2,...,N
\end{equation}
where $\mathbf{W}=\{\mathbf{w}^1,...,\mathbf{w}^N\}$ and $\mathbf{B}=\{\mathbf{b}^1,...,\mathbf{b}^N\}$ are the weights and biases of the network, respectively. %
The aim of the training stage is to then determine the network parameters $\mathbf{\Theta} = \{\mathbf{W},\mathbf{B}\}$, which is done by minimising an objective function defined so that when the value vanishes, the solution is satisfied. %
\begin{equation}
    L_o = \left\langle \mathbf{y}^* - \mathcal{N}_{\mathbf{y}} \right\rangle _{\Omega_o}, \qquad
    \langle \bullet \rangle _{\Omega_{\kappa}} = \frac{1}{N_{\kappa}}\sum_{x\in\Omega_{\kappa}}||\bullet||^2
    \label{eq:obs_loss}
\end{equation}

At the other end of the spectrum of a \emph{white-box} (model-based) approach, where the system dynamics is transparent and therefore largely prescribed, thus lies a \emph{black-box} approach, employing naive DL schemes to achieve stochastic representations of monitored systems. Linking to the BF structure described previously, Variational Autoencoders (VAE) have been extended with a temporal transition process on the latent space dynamics in order to infer dynamic models from sequential observation data \citep{bayer2015learning}. %
This approach offers greater flexibility than a scheme that relies on a prescribed physics-based model form, since VAEs are more apt to learning arbitrary nonlinear dynamics. The obvious shortcoming is that, typically, the inferred latent space need not be linked to coordinates of physical connotation. This renders such schemes more suitable for inferring dynamical features, and even condition these on operational variables \citep{MylonasCVAE}, but largely unsuitable for reproducing system response in a virtual sensing context.
Following such a scheme, Stochastic Recurrent Networks (STORN) \citep{bayer2015learning} and Deep Markov Models (DMMs) \citep{krishnan_structured_2016}, which are further referred to as Dynamic Variational Autoencoders (DVAEs), have been applied for inferring dynamics in a black box context with promising results in speech analysis, music synthesis, medical diagnosis and dynamics \citep{Vlachas2022}. In structural dynamics, in particular, previous work of the authoring team \cite{simpson2021machine} argues that use of the AutoeEncoder (AE) essentially leads in capturing a system's Nonlinear Normal Modes (NNMs), with a better approximation achieved when a VAE is employed \citep{Simpson2021}. It is reminded that, while potent in delivering compressed representations, these DL methods do not learn interpretable latent spaces. 

\begin{figure}[h]
    \centering
    \includegraphics[width=\textwidth]{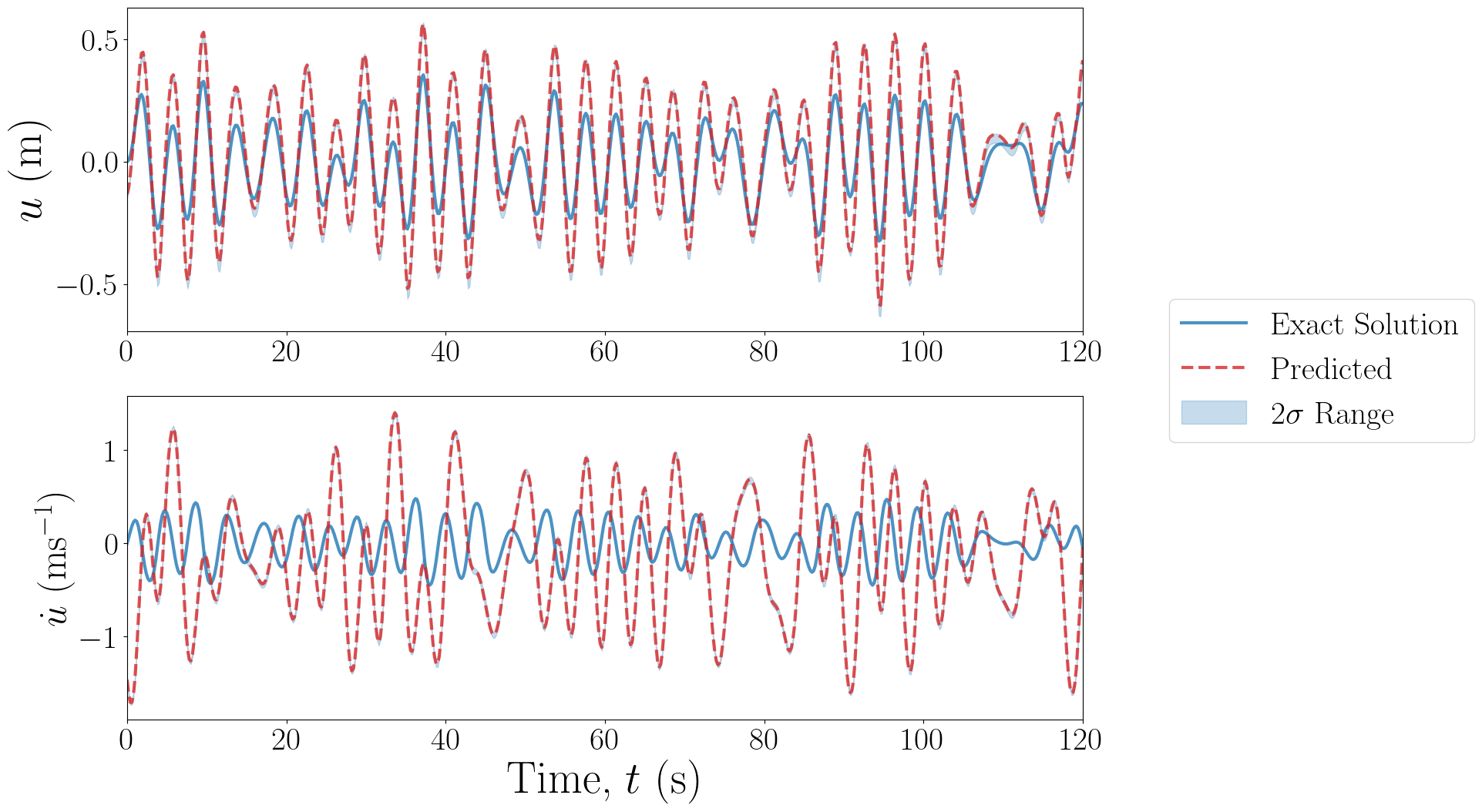}
    \caption{Predicted latent representations vs exact solutions of displacement (top) and velocity (bottom) using the DMM applied to the working example. Displacement is assume to be the only measurement. The blue bounding boxes represent the estimated $2\sigma$ range}
    \label{fig:dmm_duffing}
\end{figure}

In this paper, a rudimentary black-box method is demonstrated to provide a simple example of ML applied to the case scenario. %
Only one black-box approach is shown here to keep the focus overall to \emph{PEML} techniques. %
The Figure \ref{fig:dmm_duffing} shows the results of applying DMM to the working example. %
The $2\sigma$ uncertainty is also included, however in this case it is difficult to observe on the figure, as the uncertainty is small as a result of the low level of noise within the data. %
The data is generated for the time interval of 0 to 120 seconds with a sampling rate of 5Hz and the displacement is assumed to be the only measurement. %
All the transition and observation models, as described by \cite{krishnan_structured_2016}, are modelled by black-box neural networks, specifically DMMs. %
While it is observed that the latent representation captures certain patterns of observed data, it lacks physical interpretability. %

% ------------------------------------------------------------------------------------
\section{Light Grey PEML Schemes}
\label{sec:guided}

When the prior physics knowledge of a system is relatively well-described, i.e.\ it captures most of the physics of the true system, it is possible to rely on this knowledge as a relatively strong bias, while further exploiting learning schemes to capture any model mismatch. %
The term model mismatch or model-system discrepancy refers to the portion of the true system's behavior (or response) which remains uncaught by the known physics. %
As a result of the larger degree of reliance on the physics-based model form, we here refer to this class of methods as "light-grey". %
We will first discuss a survey of machine-learning-enhanced Bayesian filtering methods, which are still mostly driven by physics knowledge embedded in the BF technique. %
This is then followed by a section on Physics-Guided Neural Networks, which use the universal-approximation capabilities of deep-learning to determine a model of model-system discrepancy. %

\subsection{ML-Enhanced Bayesian Filtering}

As previously stated, classical Bayesian filtering requires the model form to be known a priori, implying that the resulting accuracy will depend on how exhaustive this model is. %
To overcome the inaccuracies that result from model-system discrepancy, ML can be infused with BF techniques to improve inference potential. %
To this end, \cite{tatsis2022hierarchical} propose to fuse BF with a Covariance Matrix Adaptation scheme, to extract the unknown position and location of flaws in the inverse problem setting of crack identification, while simultaneously achieving virtual sensing. The latter is the outcome of a hierarchical BF approach powered by reduced order modelling.

Using a different approach, \cite{revach2022kalmannet} employ a neural-network within a Kalman filter scheme to discover the full form of \emph{partially} known and observed dynamics. %
By exploiting the nonlinear estimation capabilities of the NN, they managed to overcome the challenges of model constraint, that are common in filtering methods \citep{aucejo2019practical}. %
Using a similar approach, but with a different motivation, \cite{angeli2021deep} combined Kalman filtering with a deep-learning architecture to perform model-order reduction, by learning the mapping from the full-system coordinates, to a minimal coordinate latent space. %

%Another such light grey approach was presented by \cite{figueiredo2019finite}, where a Gaussian mixture model (GMM) was combined with a finite-element scheme to improve modelling of a bridge in varying E/O conditions from sparse data. % In their work, the FE model is calibrated from observed data, which are then used to generate a less-sparse data which is then passed to the GMM. % Traditionally, Bayesian approaches require reasonably dense data, however, this combined approach presents a potential solution to overcome this challenge. %

\subsection{Physics-Guided Neural Networks}

% In recent years, the emergence of machine learning techniques, particularly deep learning, has revolutionized various fields by providing powerful tools to analyze complex data and make accurate predictions. However, in many real-world applications, relying solely on data-driven models may not be sufficient, especially when the underlying systems are governed by known physical principles. In such cases, incorporating prior knowledge about the physics of the monitored system can significantly enhance the performance and interpretability of the machine learning models. This integration of physical knowledge with deep learning methods is known as Physics-Guided Machine Learning (PGML).

% Physics-guided machine learning (PGML) is a powerful approach that leverages prior knowledge about the underlying physics of a monitored system to enhance the training of machine learning models. %
% Scientific knowledge can be used in the design of neural network architecture to restrict the space of models to physically consistent solutions, or in initializing the model with physically meaningful parameters. %
% Unlike physics-informed methods, where physical laws are imposed as constraints on the model, physics-guided machine learning employs prior knowledge as a guiding force during the training process. %
% This prior knowledge may not be entirely accurate, as it represents an initial understanding of the system's dynamics. %

In \emph{physics-guided} machine learning (PGML), deep learning techniques are employed to capture the discrepancy between an explicitly defined model based on prior knowledge and the true system from which data is attained. %
The goal is to fine-tune the overall model's parameters (i.e.\ the prior and ML model) in a way that the physical prior knowledge steers the training process toward the desired direction. %
By doing so, the model can be guided to learn latent quantities that align with the known physical principles of the system. %
This ensures that the resulting model is not only accurate in its predictions but also possesses physically interpretable latent representations. %
At this stage, we would like to remind the reader of the definitions of \emph{physics model strictness} and \emph{physics constraint} given in the \Cref{sec:introduction}. %
\emph{PGML} approaches employ a relatively high reliance on the physics-based model form, in that the assumed physics is imposed in a strict form. However, in order to allow for simulation of model discrepancy, the level of physics constraint is relatively low, which means that the learner is not forced to narrowly follow this assumed prescribed form. %
The relaxation of constraints differs between \emph{PGML} and physics-informed (PIML) approaches; \emph{PGML} techniques often reduce constraints through the use of bias or residual modelling, whereas \emph{PIML} schemes employ physics into the loss function as a \emph{target} solution (i.e.\ the solutions are weakly imposed). %
% However, the knowledge that is embedded tends to be more restricted models, as otherwise the overall level of knowledge would be too minimal to have influence. %

One of the key advantages of \emph{physics-guided} machine learning is its ability to incorporate domain knowledge into the learning process. %
This is particularly beneficial in scenarios where data may be limited, noisy, or expensive to obtain. %
In comparison to \emph{PEML} techniques that lay lower on our prescribed spectrum, \emph{PGML} methods can offer an increased level of interpretability. %
By steering the model's learning process with physics-based insights, it can more effectively generalize to unseen data and maintain a coherent understanding of the underlying physical mechanisms. %
% In order for models to be interpretable, they require a decent level of specificity in terms of their model definitions, which means that PGML-based methods have a relatively large level of physics restriction. %

% The fundamental idea behind \emph{physics-guided} machine learning is to use the known theoretical principles, empirical laws, or even expert insights of the system being studied as a prior during the model training. %
% These prior knowledge guide the learning process towards physically meaningful solutions and prevent the model from diverging into unrealistic regions of the solution space. %
% In essence, the physical prior knowledge serves as a compass, steering the model towards capturing the true dynamics of the system. %
% However, it is essential to acknowledge that this prior knowledge might not be completely accurate due to modeling assumptions, simplifications, or uncertainties in the physical laws themselves. %
% At the heart of the PGML framework is the construction of a physics-guided model that respects the underlying physical principles while being flexible enough to capture discrepancies between the prior knowledge and the true dynamics of the system. %
% Deep learning architectures, such as neural networks, offer a versatile toolset for this task due to their ability to approximate complex functions and learn intricate patterns from data. %

The training process in \emph{physics-guided} models involves two key components:
1) Incorporating Prior Knowledge: Prior knowledge on the physics of the system is integrated into the network architecture, or as part of the model; 2) Capturing Discrepancy: Deep learning models excel in learning from data, even when this contradicts prior knowledge. 
As a result, during training, such a model will gradually adapt and learn to account for discrepancies between the prior knowledge and the true dynamics of the system. %
This adaptability allows the model to converge towards a more accurate representation of the underlying physics. %

Conceptually similar to estimating a residual modeller, \cite{liu2022physics,liu2021physics} proposed a probabilistic \emph{physics-guided} framework termed a Physics-guided Deep Markov Model (PgDMM) for inferring the characteristics and latent structure of nonlinear dynamical systems from measurement data. It addresses the shortcoming of black-box deep generative models (such as the DMM) in terms of lacking physical interpretation and failing to recover a structured representation of the learned latent space. To overcome this, the framework combines physics-based models of the partially known physics with a DMM scheme, resulting in a hybrid modelling approach. The proposed framework leverages the expressive power of DL while imposing physics-driven restrictions on the latent space, through structured transition and emission functions, to enhance performance and generalize predictive capabilities. The authors demonstrate the benefits of this fusion through improved performance on simulation examples and experimental case studies of nonlinear systems.

Both residual modelling and \emph{PgNNs}, in general, share a common objective of easing the training objective of neural networks. Residual modelling achieves this by easing the learning process through a general approximation, while \emph{PgNNs} incorporate physical knowledge to provide reliable predictions even in data-limited situations. However, regarding specific implementations, there are differences between these two methods. Residual modelling operates without the need for specific domain knowledge, focusing instead on the residuals in a more general-purpose application. This makes it broadly applicable across various standard machine learning tasks, especially in the domain of computer vision. \emph{PgNNs}, on the other hand, explicitly incorporate physical laws into the model, adding interpretability related to these physical models. \emph{PgNNs} can also work in a general setting with any prior models to fit the residuals, but the key idea is to use physical prior model to obtain a physically interpretable latent representation from neural networks. This makes \emph{PgNNs} particularly suited for problems where adherence to physical laws is paramount. Additionally, while residual networks rely heavily on data for learning, \emph{PgNNs} can leverage physical models to make predictions even with limited data, demonstrating their utility in data-constrained environments. 

\begin{figure}[h]
    \centering
    \includegraphics[width=\textwidth]{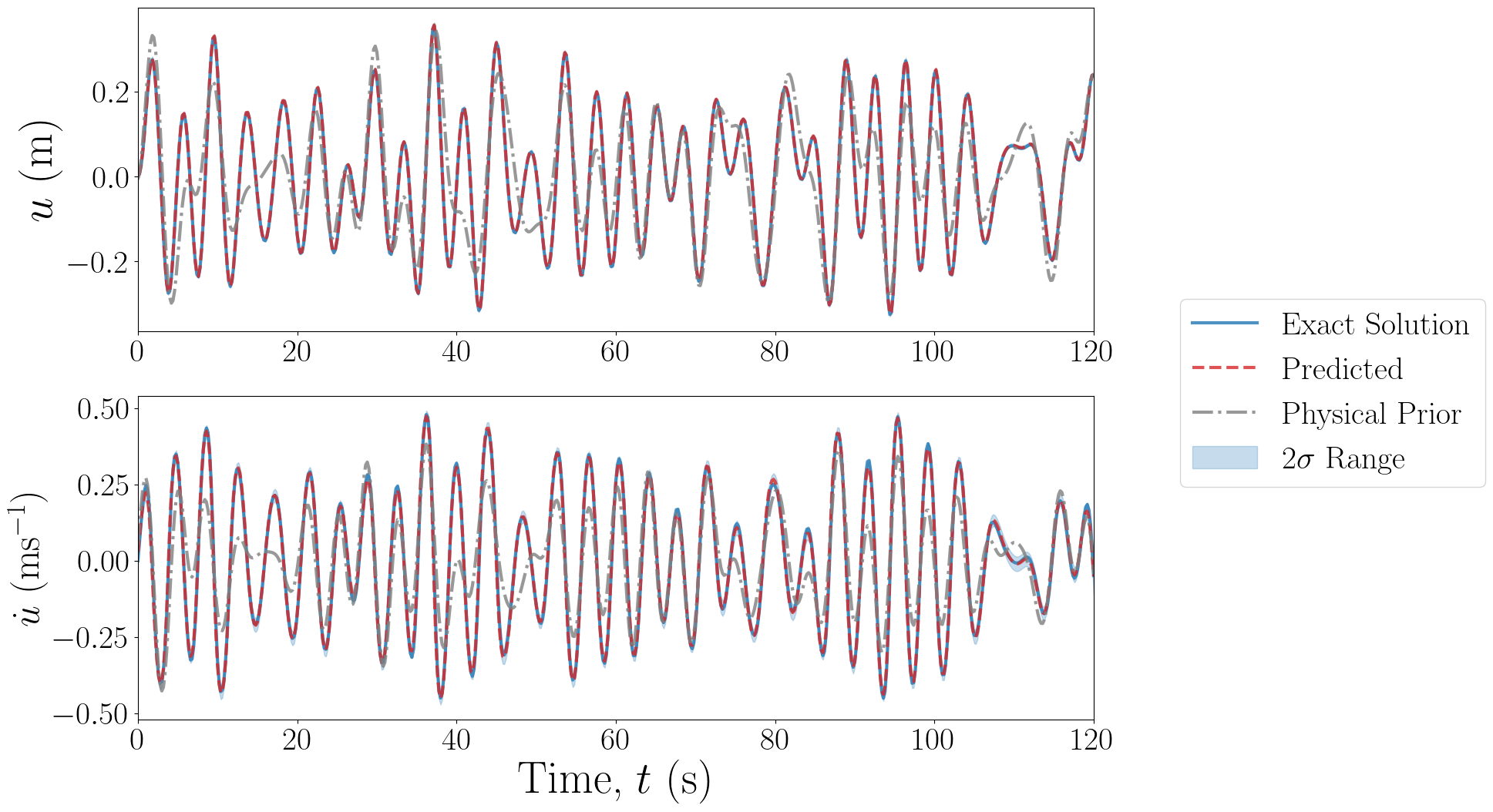}
    \caption{Predictions vs exact solutions of displacement (top) and velocity (bottom) using the PgDMM applied to the working example. Displacement is assume to be the only measurement. The gray dash-dot line is the physical prior model and the blue bounding boxes represent the estimated $2\sigma$ range}
    \label{fig:pgdmm_duffing}
\end{figure}

To demonstrate how a physical prior model can impact the training of DL models, we apply the PgDMM to our working example, using the same data generation settings as explained in Section \ref{sec:black_box}. In this case, instead of approaching the system with no prior information, we introduce a physical prior model into the DMM to guide the training process. This physical prior model is a linear model that excludes the cubic term in Equation \eqref{eq:duffing_osc}, which replicates a knowledge gap in the form of additional system complexities. The results are shown in Figure \ref{fig:pgdmm_duffing}. It can be observed that the predictions for both displacements and velocities align well with the ground-truth. The estimation uncertainty is slightly higher for velocities, which is expected since they are unobserved quantities. It's important to note that the system displays significant nonlinearity due to the presence of a cubic term with a large coefficient, causing the linear approximation to deviate noticeably from the true system dynamics. However, the learning-based model within the framework still captures this discrepancy and reconstructs the underlying dynamics through the guided training process.

A similar \emph{physics-guided} RNN was proposed by \cite{yu2020structural}, which consists of two parts: physics-based layers and data-driven layers, where physics-based layers encode the underlying physics into the network and the residual block computes a residual value which reflects the consistency of the prediction results with the known physics and needs to be optimized towards zero.

Instead of modelling the residual of the prior model, a physics-guided Neural Network (PgNN), proposed by \cite{karpatne2017physics}, not only ingests the output of a physics-based model in the neural network framework, but also uses a novel physics-based learning objective to ensure the learning of physically consistent predictions, as based on domain knowledge. Similarly, the authors proposed a Physics-guided Recurrent Neural Network scheme (PGRNN, \citep{jia2019physics}) that contains two parallel recurrent structures - a standard RNN flow and an energy flow to be able to capture the variation of energy balance over time. While the standard RNN flow models the temporal dependencies that better fit observed data, the energy flow aims to regularize the temporal progression of the model in a physically consistent fashion. Furthermore, in another \emph{PgNN} proposed by \cite{robinson2022physics}, the information from the known part of the system is injected into an intermediate layer of the neural network.

The physics-guided deep neural network (PGDNN), proposed by \cite{huang2022physics}, uses a cross-physics-data domain loss function to fuse features extracted from both the physical domain and data domain, which evaluates the discrepancy between the output of a FE model and the measured signals from the real structure. With the physical guidance of the FE model, the learned PGDNN model can be well generalized to identify test data of unknown damages. The authors also use the same idea in bridge damage identification under moving vehicle loads \citep{yin2023bridge}. Similarly, \cite{zhang2021structural} presented usage of the FE model as an implicit representation of scientific knowledge underlying the monitored structure and incorporates the output of FE model updating into the NN model setup and learning.

In \cite{chen2021probabilistic}, the physics knowledge is incorporated into the neural network by means of imposing appropriate constraints on weights, biases or both. \cite{muralidhar2020phynet} used physics-based prior model, physical intermediate variables and \emph{physics-guided} loss functions to learn physically interpretable quantities such as pressure field and velocity field.

% ------------------------------------------------------------------------------------
\section{Grey PEML Schemes}
\label{sec:informed}

The motivations which drive the `middle ground' of \emph{PEML} will - naturally - vary depending on the nature of the information deficiency, and whether this is a physics knowledge gap or data scarcity. %
For example, the knowledge could be made up of a number of possible physical phenomena, or the system could be known but the parameters not. %
Or, with the opposite problem, it may be possible to capture data well in one domain, but be limited in the relative resolution of other domains (e.g. temporal vs.\ spatial). %
These are just a few of the many examples which motivate the use of ``grey'' \emph{PEML} techniques. %
In this section, two such approaches are surveyed and discussed; first dictionary methods are shown, which select a suitably-sparse representation of the model via linear superposition from a dictionary of candidate functions. %
The second technique discussed is the physics-informed neural network, which weakly imposes conditions on the model \emph{output} in order to steer the learner. %
This differs from the previously discussed physics-guided approaches, where the physics restrictions are strongly imposed by way of a proposed solution that instantiates an inductive bias to the learner. %
As will be discussed in detail below, the latter technique can be applied in a variety of ways, each embedding different prior knowledge and beliefs, allowing for flexibility in its application. %

\subsection{Dictionary Methods}

One of the biggest challenges faced in the practical application of structural mechanics in engineering, is the presence of irregular, unknown, or ill-defined nonlinearities in the system. %
Another challenge may occur from variation in the parameters which govern the prescribed model of the system, which could be from environmental changes, or from consequential changes such as damage. %
This motivates less reliance on the physics-based model form, to allow for freedom in physical-digital system discrepancy whilst satisfying known physics. %
One approach to reducing the reliance on the prescribed physics model form is by having the learner estimate the definition of the model, which may be the sole, or additional, objective of the learner. %
Dictionary methods are well-positioned as a solution for less strict physics embedding, where the model is determined from a \emph{set} of possible model solutions, allowing freedom in a semi-discrete manner. %

The problem of estimating the existence, type, or strength of the model-governing physics is described as that of \emph{equation discovery}. %
This inverse problem can often be very computationally expensive, due to the large number of forward model calculations required to evaluate the current estimate of the parameters \citep{frangos2010surrogate}. %
When determining the presence of governing equation components, often the identification is drawn from a \emph{family} of estimated equations. %
For example, the matching-pursuit algorithm selects the most-sparse representation of a signal from a dictionary of physics-based functions \citep{vincent2002kernel}. %

For dictionary-based approaches the idea is to determine an estimate of the model output as some combination of bases or `atoms'. %
Often, these bases are formed as candidate functions of the input data, and are compiled into a dictionary-matrix $\mathbf{\Theta}(\mathbf{x})$. %
Then, linear algebra is used to represent the target signal from a sparse representation of this dictionary and a coefficient matrix $\Xi$. %
\begin{equation}
    \mathbf{y} = \mathbf{\Theta}(\mathbf{x})\Xi
\end{equation}
And so, the aim of the learner is to determine a suitably-sparse solution of $\Xi$, via some objective function. %
Typically, sparsity promoting optimisation methods are used, such as LASSO \citep{tibshirani1996regression}. %
The coefficient matrix can be constrained to be binary values, thus operating as a simple mask of the candidate function, or can be allowed to contain continuous values, and thus can simultaneously determine estimated system parameters which are used in the candidate functions. %
For the case of dynamic systems, a specific algorithm was developed by \cite{brunton2016discovering} for sparse discovery of nonlinear systems. %
The team also showed how this could be used in control by including the control parameters in the dictionary definition \citep{brunton2016sparse}. %
\cite{kaiser2018sparse} extended this nonlinear dictionary learning approach to improve control of a dynamic system where data is lacking. %
To do so, they extended the method to include the effects of actuation for better forward prediction. %

In an example of dictionary-based learning, \cite{flaschel2021unsupervised,thakolkaran2022nn} showed a method for unsupervised learning of the constitutive laws governing an isotropic or anisotropic plate. %
The approach is not only unsupervised, needing only displacement and force data, it is directly inferrable in a physical manner. %
The authoring team then extended this work further to include a Bayesian estimation \citep{joshi2022bayesian}, allowing for quantified uncertainty in the model of the constitutive laws. %

Another practical example: \cite{ren2023data} used nonlinear dynamic identification to successfully predict the forward behaviour of a 6DOF ship model, including coupling effects between the rigid body and water. %
In this work, the dictionary method does is combined with a numerical method to predict the state of the system in a short time window ahead. %
This facet is often found when applying DM-based approaches to practical examples, as the method is intrinsically a model discovery approach, and so a solution step is required if model output prediction is required. %

Data-driven approaches to equation discovery will often require an assumption of the physics models being exhaustive of the `true' solution; i.e.\ all the parameters being estimated will fully define the model, or the solution will lie within the proposed family of equations. %
One of the challenges faced with deterministic methods, such as LASSO \citep{tibshirani1996regression}, is their sensitivity to hyperparameters \citep{brunton2016discovering}; a potential manifestation of which is the estimation of a combination of two similar models, which is a less accurate estimate of the `true' solution than each of these models individually. %
Bayesian approaches can help to overcome this issue, by instead providing a stochastic estimate of the model, and enforcing sparsity \citep{park2008bayesian}. %

\cite{fuentes2021equation} show a Bayesian approach for nonlinear dynamic system identification which simultaneously selects the model, and estimates the parameters of the model. %
Similarly, \cite{nayek2021spike} identify types and strengths of nonlinearities by utilising spike-and-slab priors in the identification scheme. %
As these priors are analytically intractable, this allowed them to be used along with a MCMC sampling procedure to generate posterior distributions over the parameters. %
\cite{abdessalem2018model} showed a method for approximate Bayesian computation of model selection and parameter estimation of dynamic structures, for cases where the likelihood is either intractable or cannot be approached in closed form. %

\subsection{Physics-Informed Neural Networks}
\cite{raissi2019physics} showed that by exploiting automatic differentiation that is common in practical implementation of neural networks, one can embed physics that are known in the form of ordinary/partial differential equations (ODEs/PDEs). %
Given a system of known ODEs/PDEs which define the physics, where the sum of these ODEs/PDEs should equal zero, an objective function is formed which can be estimated using automatic differentiation over the network. %
If one was to apply a \emph{PINN} to estimate the state of the example in \Cref{fig:duffing_osc}, over the collocation domain $\Omega_c$, using \Cref{eq:duffing_state_space}, the physics-informed loss function becomes,
\begin{equation}
    L_{p}(\mathbf{t};\mathbf{\Theta}) = \left\langle \partial_t\mathcal{N}_{\mathbf{z}} - \mathbf{A}\mathcal{N}_{\mathbf{z}} - \mathbf{A}_n\mathcal{N}_{u^3} - \mathbf{B}f \right\rangle _{\Omega_c}
    \label{eq:pde_loss}
\end{equation}
where $\partial_t\mathcal{N}_{\mathbf{z}}$ is the estimated first order derivative of the state using automatic differentiation, and the system parameters are $\theta = \{m, c, k, k_3\}$. %
Boundary conditions are embedded into \emph{PINNs} in one of two ways; the first is to embed them in a `soft' manner, where an additional loss term is included based on a defined boundary condition $\xi(\mathbf{y})$ \citep{sun2020surrogate}. %
Given the boundary domain $\partial\Omega\in\Omega_c$,
\begin{equation}
    L_{bc} = \langle \xi^* - \xi(\mathcal{N}_{\mathbf{y}}) \rangle _{\partial\Omega}
\end{equation}
The second case, so called `hard' boundary conditions, involves directly masking the outputs of the network with the known boundary conditions,
\begin{equation}
    \mathcal{N}_{\mathbf{y}} = \xi^* |_{\partial\Omega}
\end{equation}
where often gradated masks are used to avoid asymptotic gradients in the optimisation \citep{sun2020surrogate}. %
In the case of estimating the state over a specified time window, the boundary condition becomes the initial condition (i.e.\ the state at $t=0$). %
\begin{equation}
    \xi = \mathbf{z}(0) = \{u(0),\dot{u}(0)\}^T, \qquad L_{bc} = \left\langle \xi - \mathcal{N}_{\mathbf{z}} \right\rangle _{\Omega\in t=0}
    \label{eq:ic_definition}
\end{equation}

The total objective function of the \emph{PINN} is then formed as the weighted sum of the observation, physics, and boundary condition losses,
\begin{equation}
    L = \lambda_oL_o + \lambda_pL_p + \lambda_{bc}L_{bc}
\end{equation}
The flexibility of the \emph{PINN} can then be highlighted by considering how the belief of the architecture is changed when selecting the corresponding objective weighting parameters $\lambda_k$. %
By considering the weighting parameters as 1 or 0, \Cref{tab:PINN_app_types} shows how the selection of objective terms to include changes the application type of the \emph{PINN}. %
The column titled $\Omega_c$ indicates the domain used for the PDE objective term, where the significant difference is when the observation domain is used, and therefore the \emph{PINN} becomes akin to a system identification problem solution. %

\begin{table}[h!]
    \centering
    \begin{tabular}{c|c|c|c|l||l}
        $\lambda_{obs}$  &  $\lambda_{pde}$  & $\lambda_{bc}$ & $\Omega_c$ & $\theta$ & PE-ML Subcategory\\
        \hline
        1 & 0 & 0 & $\Omega_o$ & - & Purely data-driven \\
        1 & 1 & 0/1 & $\Omega_o$ & Unknown & System identification \\
        1 & 1 & 1 & $\Omega_c$ & Known & Physics-informed learner \\
        0 & 1 & 1 & $\Omega_c$ & Known & Forward-modeller \\
    \end{tabular}
    \caption{Summary of PINN application types, and the \emph{physics-enhanced} machine learning genre/category each would be grouped into.}
    \label{tab:PINN_app_types}
\end{table}

\begin{figure}[h!]
    \centering
    \includegraphics[width=\textwidth]{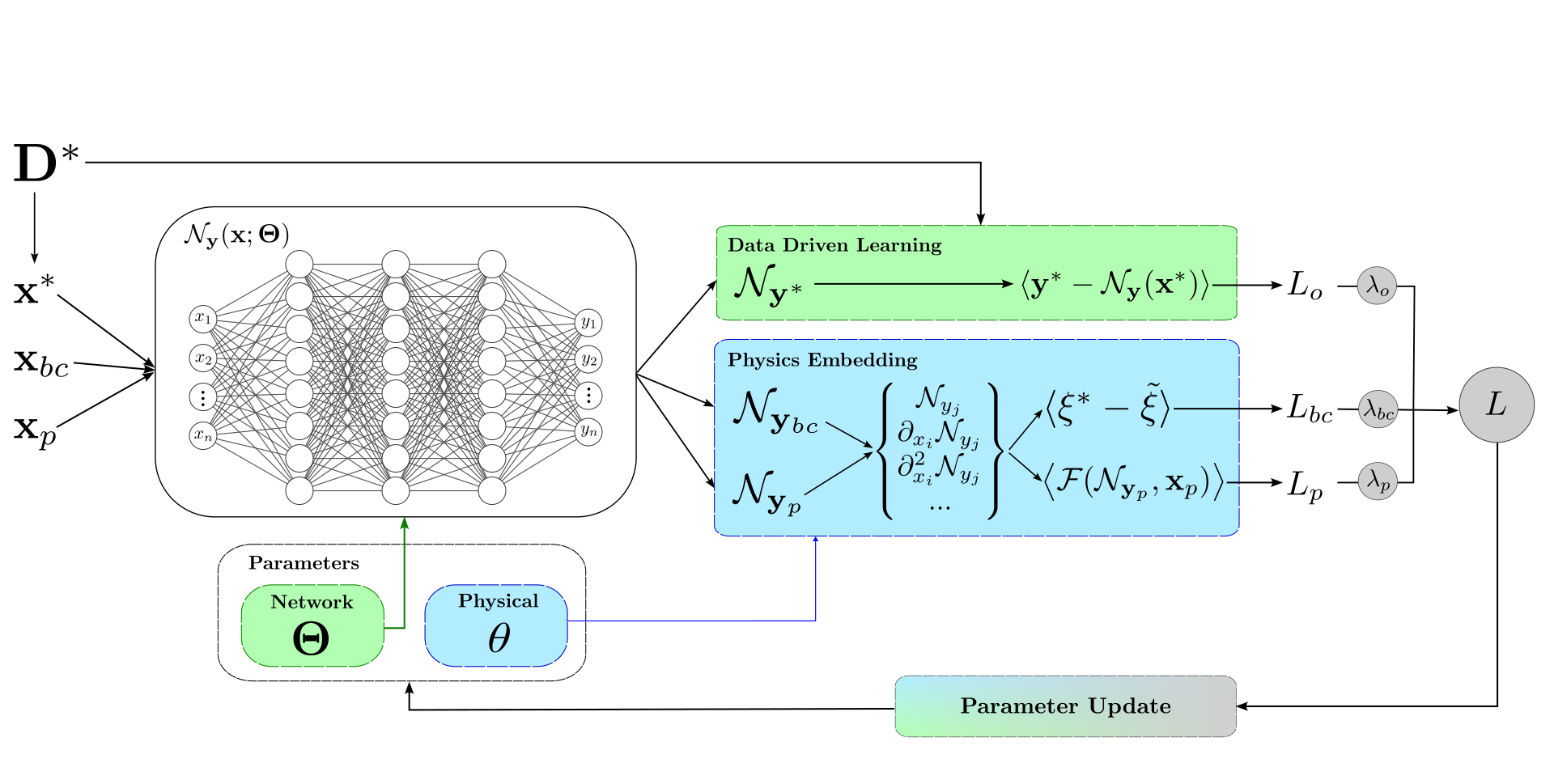}
    \caption{Framework of a general \emph{PINN}, highlighting where the data-driven and physics-knowledge are embedded within the process}
    \label{fig:pinn_framework}
\end{figure}

\Cref{fig:pinn_framework} shows the framework of a generic \emph{PINN}, highlighting where the framework, specifically the loss function formulation, can be broken down into the data-driven and physics-embedding components. %
It is also possible to include the system parameters $\theta$ as unknowns which are determined as part of the optimisation process, i.e.\ $\mathbf{\Theta} = \{\mathbf{W},\mathbf{B},\theta\}$. %
Doing so adds the capability as an system identification tool, as well as equation solution discovery. %
In system identification, either equation discovery can be performed, where the aim is to determine the definition of $\mathcal{F}$, or parameter estimation, where the aim is to determine the values of $\theta$. %
The framework shown is for the training stage of the model, where the network parameters $\mathbf{\Theta}$, and optionally the physical parameters $\theta$, are updated using an optimisation algorithm such as LBFGS \citep{liu1989limited} or Adam \citep{kingma2014adam}. %
Further prior physical knowledge can be added to \emph{PINNs} by embedding more objective functions, such as initial conditions or continuity conditions, which can be formed directly from the output values of the network, or from the derivatives. %

To illustrate the different approaches to implementing a \emph{PINN}, three paradigms of the method are discussed here, each to aiming to tackle a different objective. %
\begin{enumerate}
    \item \emph{System Identification}; the aim is to determine the physics description of the system, either by equation discovery or by parameter estimation the system parameters $\theta$ in \Cref{eq:pde_loss} via estimating the system state
    \item \emph{Enhanced Learning}; the aim is to enhance the learner to either better estimate the model given a more sparse set of data, or to improve learning efficiency
    \item \emph{Forward Modelling}; where the \emph{PINN} acts to generate a `simulated' model of the system, given the system equations and parameters, and the domain of interest
\end{enumerate}
These points will form the remaining parts of this subsection, where surveys will be shown and an example of each approach applied to the working example of the Duffing oscillator. %
The \emph{PINN} is applied here as an `instance-modeller'; where the estimated model is only applicable in the case of the prescribed initial conditions and forcing signal. %
Therefore, new estimations made with this model are only applicable within the training domain. %
The generalisability and extrapolability presented with the \emph{PINN} is in terms of extending beyond the domain of observations. %

Physics-informed neural networks are gaining increased attention thanks to some of the advantages stated above, however, there exist a number of drawbacks which should be noted. %
Firstly, in many formulations of \emph{PINNs}, they suffer from a lack of generalisability - which is a common motivation for \emph{PEML} schemes - as they are restricted to the domain on which they were trained \citep{haghighat2021physics}. %
This domain may be extended beyond that of observations, however, computationally-intensive training must still be performed before prediction. %
Furthermore, there is a lack of intuition or knowledge on the optimisation task of \emph{PINNs}; often, weighting of the losses is done via trial and improvement \citep{wang2021understanding}, and an under-defined, or ill-posed, physics prescription may result in many local minima \citep{nandi2021progress}. %
A related challenge is in the computational effort of training, which can be a considerable task. %
This challenge has garnered fair criticism against the use of \emph{PINNs} as opposed to other numerical solving schemes \citep{grossmann2023can}. %

\subsubsection{PINNs for System Identification}

As discussed above, \emph{PINNs} can be used as an approach to system identification, by applying a `soft' condition on the governing physical laws. %
The use of a variety of soft conditions, in the form of the loss functions, allows for discrepancies between the model and data, making them useful for noisy observations. %
A good demonstration of \emph{PINNs} for simultaneous state-system estimation is shown by \cite{yuan2020machine}, and \cite{moradi2023novel}, where in both examples, they model the displacement of a vibrating beam, with accurate estimation of the governing equation parameters also. %
\cite{zhang2020physics} use a \emph{PINN} to solve an identification problem for nonhomogeneous materials, using elasticity imaging. %
An elegant approach here was done so by utilising a multi-network architecture to include nonhomogeneous parameter fields, removing the potential issue caused by exploding dimensionality when including the spatially-dependent material parameters. %
\cite{sun2023pisl} recently showed an approach to discovering the parameters of the complex nonlinear dynamic systems using a Physics-informed Spline Learning approach. %
This approach employs the same exploitation as \emph{PINNs}, however, the splines are used to allow for differentiation from a more sparse set of observed data, by interpolating the underlying dynamics. %

Instead of determining the governing parameters of the PDEs, \cite{wu2020data} proposed a method to estimate the forward-in-time solution to a model. %
By recovering the evolution operator in the modal space, this reduces the problem from an infinite-dimension to a finite-dimension space. %
As opposed to value-based parameter identification, \cite{ritto2021digital} used measurement from a vibrating bar to update a ML-based classifier which directly infers the damage state of the structure. %
As the governing equations are often in the form of PDEs, an estimation of the derivatives is required in the learning process of many equation discovery approaches. %
However, \cite{goyal2022discovery} proposed a numerical integration framework with dictionary learning, along with a `Runge-Kutta inspired' numerical procedure, overcoming the issues presented with derivative approximation from corrupted or sparse data. %

For the working example, the \emph{PINN} was applied to the Duffing oscillator, shown in \Cref{fig:duffing_osc}, and an equivalent linear system (i.e.\ $k_3=0$). %
Here, 256 data points were sub-sampled using a Sobol sampler and passed to the learner, and the physics-loss domain is set to the same as the observation domain (i.e.\ $\Omega_p=\Omega_o$), positing the framework akin to a system identification scheme. %
The observation loss (\Cref{eq:obs_loss}) penalises the output enough to result in accurate estimation of the displacement, and satisfying the physics loss will drive the estimation of the physical parameters. %

\begin{figure}[h!]
    \centering
    \includegraphics[width=1.0\textwidth]{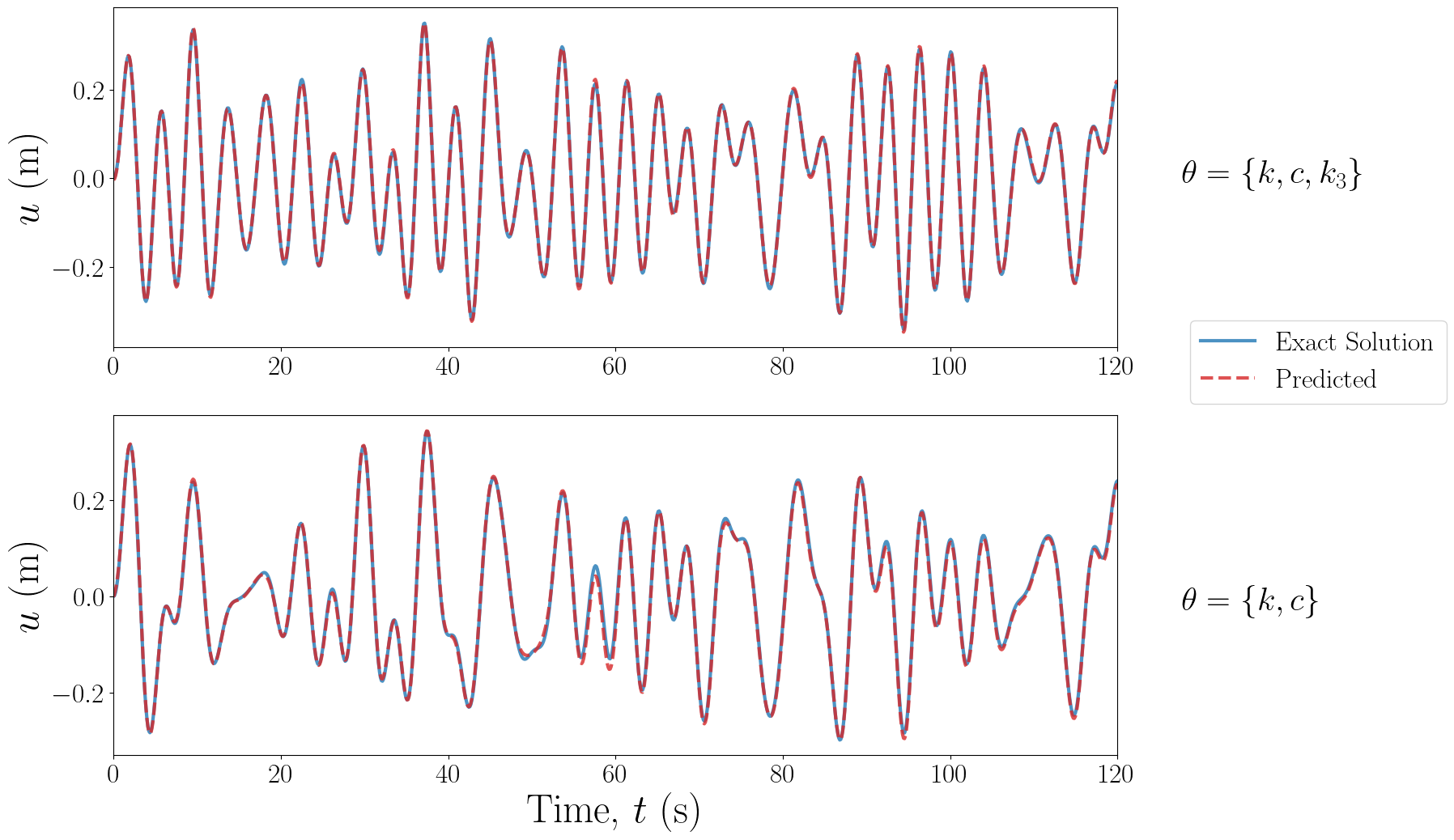}
    \caption{Predicted vs exact solution of simultaneous system-state estimation approach to solving the working example for the nonlinear case (top) and linear case ($k_3=0$) (bottom)}
    \label{fig:ed_pinn_duffing}
\end{figure}

The results of the state-estimation are shown in \Cref{fig:ed_pinn_duffing}, and the results of the parameter-estimation are shown in \Cref{tab:ed_pinn_results}. %
The state estimation results show accurate modelling, and the estimated values for the physical parameters also show a good level of accuracy. %
It is important to note here that accurate estimation of the state is not the primary objective, given that the domain of interest is well-covered by the observation data. %
A notable result is the increased accuracy when modelling only the linear system. %
This result can be explained by well known machine learning intuition that with an increased dimensionality of estimation-space, an increased number of information is also required. %
Therefore, with the same \emph{level} of information provided, the accuracy of the results for estimating more physical parameters will likely be more of a challenge for the learner. %

\begin{table}[h!]
    \centering
    \begin{tabular}{r|c|c|c|c|c|c}
         & \multicolumn{2}{c|}{$c$} & \multicolumn{2}{c|}{$k$} & \multicolumn{2}{c}{$k_3$} \\
        % \cline{2-7}
        \hline
        System Case & Predicted & Error & Predicted & Error & Predicted & Error \\
        \hline
        Nonlinear & 1.002 & 0.26\% & 15.02 & 0.12\% & 99.22 & 0.78\% \\
        Linear & 1.002 & 0.213\% & 15.00 & 0.011\% & - & - \\
    \end{tabular}
    \caption{Results of system-estimation for the SDOF oscillator for both the nonlinear and linear case. For all parameters, the estimated value and the percentage error are shown.}
    \label{tab:ed_pinn_results}
\end{table}

\subsubsection{PINNs for Enhanced Learning}

Another utility of \emph{PINNs} is domain enhancement potential, either by improving domain density from sparse observations, or by extending the domain beyond that of the observation domain. %
Practically for spatio-temporal models, increasing density is often only motivated in the spatial domain, as a result of the ease of improving time-domain sampling. %
However, domain-extension approaches are found to be practically motivated in both space and time. %

\cite{xu2022physics} used \emph{PINNs} to accurately model the rigid-body dynamics of an unmanned surface vehicle as it voyages along a river. %
This is a nice example of a practical implementation of improving state prediction from sparse data, and a good example of how to formulate \emph{PINNs} for relatively complex descriptions of dynamics. %
\cite{chen2021probabilistic} use a \emph{PINN} for estimating the fatigue $S-N$ curves, where even on seemingly-sufficient data, the uninformed ANN fails to accurately predict. %
A particular note of this work, was the inclusion of a probabilistic framework, allowing both freedom in the model construction, as a result of stochastic considerations, and a quantified estimate of the uncertainty. %
By utilising a finite-element model, in which the parameters were updated using a \emph{PINN}, \cite{zhang2021structural} developed a NN-based method for detecting damage. %

By utilising an energy-based formulation of the loss function, \cite{zhuang2021deep} modelled bending, vibration and buckling of a Kirchoff plate. As well as the informed loss function, the authors used a non-standard activation function to better emulate the underlying operations which govern the physics. %
Another example of an energy-based loss function approach is shown by \cite{goswami2020transfer}, who combined a \emph{PINN} with transfer learning to model the phase-field of fracture in a material. %
They showed how a well-trained model could drastically reduce the computational requirements of this problem. %

The deformation of elastic plates with \emph{PINNs} was shown by \cite{li2021physics}, where they made a comparison between purely data-driven, PDE-based, and energy-based physics informing, finding each to have different advantages. %
The PDE-based approach was less dependent on sampling size and resolution, whereas the energy-based approach had less hyperparameter, and therefore was more efficient and easier to train. %
This information may be useful for future development of \emph{PINNs}, which suffer from a lack of understanding of the hyperparameter space pathology, to allow for robust optimisation strategies, as discussed by \cite{wang2021understanding}. %

\cite{yucesan2020physics} present a methodology for predicting the damage level in a wind turbine blade, in the form of grease degradation, using \emph{PINNs}. %
In their work, the advantage of \emph{PINNs} is that instead of aiming to directly model the value of grease degradation, they aim to predict the \emph{increment} of grease degradation based on the current value and a number of other measurable quantities. %
This would be a difficult task to perform in a black-box manner, as the values of the increment are not the observed values, and so the physics embedding helps to overcome this. %

\begin{figure}[h!]
    \centering
    \includegraphics[width=1.0\textwidth]{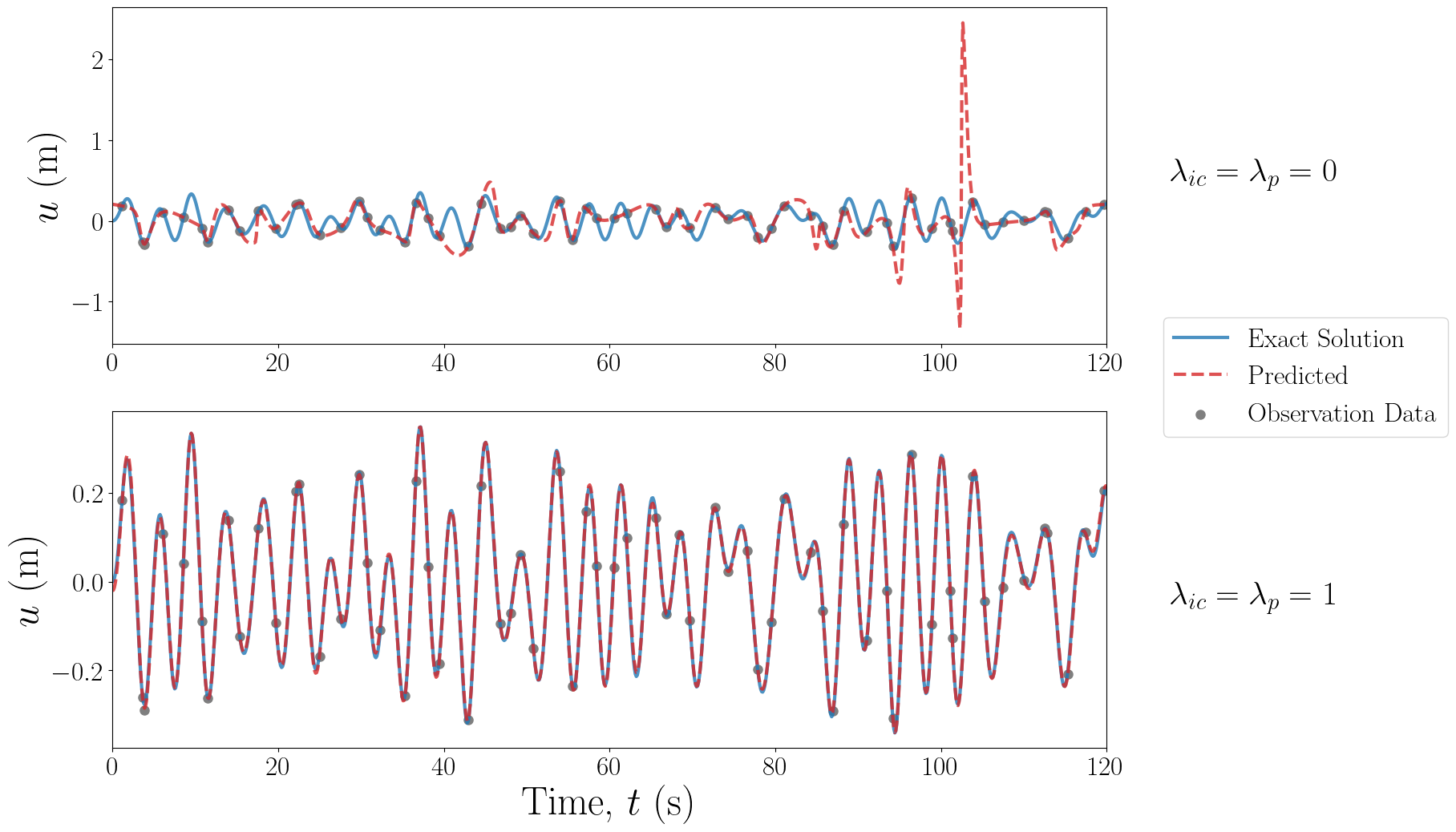}
    \caption{Predicted vs exact solution of state estimation approach applied to a subsample of the working example with no physics embedded (top) and physics-informed embedding (bottom)}
    \label{fig:pg_pinn_duffing}
\end{figure}

Now, we will show the \emph{PINN} applied as an enhanced learner to the working example, where the physical parameters are known \emph{a priori}. %
The advantages of the \emph{PINN} in this aspect is more efficient learning, and for improved extrapolation of the data from sparse observations. %
To demonstrate, only every sixteenth sample is instead fed to the learner, emulating a sampling frequency of 0.5328\\si{Hz}. %
Then, the \emph{PINN} is applied as a `black-box' architecture (i.e.\ $\lambda_{ic}=\lambda_p=0$), and as a physics-informed modeller, where all loss weight parameters are included. %
The results of the two methods applied are shown in \Cref{fig:pg_pinn_duffing}, where it is clear to see the strength of embedding physics, demonstrated in the context of sparse data. %

\subsubsection{PINNs for Forward Modelling}

When applying \emph{PINNs} for forward modelling of systems and structures, no prior observations are given, i.e.\ $D^*=\emptyset$, however, it is necessary to provide sufficient boundary and initial conditions, and physics constraints which describe a complete model. %
An advantage of \emph{PINNs} for forward modelling is their simple implementation; embedding boundary conditions, complex geometries, or new governing equations is relatively straightforward. %
The training time of \emph{PINNs} as forward modellers, in relation to traditional finite-element methods, is often greater, leading to statements implying their impracticality \citep{rezaei2022mixed}. %
In the context of forward modelling of a known model, and at pre-set collocation points, this statement of impracticality is well-founded. %
However, \emph{PINNs} may also provide a convenient solution for problems such as; in-time control/prediction, where a \emph{PINN} could predict an output rapidly, as the computational effort is done \emph{a priori} during training, or for determining a better generalisation over high-dimensional modelling space, where the computational cost of FEM methods can rapidly increase with dimensionality. %

\emph{PINNs} for forward modelling has gained a lot of traction in micro-scale problems; \cite{haghighat2021physics} and \cite{henkes2022physics} use \emph{PINNs} to model the instance of displacements and stresses in a unit cell. %
In the latter, they showed the capability of the approach to model nonlinear stresses by including a sharp phase transition within the material. %
Haghighat and the team also showed how the method could accurately model structural vibration \citep{haghighat2021deep}, giving only initial and boundary conditions. %
\cite{abueidda2021meshless} showed \emph{PINNs} for modelling various solid mechanic effects; elasticity, hyperelasticity, and plasticity, where the method performed well on all types of materials. %
In work by \cite{zheng2022physics}, fracture mechanics were modelled to a decent accuracy with only principle physics. %

The above examples all follow a fairly straightforward path to the model formulation, by employing the variables that are inherent to the governing equations as the variables of the ML model. %
Going beyond this, \cite{huang2020machine} used a \emph{proper orthogonal decomposition} (POD) neural network to model plasticity in a unit cell. %
The advantage of the POD approach is to decouple the multi-dimensional stress, allowing the use of individual NNs for each stress variable, reducing computation time, and increasing learning efficiency. %
Straying away from the common approach of using PDEs to form the physics-based loss, \cite{abueidda2022deep} formulated an energy-based loss term to successfully forward model hyperelasticity and viscoelasticity in a given material. %

\begin{figure}
    \centering
    \includegraphics[width=0.95\textwidth]{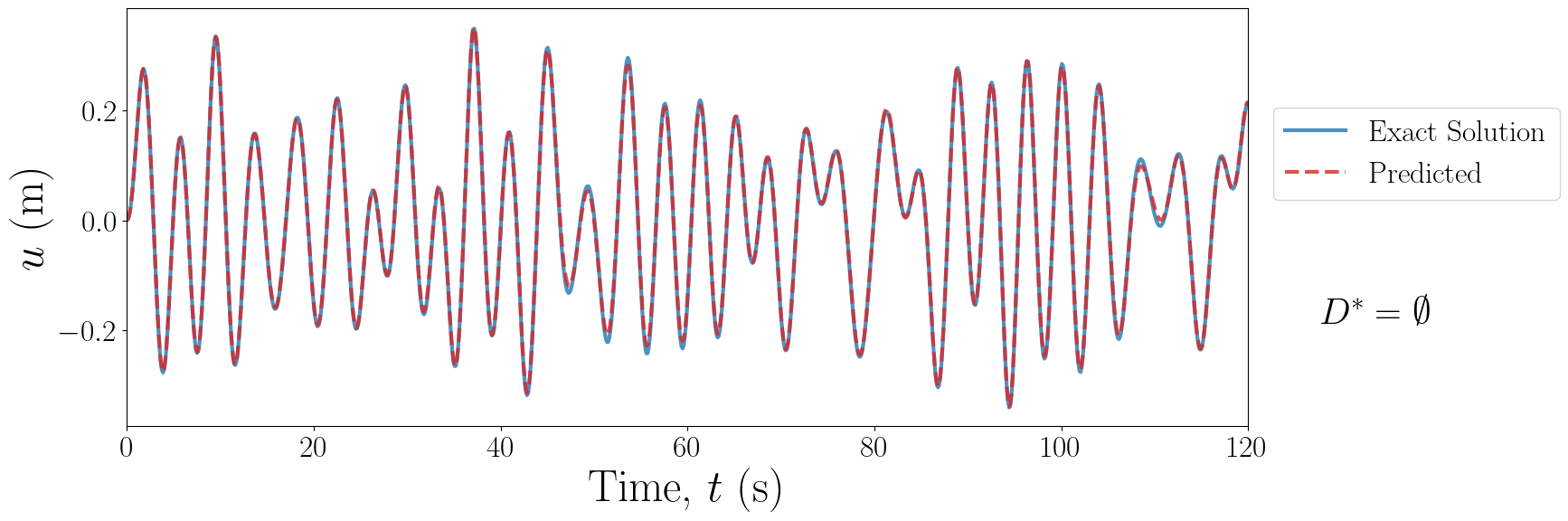}
    \caption{Exact solution vs \emph{PINN}-based forward modelling solutions of the SDOF Duffing oscillator example, where no observations of the state are given to the learner}
    \label{fig:fm_pinn_duffing}
\end{figure}

As a final demonstrator of \emph{PINN} utility, we apply it as a forward modeller, with no observed data offered to the learner, i.e.\ $\Omega_o=\emptyset$. %
However, it is important to note that there is still \emph{training data} reflecting time domain information. 
For this implementation, initial conditions, in the form of Dirichlet and Neumann boundary conditions from \Cref{eq:ic_definition}, and the forcing signal are provided. %
In \Cref{fig:pinn_framework}, therefore, only the physics embedding portion of the framework is included. %
The results of this forward modelling approach are shown in \Cref{fig:fm_pinn_duffing}, where it can be seen the solution matches well with the exact solution. %

% --------------------------------------------------------------------------
\section{Dark Grey PEML schemes}
\label{sec:encoded}

In this final section of the \emph{physics-enhanced} discussion, the `darkest' genres of \emph{PEML} techniques (that are included in this paper) are discussed. %
These techniques have a lower reliance on the physics-based model form, as the prescribed model has a lower level of strictness. 

\subsection{Constrained GPs}
\label{sec:const_gps}

One example of \emph{physics-encoded} learners are constrained Gaussian processes (GPs), which, depending on kernel design, can be viewed as embedding the general shape of the function, as GPs are a problem of discovery over the \emph{function-space}, as opposed to the weight-space \citep{williams2006gaussian, o1978curve}. %
% The Gaussian process (GP) is a flexible Bayesian regression method, which works by placing a prior over functions, which is then updated on the basis of data, to return a posterior distribution over functions \cite{williams2006gaussian, o1978curve}. %
Conceptually, this process can be thought of as estimating a distribution over all the possible functions that could explain the data, as opposed to one ``best fit'' model. %
The aim is to estimate a nonlinear regression model given a set of observed output data $\mathbf{y}$, and observed input data $\mathbf{x}$,
\begin{equation}
    \mathbf{y} = f(\mathbf{x}) + \varepsilon, \qquad \varepsilon \sim \mathcal{N}(0, \sigma_n^2\mathbb{I})
\end{equation}
where $\varepsilon$ is a zero-mean Gaussian white-noise process with variance $\sigma_n^2$. %
A GP is fully defined by its mean and covariance functions,
\begin{equation}
    f(\mathbf{x}) \sim \mathcal{GP}(m(\mathbf{x}), K(\mathbf{x},\mathbf{x}'))
\end{equation}
The mean function $m(\mathbf{x})$ can be any parametric mapping of $\mathbf{x}$, and the covariance function expresses the similarity between two input vectors $\mathbf{x}$ and $\mathbf{x}'$. %
The primary influence of the user, when implementing a GP, is in the choice of the covariance kernel, which is calculated as any other kernel; linear pair-wise distances between points to form a covariance matrix. %
There are a number of popular kernels that are used as standard, each of which embed a different belief as to which \emph{family of functions} the model solution is drawn from. %
There are two primary methods of embedding physical knowledge in GPs; the first of which is to include an initial estimate of the model into the mean function, and so the problem can then be envisioned as determining the solution of the remaining, unknown physics. %

The second approach is to design, or select, the covariance kernels to constrain the shape of the function estimate, and often combinations of kernels can provide varying levels of physical knowledge embedding \citep{cross2021physics}. %
In this case, it is possible to relax the reliance on the physics model form, by for example simply dictating the design of the kernel for the domain on which the physics is expected to operate. %
For example, \cite{padonou2016polar} showed a kernel for predicting on circular domains. %
More strictly-prescribed model forms can be applied by designing kernels to include physical knowledge in the form of partial differential equations, or boundary conditions \citep{solin2020hilbert}. %

An example of embedding prior knowledge via the mean function is shown by \cite{zhang2021gaussian}, where they showed an improved GP model for the deflection of the Tamar bridge \citep{cross2013long} under time-varying environmental conditions, by including an expected linear deflection of the cable due to temperature. %
Data from the Tamar bridge represents a relatively complex problem; a large-scale complicated structure under varying (and non-exhaustive) environmental conditions. %
However, even by applying the simplest method of embedding physics into a GP, the modelling was improved. %
\cite{petersen2022wind} also applied a novel physics-informed GP method to a bridge problem, but with the aim of estimating wind load from acceleration data. %
In their work, they developed a novel infusion of GP latent-force model (GP-LFM) with a Kalman filter-based approach. %
The inclusion of the GP-LFM allowed for characterisation of the \emph{evolution} of the wind-load, and this is enriched with prior physical knowledge in the form of stochastic information on wind-loads taken from wind-tunnel tests. %
This work provides an excellent demonstration of how physics information can be embedded to allow transfer of information from scaled structures. %

\cite{haywood2021structured} used constrained GPs to model physical characteristics of guided-waves in a complex material. %
Guided waves in complex materials are famously difficult to model due to their relatively short wavelength in comparison to the material structure. %
By designing a variety of kernels, they demonstrated the performance of the GP modelling with varying levels of physics information embedded. %
Notably, they showed that by considering the space in which the physics operates, one can already improve the modelling capabilities, even before any physics equations are embedded. %
Continuing on the topic of elastic-waves, \cite{jones2023constraining} applied PI-GPs to localise acoustic source emission in a complex domain. %
They developed constrained GPs further by embedding boundary conditions and the spatial domain of the problem. %
This approach has potential for use in modelling on structures with relatively high geometrical complexity, such as those with lots of joints, or with layered materials. %

As stated above, the GP is unique in that it operates in the \emph{function} space, thus, prior knowledge embedded is on that of the shape of the function. %
This characteristic was utilised by \cite{dardeno2021investigating}, who used weak-form dynamics equations as a mean function within a novel overlapping mixture of Gaussian processes (OMGP) method. %
By constraining the expected shape of the functions, this allows the learner to separate out unsorted data of dynamic structures from within a population. %

\begin{figure}[h]
    \centering
    \includegraphics[width=0.95\textwidth]{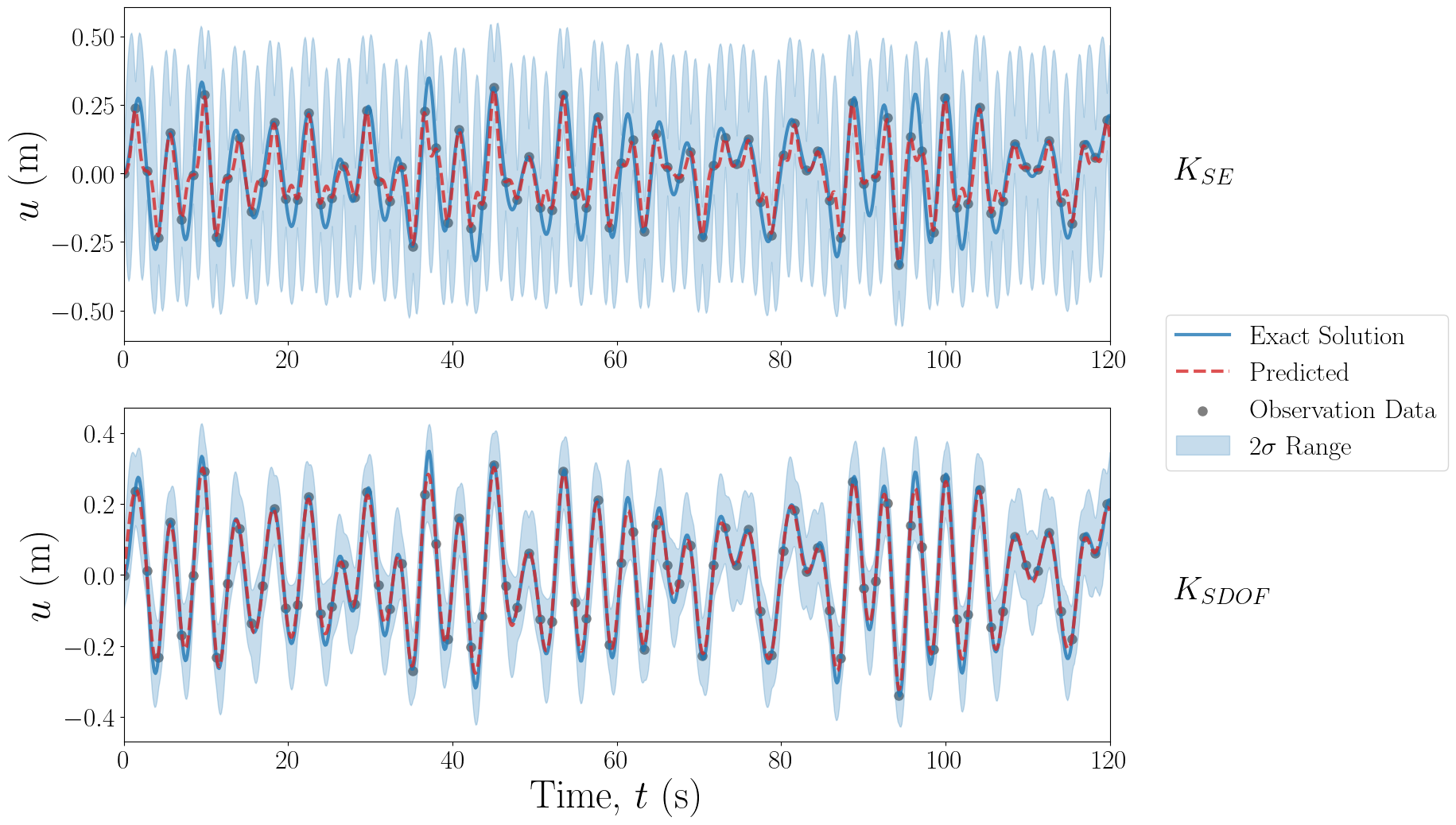}
    \caption{Predicted vs exact solutions of displacement estimation using a GP applied to a subsample of the working example, with (top) no physics embedded and (bottom) constrained GP. The blue bounding boxes represent the estimated $2\sigma$ range}
    \label{fig:cgp_duffing}
\end{figure}

To demonstrate the constrained GP, here were apply this to the working example in \Cref{fig:duffing_osc}. %
In this context, the GP estimates a nonlinear operator where the input is time (i.e. $\mathbf{x}=\mathbf{t}$). %
The first kernel we will select is the scaled squared-exponential kernel, where $l$ is a lengthscale hyperparameter, and $\alpha$ is the scaling hyperparameter,
\begin{equation}
    K_{SE}(\mathbf{t},\mathbf{t}^*) = \exp\left( -\frac{1}{2l^2}(\mathbf{t}-\mathbf{t}^*)^T(\mathbf{t}-\mathbf{t}^*) \right)
    \label{eq:se_kernel}
\end{equation}
which embeds only a belief that the function is smooth. %
Then, following the work of \cite{cross2021physics}, if we assume a Gaussian white noise force input, an additional physics-derived kernel can be included,
\begin{equation}
    K_{SDOF}(\mathbf{t},\mathbf{t}^*) = \frac{\sigma_f^2}{4m^2 \zeta\omega_n^3} e^{-\zeta\omega_n |\tau|} \left(\cos(\omega_d \tau) + \frac{\zeta\omega_n}{\omega_d}\sin(\omega_d|\tau|)\right), \quad \tau=\mathbf{t}-\mathbf{t}^*
    \label{eq:sdof_kernel}
\end{equation}
where $\sigma_f$ is an additional hyperparameter, representing the square root of the variance of the forcing. %
To demonstrate the constrained GP approach, the data from \Cref{fig:duffing_osc} is applied in a similar fashion to the informed-learner approach used with the \emph{PINN} in \Cref{fig:pg_pinn_duffing}. %
Here, only every twelfth sample is fed to the learner, and is first applied with the uninformed squared-exponential kernel in \Cref{eq:se_kernel}, and then with the constrained SDOF kernel in \Cref{eq:sdof_kernel}. %
The results of the estimated solutions using each of these kernels is shown in \Cref{fig:cgp_duffing}, along with the estimated 95\% ($2\sigma$) confidence range. %
Note, the value of $\sigma$ here is taken from the estimated covariance matrix, and is not the hyperparameter $\sigma_f$. %
It is clear to see the improved estimation of the displacement of the system, but on top of this, the estimated uncertainty is also reduced. %
An interesting observations from this is that when using a linear SDOF kernel to model a nonlinear system, this still results in improved modelling from sparse data. %

Constrained Gaussian processes are often used to improve forecasting of temporal data, which falls under the umbrella of domain extension schemes. In the context of the example prescribed here, for accurate forecasting one would need to determine an analytical solution for the covariance of a forced nonlinear system, as with the prescribed kernel, prediction beyond the scope of the data results in simply a free-vibration system. 

\subsection{Physics-Encoded Neural Networks}

So far we have only explored how the automatic differentiation mechanisms can help embed physics into a neural networks. %
This process can be applied blindly to any type of neural network architecture. %
A different approach is to modify the architecture of the neural network itself such that its hidden states conform to a domain closer to that of the physical problem of study. %
By imposing specific geometric constraints, it is possible to bias neural networks towards such domains. %
Adding additional biases to the model reduces the variance during the training phase, which leads to a faster convergence. %
However, such biases must be carefully imposed such that they replicate the physics of the system of study, else they add a constant bias error into the model. %

The biases that one can introduce in a machine learning model are in most cases symmetries that are imposed on the nonlinear mappings of the model \citep{bronstein2021geometric}. %
The rational for imposing symmetries stems from Noether's theorem: every continuous symmetry in a physical system corresponds to a conserved quantity. %
Therefore, symmetries in a model's architecture should be able to encode the physical properties of the system that are conserved. %
From henceforth, we will refer to models as ``\emph{physics-encoded}" whenever their architecture is biased to specifically reproduce the symmetries arising from the properties of the system of study. %
Such symmetries can be tailored depending on the prior knowledge available on the system. %

\subsubsection{Neural ODE}

Neural ordinary differential equations \citep{chen2018neural} are a framework that unifies neural networks and ordinary differential equations to model dynamical systems. %
Unlike traditional neural networks that operate on discrete time steps, Neural ODEs model the evolution of a system continuously over time. %
They leverage the powerful tools of ODE theory to learn and infer hidden states, trajectories, and dynamics from observed data. At the core of Neural ODEs is the use of the adjoint sensitivity method, which enables efficient gradient computation. %
This technique allows gradients to be backpropagated through the ODE solver, enabling end-to-end training of the model using standard gradient-based optimization algorithms. 

Neural ODEs offer several advantages over traditional neural networks. %
Firstly, they provide a flexible and expressive modelling framework that can capture complex temporal dependencies and nonlinear dynamics. %
Secondly, they inherently handle irregularly sampled or sparse data since the ODE solver can handle time interpolation. %
This is particularly valuable when dealing with real-world physical systems where data may be scarce or unevenly sampled. %
Lastly, Neural ODEs can exploit known physical laws or priors by incorporating them into the ODE function, thereby enhancing the interpretability and generalization of the model.

Physics-encoded machine learning using Neural ODEs has found applications in various domains, such as fluid dynamics, particle physics, astronomy, and material science. %
By incorporating physical knowledge into the model architecture, Neural ODEs can leverage the underlying physics to improve predictions, simulate systems, and discover new phenomena. %
Using \emph{PeNN}s in an equation-discovery context, \cite{lai2021structural} utilise physics-informed Neural ODEs for a structural identification problem, where varying levels of prior constraints are embedded in the learner. %
They showcased a framework that allows for an inferrable model, by adopting a sparse identification of nonlinear dynamical systems. % 
Further works by \cite{lai_liu_jian_bacsa_sun_chatzi_2022} show that such a scheme can be integrated into generative models such as VAEs. %
This framework leverages physics-informed Neural ODEs via embedding eigenmodes derived from the linearized portion of a physics-based model to capture spatial relationships between DOFs. %
This approach is notably applied to virtual sensing, i.e., the recovery of generalized response quantities in unmeasured DOFs from spatially sparse data. %

\subsubsection{Hamiltonian Neural Networks}

An alternative method for embedding physics into the architecture of the network is to constraint them according the Hamiltonian formalism. %
The intuition behind this method originates from Noether's first theorem which states that every differentiable symmetry of the action of a physical system with conservative forces has a corresponding conservation law. %
In layman terms, this means that the symmetries that are observed within the dynamics of the system are the result of a conservation of the properties of said system. %

With the success of \emph{PINNs}, more and more theoretical research on the integration of physical principles has been taking place. %
Namely, how do different formulations of a system's equations can be used as prior knowledge to bias our model. %
The most common way in mechanics is to represent a system as its state-space.
An alternative formulation can be done from the point of view of the energy of the system through the Hamiltonian formulation. %
For our state-space $\mathbf{z}$ in the case of a MDOF formed by the pair $(\mathbf{q}, \mathbf{p})$, the Hamiltonian is formulated as: %

\begin{equation}
    H(\mathbf{q}, \mathbf{p}) = \frac{1}{2} \mathbf{p}^{\textrm{T}} \mathbf{M}^{-1}(\mathbf{q}) \mathbf{p} + V(\mathbf{q})
\end{equation} 

With $\mathbf{M}^{-1}(\mathbf{q})\succcurlyeq 0$.
The Hamiltonian is considered to be separable when $H(\mathbf{q}, \mathbf{p}) = T(\mathbf{p}) + V(\mathbf{q})$. %
When looking at the variations of the Hamiltonian over time, one notices that $\dot{\mathbf{q}} = \frac{\partial H}{\partial \mathbf{p}}$ and $\dot{\mathbf{p}} = - \frac{\partial H}{\partial \mathbf{q}}$. %
This leads to the Hamiltonian being time invariant since:

\begin{equation}
    \dot{H} = (\frac{\partial H}{\partial \mathbf{q}})^{\textrm{T}} \dot{\mathbf{q}} + (\frac{\partial H}{\partial \mathbf{p}})^{\textrm{T}} \dot{\mathbf{p}} = 0
\end{equation}

\cite{greydanus2019hamiltonian} is the first to have introduced the Hamiltonian formalism to neural networks to bias the model for physical data. %
The method is closer to a Physics-informed than to a Physics-encoded model since the formalism is added through the loss function. %
The Hamiltonian loss is given by (we remind the reader that $\mathbf{\Theta}$ are the parameters of the model): %

\begin{equation}
    \mathcal{L}_{\textrm{HNN}} = || \frac{\partial H_{\mathbf{\Theta}}}{\partial \mathbf{p}} - \frac{\partial \mathbf{q}}{\partial t}||_2 + || \frac{\partial H_{\mathbf{\Theta}}}{\partial \mathbf{q}} - \frac{\partial \mathbf{p}}{\partial t}||_2
\end{equation}

In this case, is the neural network is learn the gradients of the system. %
Such a formalism prevents to the neural network prediction to stray away from the true state of the system by grounding it in the physical domain. %

This property can also be found in symplectic integrators, i.e., integrators derived from the Hamiltonian formalism.  %
For a given initial value problem, a discrete integration of the system can be performed with explicit integrators such as Euler integration. %
Such integration relies on the local Taylor expansion of the flow of the system. 
For an integration of the n-th order, an integration error of the n+1-th order accumulates at every time step, leading to a drift from the true dynamics of the system. %
This drift can be avoid by opting for symplectic gradients instead. %
The first order symplectic gradients can be derived as follows:

\begin{equation}
    \dot{\mathbf{q}}  = \frac{\partial H}{\partial \mathbf{p}} \Rightarrow \dot{\mathbf{q}}_t = \frac{\mathbf{q}_{k+1} - \mathbf{q}_k}{h} \Rightarrow \mathbf{q}_{k+1} = \mathbf{q}_k + h \frac{\partial H}{\partial \mathbf{p}_k}
\end{equation}

\begin{equation}
    \dot{\mathbf{p}}  = - \frac{\partial H}{\partial \mathbf{q}} \Rightarrow \dot{\mathbf{p}}_t = \frac{\mathbf{p}_{k+1} - \mathbf{p}_k}{h} \Rightarrow \mathbf{p}_{k+1} = \mathbf{p}_k - h \frac{\partial H}{\partial \mathbf{q}_k}
\end{equation}

For $t$ the continuous time, $k$ the discrete time and $h$ the discretisation. %
For the coefficients of symplectic integrators of the n-th order (equivalent to their explicit counterparts), we refer to the formula derived by \cite{yoshidasymplectic1990} ($c_i$ and $d_i$ given in the paper):

\begin{equation}
    e^{h(T(\mathbf{p}) + V(\mathbf{q}))} = \prod_{i=1}^n e^{c_i h T(\mathbf{p})} e^{d_i h V(\mathbf{q})} + \mathcal{O}(h^{n+1})
\end{equation}

The symplectic integration principle has been extended to a plethora of \emph{physics-encoded} models. %
\cite{chen2020symplectic} modified the original HNN by Greydanus by replacing the ANN with an RNN and by updating the gradients in a symplectic manner. %
\cite{saemundsson2020variational} trains a Neural ODE with a split latent space with symplectic integrators. %
\cite{sanchezgonzalez2019hamiltonian} estimates the Hamiltonian of the system, then derives the gradients from its estimate to update the model. %
\cite{david2021symplectic} employs the same principle but replaces the ANN with a Graph Neural Network (GNN). %
One of the issues with the Hamiltonian approach is that it assumes constant energy within the system, something that is usually not the case for most real-world applications. %
Many methods attempt to resolve this issue by incorporating dissipation into their formulation. %
\cite{sosanyadissipative2022}, an adaptation of the original HNN that splits the gradients into their dissipative and non-dissipative components. %
\cite{desaiporthamiltonian2021} adopts the port-Hamiltonian formalism to adopt HNNs, making them apt to learn the dynamics of control systems. %

The first work to extend this notion to DL is that of \cite{greydanus2019hamiltonian}, which enforces a symmetric gradient on a neural network trained to predict the dynamics of a conservative system. %
\cite{saemundsson2020variational} showed that the Hamiltonian formalism could be combined with the previously mentioned Neural ODEs, yielding the so-called Symplectic (state-space area preservation) Neural ODEs. %
\cite{zhong2020symplectic} also used neural ODEs to learn the physics in an inferrable manner, applying Hamiltonian dynamics. %
Particularly, they parameterised the model in order to enforce Hamiltonian mechanics, even when only velocity data can be accessed as opposed to momentum. %
\cite{bacsa2023symplectic} propose a method extending this reasoning to stochastic learning, where a symplectic encoder learns an energy-preserving latent representation of the system, and opens up new considerations for physics-embedded NN architectures. %

\begin{figure}[h]
    \centering
    \includegraphics[width=\textwidth]{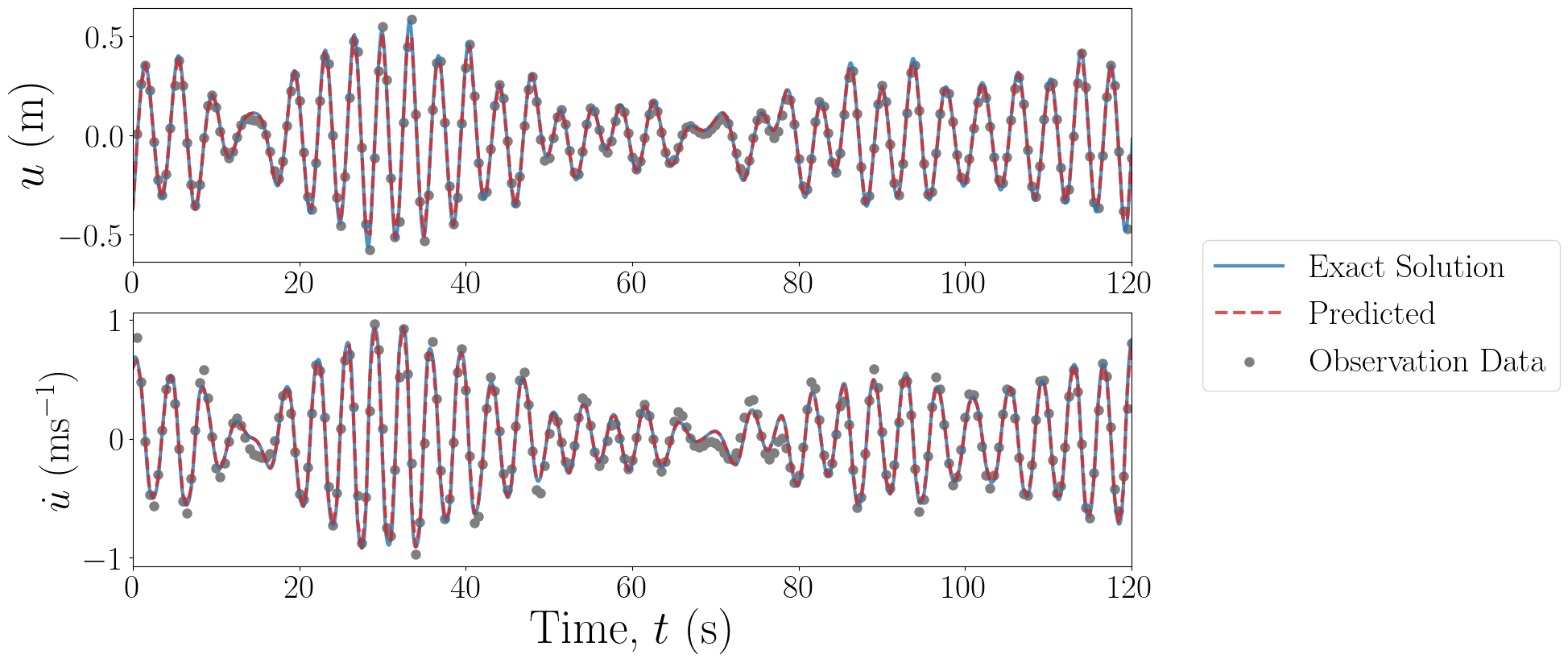}
    \caption{Predicted vs exact solutions of state-space estimation of the Neural-ODE k+1 predictor}
    \label{fig:node_duffing}
\end{figure}

We extend our SDOF oscillator example to other tasks to demonstrate the use of Neural ODEs.
Neural ODEs are trainable forward models in that a neural network is used to approximate the flow of the system of study.
In this context, the neural ODE estimates the latent space $\mathbf{z}_t$ given the initial value problem starting at $(\mathbf{z}_0, t)$. 
The Neural ODE flow estimation is done using the ResNet \citep{he2015deep} such that the integration is accumulated on top of the residuals of the neural network.
The neural network is optimized using the \emph{adjoint sensitivity method} \citep{pontryagin1962mathematical}.
The results are seen in \Cref{fig:node_duffing}.

\begin{figure}[h]
    \centering
    \includegraphics[width=\textwidth]{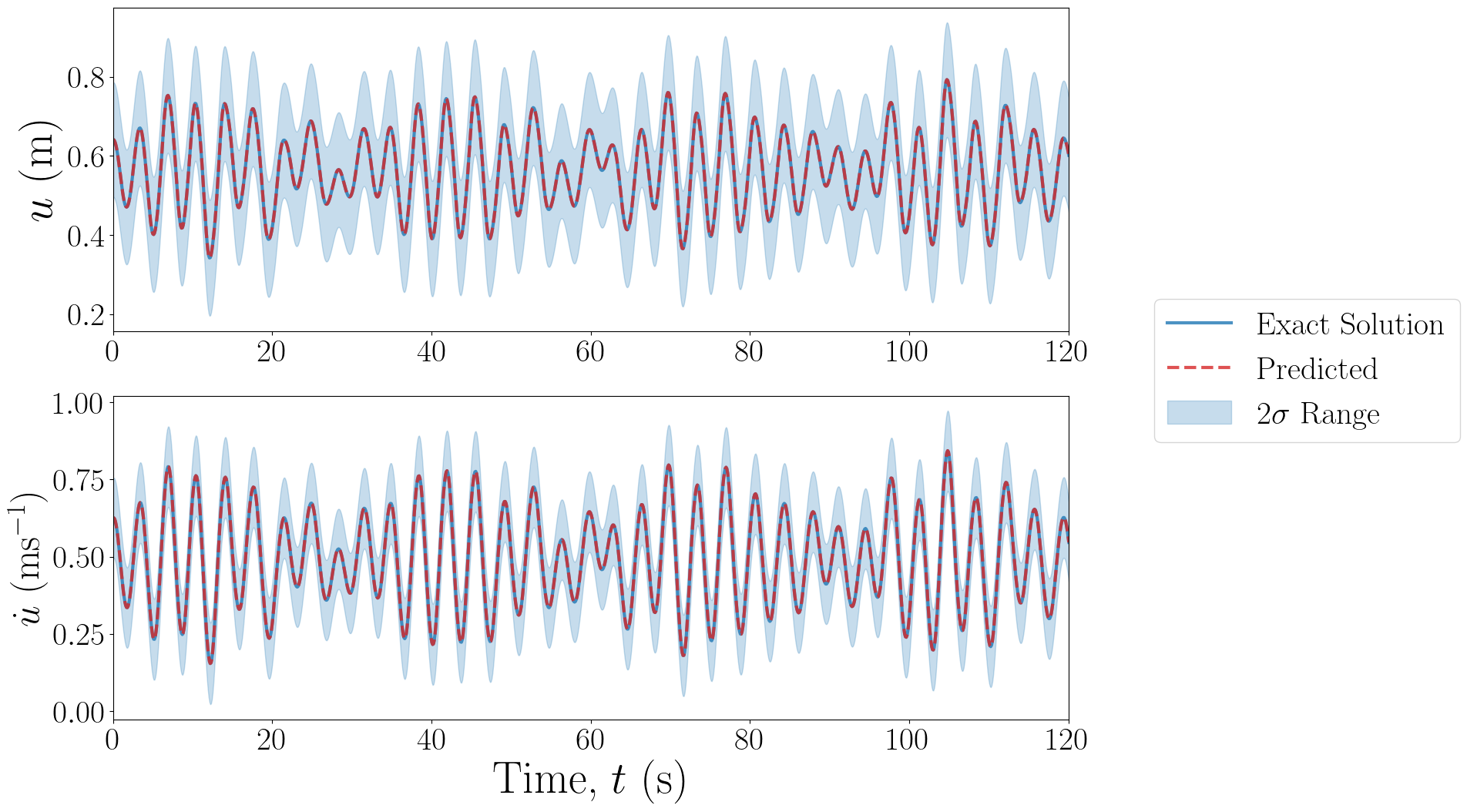}
    \caption{Predicted vs exact solutions of state-space estimation of the Symplectic Neural-ODE encoded DMM k+1 predictor with uncertainty}
    \label{fig:dmm_symplectic_duffing}
\end{figure}

We can extend this method within the DMM framework, using the symplectic Neural ODEs, as per \cite{bacsa2023symplectic}.
We change the problem accordingly: given that symplectic networks are made to deal with limit cycles, the forcing is switched from white excitation to a multisine excitation.
The results are seen in \Cref{fig:dmm_symplectic_duffing}.

% -----------------------------------------------------------------------------
\section{Discussion}

In \Cref{sec:introduction}, we discussed variants situated across the spectrum of \emph{physics-enhanced} machine learning, and examine the characterisation of such methods based on their reliance on the prescription of the physics-based model form (and physics constraints) embedded within the learner, and the amount of data used. %
The selection of an appropriate scheme is driven by the motivation or, in other words, the nature of the downstream task, the level and type of prior scientific/physical knowledge, and the amount of data available. %
When the true system is unknown, or much too complex to define an adequate model, purely data-driven (black-box) models are used. %
As these models are extremely non-generalisable, and are limited to only the scenario for which the data has been collected, they can only reasonably be applied in a sufficiently similar scenario. %
Furthermore, such types of models require training on large amounts of data for reasonable accuracy. %
The advantage, however, of such models is their extreme levels of flexibility; they embed zero prior belief of the true system, and are often described as universal approximators. %

When the true system is relatively simple, and can be adequately modelled with only prior scientific knowledge, white-box models can be used. %
Here, the Bayesian filter approach was presented as an approximation of a white box model, which embeds a strong prior belief on the description of the physics, albeit allowing for some modelling and measurement errors, typically (but not necessarily) assumed Gaussian. %
This results in the physics prescription imposed being highly strict, in that it defines the dynamics model within its specification. %
Comparatively, the \emph{physics-guided} neural network also embeds a relatively high reliance on the physics-based model prescription, however, the NN allows for freedom in the estimation of the model output, akin to a residual modelling scheme. %
Comparing the two light-grey methods discussed in this paper, namely; ML-enhanced Bayesian filtering and physics-guided neural networks, both are often used when supplementary knowledge is required, but most of the underlying physics is well described with prior knowledge. %
Naturally, the flexibility of these approaches is relatively low, a facet sacrificed in order to improve precision in the physical descriptions. %
ML-enhanced BF techniques are often applied as estimators, e.g. serving for the purpose of virtual sensing, within a system identification context. %
\emph{PgNNs}, on the other hand, often aim to determine improved estimates of the measured output (e.g.\ displacement, velocity). %

Light-grey methods often have shared motivations and characteristics as grey methods, but the grey approaches contain a higher degree of flexibility in terms of the embedded belief. %
In the case of `darker' such schemes, the model is still defined with a certain specificity, but they inherit greater flexibility owing to the more dominant incorporation of an ML method. %
However, such a lifting of restrictions stems from a different motivation for each approach. %
Dictionary methods allow for greater flexibility as they allow to combine several possible model forms, whereas the flexibility of \emph{PINNs} maybe attributed to the use of weak-form boundary conditions. %
Thus, in the former class, the embedded belief is essentially summarized in that a sparse representation of the defined dictionary will exhaustively describe the physics of the true system. %
For \emph{PINNs}, the prior belief can be described as a reasonably accurate estimate of the model, but with some discrepancy resulting from either uncertainty in the governing parameters, or in the boundary conditions specification. %

Dark-grey approaches offer maximal flexibility, i.e., maximal potential to deviate from prior assumptions, among the \emph{physics-enhanced} examples shown here. %
For such models, the embedded belief can be translated into regarding prior knowledge as a rather loose description of the \emph{type} of system handled, or the \emph{class}/\emph{family} to which it belongs. %
Thus, such models are useful for improved generalisability, if this is a primary objective of the scheme, but also require a reasonably large amount of data to determine adequate estimates of the system. %
Inadequate levels of data may leave the system underfit, similar to issues with black-box models. %
However, in comparison to black-box models, the encoding the family of the system will allow for better interpolation, or potentially extrapolation, of the model output. %

So far, most of the discussion has been centred around the flexibility and the beliefs embedded, and the fundamental facets the model aims to learn (model output or system parameters). %
But another aspect to consider of these methods is their enhanced-model structure, which can be defined as either unified or superimposed architectures. %
For a unified architecture, the model itself (e.g.\ the network or the kernels) will contain both the machine learning procedure, and the physics embedded. %
In the case of superimposed methods, the ML and the physics models are separate and the output of the model is formed by some combination of these two. %
This characteristic is not as conveniently correlated to a specific location within the spectrum plotted in Figure \ref{fig:paper_layout} as is the aspect of flexibilty. %
The dictionary, \emph{PINN}, and \emph{physics-encoded neural network} (PeNN) approaches can all be considered to be inherently unified models, whilst \emph{PgNN} techniques are naturally superimposed models. %
However, depending on the specific type of approach, constrained GP and ML-enhanced filtering techniques can be either unified or superimposed architectures. %

% -------------------------------------------------------------------------------
\section{Conclusions and Looking to the Future}

This paper has discussed, exhibited, and surveyed the spectrum of \emph{physics-enhanced} machine learning (\emph{PEML}), using the varied attributes of different methods to define and characterise them with respect to such a spectrum. %
This was done via a survey of recent applications/development of \emph{PEML} within the wide field of structural mechanics, and through further demonstration of the alternate schemes on a simple running example of a Duffing oscillator. %
The motivation for, and application of, each of these variants will strongly depend on the use case, and the discussion and detailing of these methods in this paper should help not only the implementation of these techniques, but also for further research using an understanding of the - almost - philosophical implications of each method. %

As we look towards the future of \emph{physics-enhanced} machine learning, many pathways are opened in terms of development, understanding, and improvement. %
An existing challenge in machine learning techniques is to overcome the difficulties that manifest as a result of increased dimensionality of problems. %
This challenge is far from circumvented in \emph{PEML}; in fact, it becomes potentially more prominent in that \emph{PEML} forms a compound of computational paradigms (both physics-based and data-driven), for each of which, dimensionality has a strong influence on the difficulty of implementation. %
One of the biggest advantages of informed ML, in comparison to black-box approaches, is their potential for improved inferability. %
However, the interpretability of the model is more difficult to immediately receive, and so further development can be done to improve this aspect, by utilising domain transforms. %

In order to improve perception and utilisation of \emph{PEML} techniques, the unification of architecture styles, particularly for similar problems, is an invaluable development pathway. %
This could be to provide general design approaches given \emph{PEML} design axioms, or to develop software packages or tools. %
Continuing on the technical development of \emph{PEML} techniques, much like with black-box ML techniques, work is required to improve computing efficiency, especially at a time where reduction of energy consumption is increasingly important. %

This paper has focused on \emph{physics-enhanced} machine learning for structural mechanics, but one could also call attention to the potentially large impact of \emph{PEML} as a natural next step in a society that is increasingly adopting, or opposing, AI. %
There are the clear advantages of improved efficiency, lesser data requirements, and better generalisability. %
A highly impactful societal benefit may arise from improvement of public trust in ML/AI when using informed models, since the opacity of ML models, and the lack of inferability, form a key contributor to public distrust \citep{toreini2020relationship}. %
The techno-societal study of the potential for improved public trust on the basis of \emph{PEML}, would also provide an invaluable knowledge source for modern engineers and researchers, who can leverage such a knowledge for development of tools with high societal impact. %

\begin{Backmatter}

\paragraph{Acknowledgments}
The research was conducted as part of the Future Resilient Systems (FRS) program at the Singapore-ETH Centre, which was established collaboratively between ETH Zurich and the National Research Foundation Singapore. This research is supported by the National Research Foundation, Prime Minister’s Office, Singapore under its Campus for Research Excellence and Technological Enterprise (CREATE) programme.

\paragraph{Funding Statement}
This work received no specific grant from any funding agency, commercial or not-for-profit sectors.

\paragraph{Competing Interests}
None

\paragraph{Data Availability Statement}
The code for the working example shown throughout this paper, including the `ground truth' simulator, is provided in a \hyperlink{https://github.com/ETH-IBK-SMECH/PIDyNN}{Github repository}. The code is freely available, and is written in Python, the specific requirements are provided in the repository. 

\paragraph{Ethical Standards}
The research meets all ethical guidelines, including adherence to the legal requirements of the study country.

\paragraph{Author Contributions}

Conceptualisation: MHA, WL, KB, EC; Methodology: MHA, WL, KB, ZL, EC; Software: MHA, WL, KB, ZL, EC; Investigation: MHA, WL, KB, ZL, EC; Resources: EC; Data Curation: MHA, WL, KB, ZL; Writing - Original Draft: MHA, WL, KB, ZL, EC; Writing - Review \& Editing: MHA, WL, KB, ZL, EC; Supervision: ZL, EC; Project administration: EC; Funding acquisition: EC. All authors approved the final draft.

\paragraph{Supplementary Material}
No supplementary material is given for this paper.

\bibliographystyle{apalike}

\bibliography{DCE_SI_bib}

\begin{thebibliography}{}

\bibitem[Abdessalem et~al., 2018]{abdessalem2018model}
Abdessalem, A.~B., Dervilis, N., Wagg, D., and Worden, K. (2018).
\newblock Model selection and parameter estimation in structural dynamics using
  approximate {Bayesian computation}.
\newblock {\em Mechanical Systems and Signal Processing}, 99:306--325.

\bibitem[Abueidda et~al., 2022]{abueidda2022deep}
Abueidda, D.~W., Koric, S., Al-Rub, R.~A., Parrott, C.~M., James, K.~A., and
  Sobh, N.~A. (2022).
\newblock A deep learning energy method for hyperelasticity and
  viscoelasticity.
\newblock {\em European Journal of Mechanics-A/Solids}, 95:104639.

\bibitem[Abueidda et~al., 2021]{abueidda2021meshless}
Abueidda, D.~W., Lu, Q., and Koric, S. (2021).
\newblock Meshless physics-informed deep learning method for three-dimensional
  solid mechanics.
\newblock {\em International Journal for Numerical Methods in Engineering},
  122(23):7182--7201.

\bibitem[Angeli et~al., 2021]{angeli2021deep}
Angeli, A., Desmet, W., and Naets, F. (2021).
\newblock Deep learning of multibody minimal coordinates for state and input
  estimation with {Kalman} filtering.
\newblock {\em Multibody System Dynamics}, 53(2):205--223.

\bibitem[Aucejo et~al., 2019]{aucejo2019practical}
Aucejo, M., De~Smet, O., and De{\"u}, J.-F. (2019).
\newblock Practical issues on the applicability of {Kalman} filtering for
  reconstructing mechanical sources in structural dynamics.
\newblock {\em Journal of Sound and Vibration}, 442:45--70.

\bibitem[Avenda{\~n}o-Valencia et~al., 2017]{avendano2017gaussian}
Avenda{\~n}o-Valencia, L.~D., Chatzi, E.~N., Koo, K.~Y., and Brownjohn, J.~M.
  (2017).
\newblock Gaussian process time-series models for structures under operational
  variability.
\newblock {\em Frontiers in Built Environment}, 3:69.

\bibitem[Bacsa et~al., 2023]{bacsa2023symplectic}
Bacsa, K., Lai, Z., Liu, W., Todd, M., and Chatzi, E. (2023).
\newblock Symplectic encoders for physics-constrained variational dynamics
  inference.
\newblock {\em Scientific Reports}, 13(1):2643.

\bibitem[Bayer and Osendorfer, 2015]{bayer2015learning}
Bayer, J. and Osendorfer, C. (2015).
\newblock Learning stochastic recurrent networks.

\bibitem[Bergen et~al., 2019]{bergen2019machine}
Bergen, K.~J., Johnson, P.~A., de~Hoop, M.~V., and Beroza, G.~C. (2019).
\newblock Machine learning for data-driven discovery in solid {Earth}
  geoscience.
\newblock {\em Science}, 363(6433):eaau0323.

\bibitem[Bronstein et~al., 2021]{bronstein2021geometric}
Bronstein, M.~M., Bruna, J., Cohen, T., and Veličković, P. (2021).
\newblock Geometric deep learning: Grids, groups, graphs, geodesics, and
  gauges.

\bibitem[Brunton and Kutz, 2022]{brunton2022data}
Brunton, S.~L. and Kutz, J.~N. (2022).
\newblock {\em Data-driven science and engineering: Machine learning, dynamical
  systems, and control}.
\newblock Cambridge University Press.

\bibitem[Brunton et~al., 2016a]{brunton2016discovering}
Brunton, S.~L., Proctor, J.~L., and Kutz, J.~N. (2016a).
\newblock Discovering governing equations from data by sparse identification of
  nonlinear dynamical systems.
\newblock {\em Proceedings of the national academy of sciences},
  113(15):3932--3937.

\bibitem[Brunton et~al., 2016b]{brunton2016sparse}
Brunton, S.~L., Proctor, J.~L., and Kutz, J.~N. (2016b).
\newblock Sparse identification of nonlinear dynamics with control {(SINDYc)}.
\newblock {\em IFAC-PapersOnLine}, 49(18):710--715.

\bibitem[Chatzi and Smyth, 2009]{Chatzi2009}
Chatzi, E.~N. and Smyth, A.~W. (2009).
\newblock The unscented {Kalman} filter and particle filter methods for
  nonlinear structural system identification with non-collocated heterogeneous
  sensing.
\newblock {\em Structural Control and Health Monitoring}, 16(1):99--123.

\bibitem[Chatzi et~al., 2010]{CHATZI2010}
Chatzi, E.~N., Smyth, A.~W., and Masri, S.~F. (2010).
\newblock Experimental application of on-line parametric identification for
  nonlinear hysteretic systems with model uncertainty.
\newblock {\em Structural Safety}, 32(5):326--337.
\newblock Probabilistic Methods for Modeling, Simulation and Optimization of
  Engineering Structures under Uncertainty in honor of Jim Beck’s 60th
  Birthday.

\bibitem[Chatzis et~al., 2015]{Chatzis2015}
Chatzis, M.~N., Chatzi, E.~N., and Smyth, A.~W. (2015).
\newblock On the observability and identifiability of nonlinear structural and
  mechanical systems.
\newblock {\em Structural Control and Health Monitoring}, 22(3):574--593.

\bibitem[Chen and Liu, 2021]{chen2021probabilistic}
Chen, J. and Liu, Y. (2021).
\newblock Probabilistic physics-guided machine learning for fatigue data
  analysis.
\newblock {\em Expert Systems with Applications}, 168:114316.

\bibitem[Chen et~al., 2018]{chen2018neural}
Chen, R.~T., Rubanova, Y., Bettencourt, J., and Duvenaud, D.~K. (2018).
\newblock Neural ordinary differential equations.
\newblock {\em Advances in neural information processing systems}, 31.

\bibitem[Chen et~al., 2020]{chen2020symplectic}
Chen, Z., Zhang, J., Arjovsky, M., and Bottou, L. (2020).
\newblock Symplectic recurrent neural networks.

\bibitem[Choudhary et~al., 2020]{choudhary2020physics}
Choudhary, A., Lindner, J.~F., Holliday, E.~G., Miller, S.~T., Sinha, S., and
  Ditto, W.~L. (2020).
\newblock Physics-enhanced neural networks learn order and chaos.
\newblock {\em Physical Review E}, 101(6):062207.

\bibitem[Christodoulou and Papadimitriou, 2007]{christodoulou2007structural}
Christodoulou, K. and Papadimitriou, C. (2007).
\newblock Structural identification based on optimally weighted modal
  residuals.
\newblock {\em Mechanical Systems and Signal Processing}, 21(1):4--23.

\bibitem[Cross et~al., 2013]{cross2013long}
Cross, E., Koo, K., Brownjohn, J., and Worden, K. (2013).
\newblock Long-term monitoring and data analysis of the {Tamar} bridge.
\newblock {\em Mechanical Systems and Signal Processing}, 35(1-2):16--34.

\bibitem[Cross et~al., 2022]{cross2022physics}
Cross, E.~J., Gibson, S.~J., Jones, M.~R., Pitchforth, D.~J., Zhang, S., and
  Rogers, T.~J. (2022).
\newblock Physics-informed machine learning for structural health monitoring.
\newblock {\em Structural Health Monitoring Based on Data Science Techniques},
  pages 347--367.

\bibitem[Cross and Rogers, 2021]{cross2021physics}
Cross, E.~J. and Rogers, T.~J. (2021).
\newblock Physics-derived covariance functions for machine learning in
  structural dynamics.
\newblock {\em IFAC-PapersOnLine}, 54(7):168--173.

\bibitem[Cross et~al., 2011a]{cross2011cointegration_a}
Cross, E.~J., Worden, K., and Chen, Q. (2011a).
\newblock Cointegration: a novel approach for the removal of environmental
  trends in structural health monitoring data.
\newblock {\em Proceedings of the Royal Society A: Mathematical, Physical and
  Engineering Sciences}, 467(2133):2712--2732.

\bibitem[Cross et~al., 2011b]{cross2011cointegration_b}
Cross, E.~J., Worden, K., and Chen, Q. (2011b).
\newblock Cointegration: a novel approach for the removal of environmental
  trends in structural health monitoring data.
\newblock {\em Proceedings of the Royal Society A: Mathematical, Physical and
  Engineering Sciences}, 467(2133):2712--2732.

\bibitem[Cuomo et~al., 2022]{cuomo2022scientific}
Cuomo, S., Di~Cola, V.~S., Giampaolo, F., Rozza, G., Raissi, M., and Piccialli,
  F. (2022).
\newblock Scientific machine learning through physics--informed neural
  networks: where we are and what’s next.
\newblock {\em Journal of Scientific Computing}, 92(3):88.

\bibitem[D'Amico et~al., 2019]{d2019machine}
D'Amico, B., Myers, R.~J., Sykes, J., Voss, E., Cousins-Jenvey, B., Fawcett,
  W., Richardson, S., Kermani, A., and Pomponi, F. (2019).
\newblock Machine learning for sustainable structures: a call for data.
\newblock In {\em Structures}, volume~19, pages 1--4. Elsevier.

\bibitem[Dardeno et~al., 2021]{dardeno2021investigating}
Dardeno, T.~A., Haywood-Alexander, M., Mills, R.~S., Bull, L.~A., Dervilis, N.,
  and Worden, K. (2021).
\newblock Investigating the effects of ambient temperature on feature
  consistency in vibration-based {SHM}.
\newblock In {\em International Workshop on Structural Health Monitoring 2021}.

\bibitem[David and Méhats, 2021]{david2021symplectic}
David, M. and Méhats, F. (2021).
\newblock Symplectic learning for {Hamiltonian} neural networks.

\bibitem[Dertimanis et~al., 2019]{dertimanis2019input}
Dertimanis, V.~K., Chatzi, E., Azam, S.~E., and Papadimitriou, C. (2019).
\newblock Input-state-parameter estimation of structural systems from limited
  output information.
\newblock {\em Mechanical Systems and Signal Processing}, 126:711--746.

\bibitem[Desai et~al., 2021]{desaiporthamiltonian2021}
Desai, S.~A., Mattheakis, M., Sondak, D., Protopapas, P., and Roberts, S.~J.
  (2021).
\newblock {Port-Hamiltonian} neural networks for learning explicit
  time-dependent dynamical systems.
\newblock {\em Physical Review E}, 104(3).

\bibitem[Dhadphale et~al., 2022]{dhadphale2022neural}
Dhadphale, J.~M., Unni, V.~R., Saha, A., and Sujith, R. (2022).
\newblock Neural ode to model and prognose thermoacoustic instability.
\newblock {\em Chaos: An Interdisciplinary Journal of Nonlinear Science},
  32(1).

\bibitem[Dou et~al., 2023]{dou2023machine}
Dou, B., Zhu, Z., Merkurjev, E., Ke, L., Chen, L., Jiang, J., Zhu, Y., Liu, J.,
  Zhang, B., and Wei, G.-W. (2023).
\newblock Machine learning methods for small data challenges in molecular
  science.
\newblock {\em Chemical Reviews}, 123(13):8736--8780.

\bibitem[Duenas-Osorio and Vemuru, 2009]{duenas2009cascading}
Duenas-Osorio, L. and Vemuru, S.~M. (2009).
\newblock Cascading failures in complex infrastructure systems.
\newblock {\em Structural safety}, 31(2):157--167.

\bibitem[{Eftekhar Azam} et~al., 2015]{EFTEKHARAZAM2015866}
{Eftekhar Azam}, S., Chatzi, E., and Papadimitriou, C. (2015).
\newblock A dual {Kalman} filter approach for state estimation via output-only
  acceleration measurements.
\newblock {\em Mechanical Systems and Signal Processing}, 60-61:866--886.

\bibitem[Erazo et~al., 2019]{erazo2019vibration}
Erazo, K., Sen, D., Nagarajaiah, S., and Sun, L. (2019).
\newblock Vibration-based structural health monitoring under changing
  environmental conditions using {Kalman} filtering.
\newblock {\em Mechanical systems and signal processing}, 117:1--15.

\bibitem[Faroughi et~al., 2022]{faroughi2022physics}
Faroughi, S.~A., Pawar, N., Fernandes, C., Das, S., Kalantari, N.~K., and
  Mahjour, S.~K. (2022).
\newblock Physics-guided, physics-informed, and physics-encoded neural networks
  in scientific computing.
\newblock {\em arXiv preprint arXiv:2211.07377}.

\bibitem[Farrar and Worden, 2007]{farrar2007introduction}
Farrar, C.~R. and Worden, K. (2007).
\newblock An introduction to structural health monitoring.
\newblock {\em Philosophical Transactions of the Royal Society A: Mathematical,
  Physical and Engineering Sciences}, 365(1851):303--315.

\bibitem[Farrar and Worden, 2012]{farrar2012structural}
Farrar, C.~R. and Worden, K. (2012).
\newblock {\em Structural health monitoring: a machine learning perspective}.
\newblock John Wiley \& Sons.

\bibitem[Feng et~al., 2020]{feng2020force}
Feng, W., Li, Q., and Lu, Q. (2020).
\newblock Force localization and reconstruction based on a novel sparse
  {Kalman} filter.
\newblock {\em Mechanical Systems and Signal Processing}, 144:106890.

\bibitem[Figueiredo et~al., 2011]{figueiredo2011machine}
Figueiredo, E., Park, G., Farrar, C.~R., Worden, K., and Figueiras, J. (2011).
\newblock Machine learning algorithms for damage detection under operational
  and environmental variability.
\newblock {\em Structural Health Monitoring}, 10(6):559--572.

\bibitem[Flaschel et~al., 2021]{flaschel2021unsupervised}
Flaschel, M., Kumar, S., and De~Lorenzis, L. (2021).
\newblock Unsupervised discovery of interpretable hyperelastic constitutive
  laws.
\newblock {\em Computer Methods in Applied Mechanics and Engineering},
  381:113852.

\bibitem[Frangos et~al., 2010]{frangos2010surrogate}
Frangos, M., Marzouk, Y., Willcox, K., and van Bloemen~Waanders, B. (2010).
\newblock Surrogate and reduced-order modeling: a comparison of approaches for
  large-scale statistical inverse problems.
\newblock {\em Large-Scale Inverse Problems and Quantification of Uncertainty},
  pages 123--149.

\bibitem[Fuentes et~al., 2021]{fuentes2021equation}
Fuentes, R., Nayek, R., Gardner, P., Dervilis, N., Rogers, T., Worden, K., and
  Cross, E. (2021).
\newblock Equation discovery for nonlinear dynamical systems: A {Bayesian}
  viewpoint.
\newblock {\em Mechanical Systems and Signal Processing}, 154:107528.

\bibitem[Geyer et~al., 2021]{geyer2021explainable}
Geyer, P., Singh, M.~M., and Chen, X. (2021).
\newblock Explainable {AI} for engineering design: A unified approach of
  systems engineering and component-based deep learning.
\newblock {\em arXiv preprint arXiv:2108.13836}.

\bibitem[Goswami et~al., 2020]{goswami2020transfer}
Goswami, S., Anitescu, C., Chakraborty, S., and Rabczuk, T. (2020).
\newblock Transfer learning enhanced physics informed neural network for
  phase-field modeling of fracture.
\newblock {\em Theoretical and Applied Fracture Mechanics}, 106:102447.

\bibitem[Goyal and Benner, 2022]{goyal2022discovery}
Goyal, P. and Benner, P. (2022).
\newblock Discovery of nonlinear dynamical systems using a {Runge--Kutta}
  inspired dictionary-based sparse regression approach.
\newblock {\em Proceedings of the Royal Society A}, 478(2262):20210883.

\bibitem[Gre{\'s} et~al., 2021]{gres2021kalman}
Gre{\'s}, S., D{\"o}hler, M., Andersen, P., and Mevel, L. (2021).
\newblock {Kalman} filter-based subspace identification for operational modal
  analysis under unmeasured periodic excitation.
\newblock {\em Mechanical Systems and Signal Processing}, 146:106996.

\bibitem[Greydanus et~al., 2019]{greydanus2019hamiltonian}
Greydanus, S., Dzamba, M., and Yosinski, J. (2019).
\newblock Hamiltonian neural networks.
\newblock {\em CoRR}, abs/1906.01563.

\bibitem[Grossmann et~al., 2023]{grossmann2023can}
Grossmann, T.~G., Komorowska, U.~J., Latz, J., and Sch{\"o}nlieb, C.-B. (2023).
\newblock Can physics-informed neural networks beat the finite element method?
\newblock {\em arXiv preprint arXiv:2302.04107}.

\bibitem[Guo et~al., 2021]{guo2021artificial}
Guo, K., Yang, Z., Yu, C.-H., and Buehler, M.~J. (2021).
\newblock Artificial intelligence and machine learning in design of mechanical
  materials.
\newblock {\em Materials Horizons}, 8(4):1153--1172.

\bibitem[Haghighat et~al., 2021a]{haghighat2021deep}
Haghighat, E., Bekar, A.~C., Madenci, E., and Juanes, R. (2021a).
\newblock Deep learning for solution and inversion of structural mechanics and
  vibrations.
\newblock In {\em Modeling and Computation in Vibration Problems, Volume 2:
  Soft computing and uncertainty}. IOP Publishing.

\bibitem[Haghighat et~al., 2021b]{haghighat2021physics}
Haghighat, E., Raissi, M., Moure, A., Gomez, H., and Juanes, R. (2021b).
\newblock A physics-informed deep learning framework for inversion and
  surrogate modeling in solid mechanics.
\newblock {\em Computer Methods in Applied Mechanics and Engineering},
  379:113741.

\bibitem[Haywood-Alexander et~al., 2021]{haywood2021structured}
Haywood-Alexander, M., Dervilis, N., Worden, K., Cross, E.~J., Mills, R.~S.,
  and Rogers, T.~J. (2021).
\newblock Structured machine learning tools for modelling characteristics of
  guided waves.
\newblock {\em Mechanical Systems and Signal Processing}, 156:107628.

\bibitem[He et~al., 2015]{he2015deep}
He, K., Zhang, X., Ren, S., and Sun, J. (2015).
\newblock Deep residual learning for image recognition.

\bibitem[Henkes et~al., 2022]{henkes2022physics}
Henkes, A., Wessels, H., and Mahnken, R. (2022).
\newblock Physics informed neural networks for continuum micromechanics.
\newblock {\em Computer Methods in Applied Mechanics and Engineering},
  393:114790.

\bibitem[Hey et~al., 2020]{hey2020machine}
Hey, T., Butler, K., Jackson, S., and Thiyagalingam, J. (2020).
\newblock Machine learning and big scientific data.
\newblock {\em Philosophical Transactions of the Royal Society A},
  378(2166):20190054.

\bibitem[Huang et~al., 2020]{huang2020machine}
Huang, D., Fuhg, J.~N., Wei{\ss}enfels, C., and Wriggers, P. (2020).
\newblock A machine learning based plasticity model using proper orthogonal
  decomposition.
\newblock {\em Computer Methods in Applied Mechanics and Engineering},
  365:113008.

\bibitem[Huang et~al., 2022]{huang2022physics}
Huang, Z., Yin, X., and Liu, Y. (2022).
\newblock Physics-guided deep neural network for structural damage
  identification.
\newblock {\em Ocean Engineering}, 260:112073.

\bibitem[Jia et~al., 2019]{jia2019physics}
Jia, X., Willard, J., Karpatne, A., Read, J., Zwart, J., Steinbach, M., and
  Kumar, V. (2019).
\newblock Physics guided {RNNs} for modeling dynamical systems: A case study in
  simulating lake temperature profiles.
\newblock In {\em Proceedings of the 2019 SIAM international conference on data
  mining}, pages 558--566. SIAM.

\bibitem[Jones et~al., 2023]{jones2023constraining}
Jones, M.~R., Rogers, T.~J., and Cross, E.~J. (2023).
\newblock Constraining {Gaussian} processes for physics-informed acoustic
  emission mapping.
\newblock {\em Mechanical Systems and Signal Processing}, 188:109984.

\bibitem[Jordan and Mitchell, 2015]{jordan2015machine}
Jordan, M.~I. and Mitchell, T.~M. (2015).
\newblock Machine learning: Trends, perspectives, and prospects.
\newblock {\em Science}, 349(6245):255--260.

\bibitem[Joshi et~al., 2022]{joshi2022bayesian}
Joshi, A., Thakolkaran, P., Zheng, Y., Escande, M., Flaschel, M., De~Lorenzis,
  L., and Kumar, S. (2022).
\newblock {Bayesian-EUCLID: Discovering hyperelastic material laws with
  uncertainties}.
\newblock {\em Computer Methods in Applied Mechanics and Engineering},
  398:115225.

\bibitem[Kaiser et~al., 2018]{kaiser2018sparse}
Kaiser, E., Kutz, J.~N., and Brunton, S.~L. (2018).
\newblock Sparse identification of nonlinear dynamics for model predictive
  control in the low-data limit.
\newblock {\em Proceedings of the Royal Society A}, 474(2219):20180335.

\bibitem[Kamariotis et~al., 2023]{Kamariotis_2023}
Kamariotis, A., Sardi, L., Papaioannou, I., Chatzi, E., and Straub, D. (2023).
\newblock On off-line and on-line {Bayesian} filtering for uncertainty
  quantification of structural deterioration.
\newblock {\em Data-Centric Engineering}, 4:e17.

\bibitem[Karniadakis et~al., 2021]{karniadakis2021physics}
Karniadakis, G.~E., Kevrekidis, I.~G., Lu, L., Perdikaris, P., Wang, S., and
  Yang, L. (2021).
\newblock Physics-informed machine learning.
\newblock {\em Nature Reviews Physics}, 3(6):422--440.

\bibitem[Karpatne et~al., 2017]{karpatne2017physics}
Karpatne, A., Watkins, W., Read, J., and Kumar, V. (2017).
\newblock Physics-guided neural networks (pgnn): An application in lake
  temperature modeling.
\newblock {\em arXiv preprint arXiv:1710.11431}, 2.

\bibitem[Kim et~al., 2017]{kim2017failure}
Kim, J.-H., Jin, J.-W., Lee, J.-H., and Kang, K.-W. (2017).
\newblock Failure analysis for vibration-based energy harvester utilized in
  high-speed railroad vehicle.
\newblock {\em Engineering Failure Analysis}, 73:85--96.

\bibitem[Kingma and Ba, 2014]{kingma2014adam}
Kingma, D.~P. and Ba, J. (2014).
\newblock Adam: A method for stochastic optimization.
\newblock {\em arXiv preprint arXiv:1412.6980}.

\bibitem[Kontoroupi and Smyth, 2016]{doi:10.1061/AJRUA6.0000839}
Kontoroupi, T. and Smyth, A.~W. (2016).
\newblock Online noise identification for joint state and parameter estimation
  of nonlinear systems.
\newblock {\em ASCE-ASME Journal of Risk and Uncertainty in Engineering
  Systems, Part A: Civil Engineering}, 2(3):B4015006.

\bibitem[Krishnan et~al., 2016]{krishnan_structured_2016}
Krishnan, R.~G., Shalit, U., and Sontag, D. (2016).
\newblock Structured {Inference} {Networks} for {Nonlinear} {State} {Space}
  {Models}.
\newblock {\em arXiv:1609.09869 [cs, stat]}.
\newblock arXiv: 1609.09869.

\bibitem[Lai et~al., 2022]{lai_liu_jian_bacsa_sun_chatzi_2022}
Lai, Z., Liu, W., Jian, X., Bacsa, K., Sun, L., and Chatzi, E. (2022).
\newblock Neural modal ordinary differential equations: Integrating
  physics-based modeling with neural ordinary differential equations for
  modeling high-dimensional monitored structures.
\newblock {\em Data-Centric Engineering}, 3:e34.

\bibitem[Lai et~al., 2021]{lai2021structural}
Lai, Z., Mylonas, C., Nagarajaiah, S., and Chatzi, E. (2021).
\newblock Structural identification with physics-informed neural ordinary
  differential equations.
\newblock {\em Journal of Sound and Vibration}, 508:116196.

\bibitem[Langley et~al., 1994]{langley1994selection}
Langley, P. et~al. (1994).
\newblock Selection of relevant features in machine learning.
\newblock In {\em Proceedings of the AAAI Fall symposium on relevance}, volume
  184, pages 245--271. California.

\bibitem[Li et~al., 2021]{li2021physics}
Li, W., Bazant, M.~Z., and Zhu, J. (2021).
\newblock A physics-guided neural network framework for elastic plates:
  Comparison of governing equations-based and energy-based approaches.
\newblock {\em Computer Methods in Applied Mechanics and Engineering},
  383:113933.

\bibitem[Linardatos et~al., 2020]{linardatos2020explainable}
Linardatos, P., Papastefanopoulos, V., and Kotsiantis, S. (2020).
\newblock Explainable ai: A review of machine learning interpretability
  methods.
\newblock {\em Entropy}, 23(1):18.

\bibitem[Liu and Nocedal, 1989]{liu1989limited}
Liu, D.~C. and Nocedal, J. (1989).
\newblock On the limited memory {BFGS} method for large scale optimization.
\newblock {\em Mathematical programming}, 45(1-3):503--528.

\bibitem[Liu et~al., 2022]{liu2022physics}
Liu, W., Lai, Z., Bacsa, K., and Chatzi, E. (2022).
\newblock Physics-guided deep markov models for learning nonlinear dynamical
  systems with uncertainty.
\newblock {\em Mechanical Systems and Signal Processing}, 178:109276.

\bibitem[Liu et~al., 2021]{liu2021physics}
Liu, W., Lai, Z., and Chatzi, E. (2021).
\newblock A physics-guided deep learning approach to modeling nonlinear
  dynamics: A case study of a {Bouc-Wen} system.
\newblock {\em STRUCTURAL HEALTH MONITORING 2021}.

\bibitem[Lynch, 2007]{lynch2007overview}
Lynch, J.~P. (2007).
\newblock An overview of wireless structural health monitoring for civil
  structures.
\newblock {\em Philosophical Transactions of the Royal Society A: Mathematical,
  Physical and Engineering Sciences}, 365(1851):345--372.

\bibitem[Maes et~al., 2019]{MAES2019378}
Maes, K., Chatzis, M., and Lombaert, G. (2019).
\newblock Observability of nonlinear systems with unmeasured inputs.
\newblock {\em Mechanical Systems and Signal Processing}, 130:378--394.

\bibitem[Maes et~al., 2018]{MAES2018292}
Maes, K., Gillijns, S., and Lombaert, G. (2018).
\newblock A smoothing algorithm for joint input-state estimation in structural
  dynamics.
\newblock {\em Mechanical Systems and Signal Processing}, 98:292--309.

\bibitem[Molnar, 2020]{molnar2020interpretable}
Molnar, C. (2020).
\newblock {\em Interpretable machine learning}.
\newblock Lulu. com.

\bibitem[Moradi et~al., 2023]{moradi2023novel}
Moradi, S., Duran, B., Eftekhar~Azam, S., and Mofid, M. (2023).
\newblock Novel physics-informed artificial neural network architectures for
  system and input identification of structural dynamics {PDEs}.
\newblock {\em Buildings}, 13(3):650.

\bibitem[Muralidhar et~al., 2020]{muralidhar2020phynet}
Muralidhar, N., Bu, J., Cao, Z., He, L., Ramakrishnan, N., Tafti, D., and
  Karpatne, A. (2020).
\newblock Phynet: Physics guided neural networks for particle drag force
  prediction in assembly.
\newblock In {\em Proceedings of the 2020 SIAM International Conference on Data
  Mining}, pages 559--567. SIAM.

\bibitem[Mylonas et~al., 2021]{MylonasCVAE}
Mylonas, C., Abdallah, I., and Chatzi, E. (2021).
\newblock Conditional variational autoencoders for probabilistic wind turbine
  blade fatigue estimation using supervisory, control, and data acquisition
  data.
\newblock {\em Wind Energy}, 24(10):1122--1139.

\bibitem[Naets et~al., 2015]{naets2015online}
Naets, F., Croes, J., and Desmet, W. (2015).
\newblock An online coupled state/input/parameter estimation approach for
  structural dynamics.
\newblock {\em Computer Methods in Applied Mechanics and Engineering},
  283:1167--1188.

\bibitem[Nandakumar and Jacob, 2021]{nandakumar2021structural}
Nandakumar, P. and Jacob, J. (2021).
\newblock Structural and crack parameter identification on structures using
  observer {Kalman} filter identification/eigen system realization algorithm.
\newblock {\em Journal of Solid Mechanics}, 13(1):68--79.

\bibitem[Nandi et~al., 2021]{nandi2021progress}
Nandi, T., Hennigh, O., Nabian, M., Liu, Y., Woo, M., Jordan, T., Shahnam, M.,
  Syamlal, M., Guenther, C., and VanEssendelft, D. (2021).
\newblock Progress towards solving high reynolds number reacting flows in
  simnet.
\newblock Technical report, National Energy Technology Laboratory (NETL),
  Pittsburgh, PA, Morgantown, WV~….

\bibitem[Naser, 2021]{naser2021engineer}
Naser, M. (2021).
\newblock An engineer's guide to {eXplainable} artificial intelligence and
  interpretable machine learning: Navigating causality, forced goodness, and
  the false perception of inference.
\newblock {\em Automation in Construction}, 129:103821.

\bibitem[Nayek et~al., 2019]{NAYEK2019497}
Nayek, R., Chakraborty, S., and Narasimhan, S. (2019).
\newblock A {Gaussian} process latent force model for joint input-state
  estimation in linear structural systems.
\newblock {\em Mechanical Systems and Signal Processing}, 128:497--530.

\bibitem[Nayek et~al., 2021]{nayek2021spike}
Nayek, R., Fuentes, R., Worden, K., and Cross, E.~J. (2021).
\newblock On spike-and-slab priors for {Bayesian} equation discovery of
  nonlinear dynamical systems via sparse linear regression.
\newblock {\em Mechanical Systems and Signal Processing}, 161:107986.

\bibitem[N{\'u}{\~n}ez et~al., 2023]{nunez2023forecasting}
N{\'u}{\~n}ez, M., Barreiro, N.~L., Barrio, R.~A., and Rackauckas, C. (2023).
\newblock Forecasting virus outbreaks with social media data via neural
  ordinary differential equations.
\newblock {\em Scientific Reports}, 13(1):10870.

\bibitem[Odelson et~al., 2006]{ODELSON2006303}
Odelson, B.~J., Rajamani, M.~R., and Rawlings, J.~B. (2006).
\newblock A new autocovariance least-squares method for estimating noise
  covariances.
\newblock {\em Automatica}, 42(2):303--308.

\bibitem[O'Driscoll et~al., 2019]{o2019physics}
O'Driscoll, P., Lee, J., and Fu, B. (2019).
\newblock Physics enhanced artificial intelligence.
\newblock {\em arXiv preprint arXiv:1903.04442}.

\bibitem[O'Hagan, 1978]{o1978curve}
O'Hagan, A. (1978).
\newblock Curve fitting and optimal design for prediction.
\newblock {\em Journal of the Royal Statistical Society: Series B
  (Methodological)}, 40(1):1--24.

\bibitem[Olson et~al., 2018]{olson2018data}
Olson, R.~S., Cava, W.~L., Mustahsan, Z., Varik, A., and Moore, J.~H. (2018).
\newblock Data-driven advice for applying machine learning to bioinformatics
  problems.
\newblock In {\em Pacific Symposium on Biocomputing 2018: Proceedings of the
  Pacific Symposium}, pages 192--203. World Scientific.

\bibitem[Padonou and Roustant, 2016]{padonou2016polar}
Padonou, E. and Roustant, O. (2016).
\newblock {Polar {Gaussian} Processes and Experimental Designs in Circular
  Domains}.
\newblock working paper or preprint.

\bibitem[Papadimitriou and Katafygiotis, 2004]{papadimitriou2004bayesian}
Papadimitriou, C. and Katafygiotis, L.~S. (2004).
\newblock Bayesian modeling and updating.
\newblock In {\em Engineering design reliability handbook}, pages 525--544. CRC
  Press.

\bibitem[Park and Casella, 2008]{park2008bayesian}
Park, T. and Casella, G. (2008).
\newblock The {Bayesian} lasso.
\newblock {\em Journal of the American Statistical Association},
  103(482):681--686.

\bibitem[Paszke et~al., 2019]{NEURIPS2019_pytorch}
Paszke, A., Gross, S., Massa, F., Lerer, A., Bradbury, J., Chanan, G., Killeen,
  T., Lin, Z., Gimelshein, N., Antiga, L., Desmaison, A., Kopf, A., Yang, E.,
  DeVito, Z., Raison, M., Tejani, A., Chilamkurthy, S., Steiner, B., Fang, L.,
  Bai, J., and Chintala, S. (2019).
\newblock {PyTorch}: An imperative style, high-performance deep learning
  library.
\newblock In Wallach, H., Larochelle, H., Beygelzimer, A., d\textquotesingle
  Alch\'{e}-Buc, F., Fox, E., and Garnett, R., editors, {\em Advances in Neural
  Information Processing Systems 32}, pages 8024--8035. Curran Associates, Inc.

\bibitem[Pawar et~al., 2021]{pawar2021model}
Pawar, S., San, O., Nair, A., Rasheed, A., and Kvamsdal, T. (2021).
\newblock Model fusion with physics-guided machine learning: Projection-based
  reduced-order modeling.
\newblock {\em Physics of Fluids}, 33(6).

\bibitem[Peeters and De~Roeck, 2001]{peeters2001stochastic}
Peeters, B. and De~Roeck, G. (2001).
\newblock Stochastic system identification for operational modal analysis: a
  review.
\newblock {\em Journal Dynamic Systems, Measurements, and Control},
  123(4):659--667.

\bibitem[Petersen et~al., 2022]{petersen2022wind}
Petersen, {\O}., {\O}iseth, O., and Lourens, E. (2022).
\newblock Wind load estimation and virtual sensing in long-span suspension
  bridges using physics-informed {Gaussian} process latent force models.
\newblock {\em Mechanical Systems and Signal Processing}, 170:108742.

\bibitem[Pontryagin et~al., 1962]{pontryagin1962mathematical}
Pontryagin, L.~S., Mishchenko, E., Boltyanskii, V., and Gamkrelidze, R. (1962).
\newblock {\em The mathematical theory of optimal processes}.
\newblock Wiley.

\bibitem[Raissi et~al., 2019]{raissi2019physics}
Raissi, M., Perdikaris, P., and Karniadakis, G.~E. (2019).
\newblock Physics-informed neural networks: A deep learning framework for
  solving forward and inverse problems involving nonlinear partial differential
  equations.
\newblock {\em Journal of Computational physics}, 378:686--707.

\bibitem[Rebillat et~al., 2023]{rebillat2023physically}
Rebillat, M., Monteiro, E., and Mechbal, N. (2023).
\newblock Physically informed and data driven direct models for {Lamb} waves
  based shm: Advantages and drawbacks of existing approaches.
\newblock In {\em Proceedings of the 14th International Workshop on Structural
  Health Monitoring (IWSHM 2024)}, pages 645--652. IWSHM.

\bibitem[Reich, 1997]{reich1997machine}
Reich, Y. (1997).
\newblock Machine learning techniques for civil engineering problems.
\newblock {\em Computer-Aided Civil and Infrastructure Engineering},
  12(4):295--310.

\bibitem[Reimann et~al., 2019]{reimann2019modeling}
Reimann, D., Nidadavolu, K., ul~Hassan, H., Vajragupta, N., Glasmachers, T.,
  Junker, P., and Hartmaier, A. (2019).
\newblock Modeling macroscopic material behavior with machine learning
  algorithms trained by micromechanical simulations.
\newblock {\em Frontiers in Materials}, 6:181.

\bibitem[Ren et~al., 2023]{ren2023data}
Ren, Z., Han, X., Yu, X., Skjetne, R., Leira, B.~J., S{\ae}vik, S., and Zhu, M.
  (2023).
\newblock Data-driven simultaneous identification of the {6DOF} dynamic model
  and wave load for a ship in waves.
\newblock {\em Mechanical Systems and Signal Processing}, 184:109422.

\bibitem[Revach et~al., 2022]{revach2022kalmannet}
Revach, G., Shlezinger, N., Ni, X., Escoriza, A.~L., Van~Sloun, R.~J., and
  Eldar, Y.~C. (2022).
\newblock {KalmanNet: Neural network aided Kalman} filtering for partially
  known dynamics.
\newblock {\em IEEE Transactions on Signal Processing}, 70:1532--1547.

\bibitem[Rezaei et~al., 2022]{rezaei2022mixed}
Rezaei, S., Harandi, A., Moeineddin, A., Xu, B.-X., and Reese, S. (2022).
\newblock A mixed formulation for physics-informed neural networks as a
  potential solver for engineering problems in heterogeneous domains:
  comparison with finite element method.
\newblock {\em Computer Methods in Applied Mechanics and Engineering},
  401:115616.

\bibitem[Ritto and Rochinha, 2021]{ritto2021digital}
Ritto, T. and Rochinha, F. (2021).
\newblock Digital twin, physics-based model, and machine learning applied to
  damage detection in structures.
\newblock {\em Mechanical Systems and Signal Processing}, 155:107614.

\bibitem[Robinson et~al., 2022]{robinson2022physics}
Robinson, H., Pawar, S., Rasheed, A., and San, O. (2022).
\newblock Physics guided neural networks for modelling of non-linear dynamics.
\newblock {\em Neural Networks}, 154:333--345.

\bibitem[Rogers et~al., 2020]{ROGERS2020106580}
Rogers, T., Worden, K., and Cross, E. (2020).
\newblock On the application of {Gaussian} process latent force models for
  joint input-state-parameter estimation: With a view to {Bayesian} operational
  identification.
\newblock {\em Mechanical Systems and Signal Processing}, 140:106580.

\bibitem[Saemundsson et~al., 2020]{saemundsson2020variational}
Saemundsson, S., Terenin, A., Hofmann, K., and Deisenroth, M.~P. (2020).
\newblock Variational integrator networks for physically structured embeddings.

\bibitem[Sanchez-Gonzalez et~al., 2019]{sanchezgonzalez2019hamiltonian}
Sanchez-Gonzalez, A., Bapst, V., Cranmer, K., and Battaglia, P. (2019).
\newblock Hamiltonian graph networks with {ODE} integrators.

\bibitem[Sedehi et~al., 2019]{SEDEHI2019659}
Sedehi, O., Papadimitriou, C., Teymouri, D., and Katafygiotis, L.~S. (2019).
\newblock Sequential {Bayesian} estimation of state and input in dynamical
  systems using output-only measurements.
\newblock {\em Mechanical Systems and Signal Processing}, 131:659--688.

\bibitem[Shadabfar et~al., 2022]{shadabfar2022resilience}
Shadabfar, M., Mahsuli, M., Zhang, Y., Xue, Y., Ayyub, B.~M., Huang, H., and
  Medina, R.~A. (2022).
\newblock Resilience-based design of infrastructure: Review of models,
  methodologies, and computational tools.
\newblock {\em ASCE-ASME Journal of Risk and Uncertainty in Engineering
  Systems, Part A: Civil Engineering}, 8(1):03121004.

\bibitem[Shaikhina et~al., 2015]{shaikhina2015machine}
Shaikhina, T., Lowe, D., Daga, S., Briggs, D., Higgins, R., and Khovanova, N.
  (2015).
\newblock Machine learning for predictive modelling based on small data in
  biomedical engineering.
\newblock {\em IFAC-PapersOnLine}, 48(20):469--474.

\bibitem[Sharifani and Amini, 2023]{sharifani2023machine}
Sharifani, K. and Amini, M. (2023).
\newblock Machine learning and deep learning: A review of methods and
  applications.
\newblock {\em World Information Technology and Engineering Journal},
  10(07):3897--3904.

\bibitem[Shi and Chatzis, 2022]{SHI2022108345}
Shi, X. and Chatzis, M. (2022).
\newblock An efficient algorithm to test the observability of rational
  nonlinear systems with unmeasured inputs.
\newblock {\em Mechanical Systems and Signal Processing}, 165:108345.

\bibitem[Simpson et~al., 2021]{simpson2021machine}
Simpson, T., Dervilis, N., and Chatzi, E. (2021).
\newblock Machine learning approach to model order reduction of nonlinear
  systems via autoencoder and {LSTM} networks.
\newblock {\em Journal of Engineering Mechanics}, 147(10):04021061.

\bibitem[Simpson et~al., 2023]{Simpson2021}
Simpson, T., Tsialiamanis, G., Dervilis, N., Worden, K., and Chatzi, E. (2023).
\newblock On the use of variational autoencoders for nonlinear modal analysis.
\newblock In Brake, M.~R., Renson, L., Kuether, R.~J., and Tiso, P., editors,
  {\em Nonlinear Structures {\&} Systems, Volume 1}, pages 297--300, Cham.
  Springer International Publishing.

\bibitem[Sohn, 2007]{Sohn2007}
Sohn, H. (2007).
\newblock Effects of environmental and operational variability on structural
  health monitoring.
\newblock {\em Philosophical Transactions of the Royal Society A: Mathematical,
  Physical and Engineering Sciences}, 365(1851):539--560.

\bibitem[Sohn et~al., 2003]{sohn2003review}
Sohn, H., Farrar, C.~R., Hemez, F.~M., Shunk, D.~D., Stinemates, D.~W., Nadler,
  B.~R., and Czarnecki, J.~J. (2003).
\newblock A review of structural health monitoring literature: 1996--2001.
\newblock {\em Los Alamos National Laboratory, USA}, 1:16.

\bibitem[Solin and S{\"a}rkk{\"a}, 2020]{solin2020hilbert}
Solin, A. and S{\"a}rkk{\"a}, S. (2020).
\newblock Hilbert space methods for reduced-rank {Gaussian} process regression.
\newblock {\em Statistics and Computing}, 30(2):419--446.

\bibitem[Sosanya and Greydanus, 2022]{sosanyadissipative2022}
Sosanya, A. and Greydanus, S. (2022).
\newblock Dissipative {Hamiltonian} neural networks: Learning dissipative and
  conservative dynamics separately.
\newblock {\em CoRR}, abs/2201.10085.

\bibitem[Sun et~al., 2023]{sun2023pisl}
Sun, F., Liu, Y., Wang, Q., and Sun, H. (2023).
\newblock {PiSL}: Physics-informed spline learning for data-driven
  identification of nonlinear dynamical systems.
\newblock {\em Mechanical Systems and Signal Processing}, 191:110165.

\bibitem[Sun et~al., 2021]{sun2021machine}
Sun, H., Burton, H.~V., and Huang, H. (2021).
\newblock Machine learning applications for building structural design and
  performance assessment: State-of-the-art review.
\newblock {\em Journal of Building Engineering}, 33:101816.

\bibitem[Sun et~al., 2020]{sun2020surrogate}
Sun, L., Gao, H., Pan, S., and Wang, J.-X. (2020).
\newblock Surrogate modeling for fluid flows based on physics-constrained deep
  learning without simulation data.
\newblock {\em Computer Methods in Applied Mechanics and Engineering},
  361:112732.

\bibitem[Tatsis et~al., 2022a]{TATSIS2022108558}
Tatsis, K., Agathos, K., Chatzi, E., and Dertimanis, V. (2022a).
\newblock A hierarchical output-only {Bayesian} approach for online
  vibration-based crack detection using parametric reduced-order models.
\newblock {\em Mechanical Systems and Signal Processing}, 167:108558.

\bibitem[Tatsis et~al., 2023]{Tatsis_2023}
Tatsis, K., Dertimanis, V., and Chatzi, E. (2023).
\newblock On off-line and on-line {Bayesian} filtering for uncertainty
  quantification of structural deterioration.
\newblock {\em Journal of Structural Dynamics}.

\bibitem[Tatsis et~al., 2021]{TATSIS2021107223}
Tatsis, K., Dertimanis, V., Papadimitriou, C., Lourens, E., and Chatzi, E.
  (2021).
\newblock A general substructure-based framework for input-state estimation
  using limited output measurements.
\newblock {\em Mechanical Systems and Signal Processing}, 150:107223.

\bibitem[Tatsis et~al., 2022b]{tatsis2022hierarchical}
Tatsis, K.~E., Agathos, K., Chatzi, E., and Dertimanis, V.~K. (2022b).
\newblock A hierarchical output-only {Bayesian} approach for online
  vibration-based crack detection using parametric reduced-order models.
\newblock {\em Mechanical Systems and Signal Processing}, 167:108558.

\bibitem[Tchemodanova et~al., 2021]{tchemodanova2021strain}
Tchemodanova, S.~P., Sanayei, M., Moaveni, B., Tatsis, K., and Chatzi, E.
  (2021).
\newblock Strain predictions at unmeasured locations of a substructure using
  sparse response-only vibration measurements.
\newblock {\em Journal of Civil Structural Health Monitoring},
  11(4):1113--1136.

\bibitem[Teymouri et~al., 2023]{TEYMOURI2023109758}
Teymouri, D., Sedehi, O., Katafygiotis, L.~S., and Papadimitriou, C. (2023).
\newblock Input-state-parameter-noise identification and virtual sensing in
  dynamical systems: A {Bayesian} expectation-maximization {(BEM)} perspective.
\newblock {\em Mechanical Systems and Signal Processing}, 185:109758.

\bibitem[Thakolkaran et~al., 2022]{thakolkaran2022nn}
Thakolkaran, P., Joshi, A., Zheng, Y., Flaschel, M., De~Lorenzis, L., and
  Kumar, S. (2022).
\newblock {NN-EUCLID}: Deep-learning hyperelasticity without stress data.
\newblock {\em Journal of the Mechanics and Physics of Solids}, 169:105076.

\bibitem[Tibshirani, 1996]{tibshirani1996regression}
Tibshirani, R. (1996).
\newblock Regression shrinkage and selection via the lasso.
\newblock {\em Journal of the Royal Statistical Society Series B: Statistical
  Methodology}, 58(1):267--288.

\bibitem[Toreini et~al., 2020]{toreini2020relationship}
Toreini, E., Aitken, M., Coopamootoo, K., Elliott, K., Zelaya, C.~G., and
  Van~Moorsel, A. (2020).
\newblock The relationship between trust in ai and trustworthy machine learning
  technologies.
\newblock In {\em Proceedings of the 2020 conference on fairness,
  accountability, and transparency}, pages 272--283.

\bibitem[Van~der Meer et~al., 2012]{van2012computational}
Van~der Meer, F., Sluys, L., Hallett, S., and Wisnom, M. (2012).
\newblock Computational modeling of complex failure mechanisms in laminates.
\newblock {\em Journal of Composite Materials}, 46(5):603--623.

\bibitem[Vettori et~al., 2023a]{vettori2023adaptive}
Vettori, S., Di~Lorenzo, E., Peeters, B., Luczak, M., and Chatzi, E. (2023a).
\newblock An adaptive-noise augmented {Kalman} filter approach for input-state
  estimation in structural dynamics.
\newblock {\em Mechanical Systems and Signal Processing}, 184:109654.

\bibitem[Vettori et~al., 2023b]{vettori2023assessment}
Vettori, S., Lorenzo, E.~D., Peeters, B., and Chatzi, E. (2023b).
\newblock Assessment of alternative covariance functions for joint input-state
  estimation via {Gaussian} process latent force models in structural dynamics.

\bibitem[Vincent and Bengio, 2002]{vincent2002kernel}
Vincent, P. and Bengio, Y. (2002).
\newblock Kernel matching pursuit.
\newblock {\em Machine learning}, 48:165--187.

\bibitem[Vlachas et~al., 2022]{Vlachas2022}
Vlachas, P.~R., Arampatzis, G., Uhler, C., and Koumoutsakos, P. (2022).
\newblock Multiscale simulations of complex systems by learning their effective
  dynamics.
\newblock {\em Nature Machine Intelligence}, 4(4):359--366.

\bibitem[Wang et~al., 2021]{wang2021understanding}
Wang, S., Teng, Y., and Perdikaris, P. (2021).
\newblock Understanding and mitigating gradient flow pathologies in
  physics-informed neural networks.
\newblock {\em SIAM Journal on Scientific Computing}, 43(5):A3055--A3081.

\bibitem[Williams and Rasmussen, 2006]{williams2006gaussian}
Williams, C.~K. and Rasmussen, C.~E. (2006).
\newblock {\em {Gaussian} processes for machine learning}, volume~2.
\newblock MIT press Cambridge, MA.

\bibitem[Wu and Xiu, 2020]{wu2020data}
Wu, K. and Xiu, D. (2020).
\newblock Data-driven deep learning of partial differential equations in modal
  space.
\newblock {\em Journal of Computational Physics}, 408:109307.

\bibitem[Xiaowei et~al., 2021]{xiaowei2021physics}
Xiaowei, J., Shujin, L., and Hui, L. (2021).
\newblock Physics-enhanced deep learning methods for modelling and simulating
  flow fields.
\newblock {\em Chinese Journal of Theoretical and Applied Physics},
  53(10):2616--2629.

\bibitem[Xu et~al., 2022]{xu2022physics}
Xu, P.-F., Han, C.-B., Cheng, H.-X., Cheng, C., and Ge, T. (2022).
\newblock A physics-informed neural network for the prediction of unmanned
  surface vehicle dynamics.
\newblock {\em Journal of Marine Science and Engineering}, 10(2):148.

\bibitem[Yang et~al., 2020]{yang2020structure}
Yang, Y., Nagayama, T., and Xue, K. (2020).
\newblock Structure system estimation under seismic excitation with an adaptive
  extended {Kalman} filter.
\newblock {\em Journal of Sound and Vibration}, 489:115690.

\bibitem[Yin et~al., 2023]{yin2023bridge}
Yin, X., Huang, Z., and Liu, Y. (2023).
\newblock Bridge damage identification under the moving vehicle loads based on
  the method of physics-guided deep neural networks.
\newblock {\em Mechanical Systems and Signal Processing}, 190:110123.

\bibitem[Yoshida, 1990]{yoshidasymplectic1990}
Yoshida, H. (1990).
\newblock Construction of higher order symplectic integrators.
\newblock {\em Physics Letters A}, 150(5):262--268.

\bibitem[Yu et~al., 2020]{yu2020structural}
Yu, Y., Yao, H., and Liu, Y. (2020).
\newblock Structural dynamics simulation using a novel physics-guided machine
  learning method.
\newblock {\em Engineering Applications of Artificial Intelligence}, 96:103947.

\bibitem[Yuan et~al., 2020]{yuan2020machine}
Yuan, F.-G., Zargar, S.~A., Chen, Q., and Wang, S. (2020).
\newblock Machine learning for structural health monitoring: challenges and
  opportunities.
\newblock {\em Sensors and smart structures technologies for civil, mechanical,
  and aerospace systems 2020}, 11379:1137903.

\bibitem[Yucesan and Viana, 2020]{yucesan2020physics}
Yucesan, Y.~A. and Viana, F.~A. (2020).
\newblock A physics-informed neural network for wind turbine main bearing
  fatigue.
\newblock {\em International Journal of Prognostics and Health Management},
  11(1).

\bibitem[Zhang et~al., 2020]{zhang2020physics}
Zhang, E., Yin, M., and Karniadakis, G.~E. (2020).
\newblock Physics-informed neural networks for nonhomogeneous material
  identification in elasticity imaging.
\newblock {\em arXiv preprint arXiv:2009.04525}.

\bibitem[Zhang et~al., 2021]{zhang2021gaussian}
Zhang, S., Rogers, T.~J., and Cross, E.~J. (2021).
\newblock {Gaussian} process based grey-box modelling for {SHM} of structures
  under fluctuating environmental conditions.
\newblock In {\em European Workshop on Structural Health Monitoring: Special
  Collection of 2020 Papers-Volume 2}, pages 55--66. Springer.

\bibitem[Zhang and Ling, 2018]{zhang2018strategy}
Zhang, Y. and Ling, C. (2018).
\newblock A strategy to apply machine learning to small datasets in materials
  science.
\newblock {\em Npj Computational Materials}, 4(1):25.

\bibitem[Zhang and Sun, 2021]{zhang2021structural}
Zhang, Z. and Sun, C. (2021).
\newblock Structural damage identification via physics-guided machine learning:
  a methodology integrating pattern recognition with finite element model
  updating.
\newblock {\em Structural Health Monitoring}, 20(4):1675--1688.

\bibitem[Zhang and Duraisamy, 2015]{zhang2015machine}
Zhang, Z.~J. and Duraisamy, K. (2015).
\newblock Machine learning methods for data-driven turbulence modeling.
\newblock In {\em 22nd AIAA computational fluid dynamics conference}, page
  2460.

\bibitem[Zheng et~al., 2022]{zheng2022physics}
Zheng, B., Li, T., Qi, H., Gao, L., Liu, X., and Yuan, L. (2022).
\newblock Physics-informed machine learning model for computational fracture of
  quasi-brittle materials without labelled data.
\newblock {\em International Journal of Mechanical Sciences}, 223:107282.

\bibitem[Zhong et~al., 2021]{zhong2021machine}
Zhong, S., Zhang, K., Bagheri, M., Burken, J.~G., Gu, A., Li, B., Ma, X.,
  Marrone, B.~L., Ren, Z.~J., Schrier, J., et~al. (2021).
\newblock Machine learning: new ideas and tools in environmental science and
  engineering.
\newblock {\em Environmental Science \& Technology}, 55(19):12741--12754.

\bibitem[Zhong et~al., 2020]{zhong2020symplectic}
Zhong, Y.~D., Dey, B., and Chakraborty, A. (2020).
\newblock Symplectic {ODE-Net}: Learning hamiltonian dynamics with control.

\bibitem[Zhuang et~al., 2021]{zhuang2021deep}
Zhuang, X., Guo, H., Alajlan, N., Zhu, H., and Rabczuk, T. (2021).
\newblock Deep autoencoder based energy method for the bending, vibration, and
  buckling analysis of kirchhoff plates with transfer learning.
\newblock {\em European Journal of Mechanics-A/Solids}, 87:104225.

\bibitem[Zou et~al., 2023]{ZOU2023110488}
Zou, J., Lourens, E.-M., and Cicirello, A. (2023).
\newblock Virtual sensing of subsoil strain response in monopile-based offshore
  wind turbines via {Gaussian} process latent force models.
\newblock {\em Mechanical Systems and Signal Processing}, 200:110488.

\end{thebibliography}

\end{Backmatter}

\end{document}